\documentclass[11pt, a4paper, logo, copyright, nonumbering]{map}
\usepackage{CJKutf8}
\usepackage{xargs}  
\usepackage{todonotes}
\usepackage{amsmath}
\usepackage{dsfont}

\usepackage{longtable}

\usepackage{svg}
\usepackage{fontawesome}
\usepackage{array}
\usepackage{tabularx}
\usepackage{latexsym}
\usepackage{graphicx}
\usepackage[utf8]{inputenc} 
\usepackage[utf8]{inputenc} 
\usepackage{amssymb} 
\usepackage{url}            
\usepackage{amsfonts}       
\usepackage{nicefrac}       
\usepackage{microtype}      
\usepackage{xspace}

\usepackage{listings}
\usepackage{makecell}
\usepackage{inconsolata}
\usepackage{multicol}
\usepackage{amstext}

\definecolor{darkblue}{RGB}{84, 112, 198}

\definecolor{lightblue}{rgb}{0.85, 0.95, 1.0}    
\definecolor{lightgreen}{rgb}{0.90, 1.0, 0.90}    
\definecolor{lightorange}{rgb}{1.0, 0.95, 0.85}   
\definecolor{lightpurple}{rgb}{0.95, 0.90, 1.0}   
\definecolor{lightgray}{rgb}{0.97, 0.97, 0.97}    

\usepackage{tikz}
\usetikzlibrary{calc}
\definecolor{battery-empty}{rgb}{0.9, 0.9, 0.9}
\newcommand{\difficultybar}[1]{%
  \begin{tikzpicture}[baseline, scale=0.5, every node/.style={scale=0.8}]
    \foreach \i in {1,2,3,4,5} {
      \ifnum\i>#1
        \draw[fill=battery-empty] (\i*0.5-0.5, 0) rectangle (\i*0.5, 0.25);
      \else
        \pgfmathsetmacro{\colorlevel}{80 - 12*(\i)} 
        \edef\x{\noexpand\draw[fill=blue!\colorlevel!white, opacity=0.9] (\i*0.5-0.5, 0) rectangle (\i*0.5, 0.25);}
        \x
        \draw[blue!50!black] (\i*0.5-0.5, 0) rectangle (\i*0.5, 0.25);
      \fi
    }
    \fill[battery-empty!70] (2.5, 0.08) rectangle (2.6, 0.17);
    \draw[battery-empty!70!black] (2.5, 0.08) rectangle (2.6, 0.17);
  \end{tikzpicture}%
}
\renewcommand{\arraystretch}{0.96}

\definecolor{hidden-draw}{RGB}{20,68,106}
\definecolor{hidden-pink}{RGB}{255,245,247}
\usepackage[edges]{forest}

\usepackage{bm}

\usepackage{natbib}
\usepackage{CJKutf8}
\usepackage{xargs}
\usepackage{todonotes}
\usepackage{multirow}
\usepackage{cleveref}
\usepackage{subcaption}
\usepackage{url}
\usepackage{svg}
\usepackage{fontawesome}
\usepackage{xcolor}
\usepackage{float}

\usepackage[utf8]{inputenc}
\usepackage{booktabs}
\usepackage{graphicx}
\usepackage{array}
\usepackage{tabularx}
\usepackage{xcolor,colortbl}  
\definecolor{boxcolor}{HTML}{d92523} 
\definecolor{bulbcolor}{HTML}{e3b87f} 
\usepackage{url}
\usepackage{ragged2e}   
\usepackage{makecell} 

\usepackage{hyperref}       
\usepackage{url}            
\usepackage{booktabs}       
\usepackage{nicefrac}       
\usepackage{microtype}      
\usepackage{xcolor}         
\usepackage{graphicx}
\usepackage{longtable}
\usepackage{array}
\usepackage{pbox}
\usepackage{tabularx}
\usepackage{multirow}
\usepackage{wrapfig}
\usepackage{pifont}
\usepackage{lipsum}
\usepackage{subcaption}
\usepackage{enumitem}
\usepackage{tikz}
\usepackage{xstring}

\usepackage{color-edits}

\usepackage{booktabs}
\usepackage{multirow}
\usepackage[table]{xcolor}
\usepackage{tabularx}

\usepackage{parskip} 

\usepackage{threeparttable} 

\usepackage{algorithm}
\usepackage{algpseudocode}

\usepackage{xspace}
\usepackage{times}
\usepackage{latexsym}
\usepackage{booktabs}
\usepackage{fvextra}
\usepackage{graphicx}
\usepackage{subcaption} 
\usepackage[T1]{fontenc}
\usepackage{xcolor}
\usepackage[utf8]{inputenc}

\usepackage{microtype}

\usepackage{inconsolata}
\usepackage{amsmath}
\usepackage{booktabs}
\usepackage{graphicx}
\usepackage{pifont}
\usepackage{multirow}          
\usepackage{colortbl}          
\usepackage{algorithm}
\usepackage{algpseudocode}
\usepackage{xcolor}
\usepackage{amssymb}

\definecolor{rliableolive}{HTML}{BBCC33}
\definecolor{rliableblue}{HTML}{77AADD}
\definecolor{rliablered}{HTML}{f63c44}

\algrenewcommand\algorithmicrequire{\textbf{Input:}}
\algrenewcommand\algorithmicensure{\textbf{Output:}}
\algrenewcommand\alglinenumber[1]{\small #1:}

\newcommand{\algsection}[1]{\Statex \textcolor{blue}{\(\triangleright\)~#1}}
\newcommand{\algcommentgray}[1]{\Statex \hspace{\algorithmicindent}\textcolor{gray}{\textit{// #1}}}

\newcommand{\eg}{\textit{e.g.,}\xspace}

\newcommand{\paratitle}[1]{\vspace{1.5ex}\noindent\textbf{#1}}

\usepackage{alltt}
\usepackage[most,skins,theorems]{tcolorbox}
\tcbuselibrary{skins,breakable,fitting,hooks}  
\tcbset{
  aibox/.style={
    width=\linewidth,
    top=7pt,
    bottom=8pt,
    colback=white,
    colframe=black,
    colbacktitle=black,
    enhanced,
    center,
    attach boxed title to top left={yshift=-0.1in,xshift=0.15in},
    boxed title style={boxrule=0pt,colframe=white},
  }
}
\tcbset{
  aibox2/.style={
    width=\linewidth,
    top=7pt,
    bottom=2pt,
    colback=green!18!white,
    colframe=black,
    colbacktitle=black,
    enhanced,
    center,
    attach boxed title to top left={yshift=-0.1in,xshift=0.15in},
    boxed title style={boxrule=0pt,colframe=white,},
  }
}
\newtcolorbox{AIbox}[2][]{aibox,title=#2,#1}
\newtcolorbox{AIbox2}[2][]{aibox2,title=#2,#1}

\usepackage{xcolor}

\hypersetup{
    colorlinks=true,            
    linkcolor=blue,             
    filecolor=magenta,          
    urlcolor=cyan,              
    citecolor=purple,             
    pdftitle={Overleaf Example},
    pdfpagemode=FullScreen,
}

\definecolor{iquestblue}{HTML}{173C7F}
\definecolor{iquestazure}{HTML}{528FCC}

\newcommandx{\info}[2][1=]{\todo[linecolor=red,backgroundcolor=red!25,bordercolor=red,#1]{#2}}

\title{
\vspace{-0.2in}
\centering \fontsize{15pt}{16pt}\selectfont

Evolving with Open Worlds: Towards Recursive Self-Improvement for Future Forecasting

\vspace{-0.2in}
}

\renewcommand{\thefootnote}{\fnsymbol{footnote}}

\author{
Zhipeng Chen\textsuperscript{1,2*$^\S$},
Yunsen Liang\textsuperscript{1,2*$^\S$},
Yanxi Chen\textsuperscript{1},
Hongkang Zhao\textsuperscript{1},
Lingyun Feng\textsuperscript{1},\\
\normalfont
Ruiyang Ren\textsuperscript{3},
Guoxin Chen\textsuperscript{1},
Jia Deng\textsuperscript{1,2$^\S$},
Feng Chang\textsuperscript{2},
Chuan Hao\textsuperscript{2},\\
\normalfont
Ran Tao\textsuperscript{2},
Jian Yang\textsuperscript{2},
Ji-Rong Wen\textsuperscript{1$\dagger$},
Bryan Dai\textsuperscript{2$\dagger$},
Wayne Xin Zhao\textsuperscript{1$\dagger$}\\
\normalfont
\textsuperscript{1}Gaoling School of Artificial Intelligence, Renmin University of China \\
\normalfont
\textsuperscript{2}IQuest Research\quad
\textsuperscript{3}Nanyang Technological University \\
\normalfont
\normalsize{\textsuperscript{*}Equal contributors\quad
\textsuperscript{$\dagger$}Corresponding Authors \\
\normalfont
\textsuperscript{$\S$}Work done during an internship at IQuest Research. \\
\normalfont
Email: \texttt{\{zhipeng\_chen,yunsenliang04,jrwen\}@ruc.edu.cn}, \\
\normalfont
\texttt{batmanfly@gmail.com}
}}

\newcommand{\ignore}[1]{}

\usepackage[most]{tcolorbox}
\definecolor{darkorange}{RGB}{255, 140, 0}
\definecolor{lightgreen}{RGB}{145, 204, 117}
\definecolor{lightyellow}{RGB}{250, 200, 88}
\definecolor{lightred}{RGB}{238, 102, 102}
\definecolor{lightblue}{RGB}{115, 192, 222}

\definecolor{gray_1}{HTML}{B7B7B7}
\definecolor{gray_2}{HTML}{F0F0F0} 
\definecolor{frame_blue}{HTML}{A9D18E}

\newtcolorbox[auto counter, number within=section]{PromptBox}[2][]{
    enhanced,
    breakable,
    colback=gray_2, 
    colframe=gray_1,
    coltitle=white,
    fontupper=\small,
    fonttitle=\bfseries,
    title={#2},
    label={#1},
    arc=2pt,
    boxrule=1pt,
    left=2mm, right=2mm, top=2mm, bottom=2mm,
}

\definecolor{caseblue}{HTML}{E8F1FB}       
\definecolor{caseblueframe}{HTML}{7FA8D9}  
\newtcolorbox[use counter from=PromptBox]{CasePromptBox}[2][]{
    enhanced,
    breakable,
    colback=caseblue,
    colframe=caseblueframe,
    coltitle=white,
    colbacktitle=caseblueframe,
    fontupper=\small,
    fonttitle=\bfseries,
    title={#2},
    label={#1},
    arc=2pt,
    boxrule=1pt,
    left=2mm, right=2mm, top=2mm, bottom=2mm,
}

\usepackage{etoolbox}
\makeatletter
\patchcmd{\@maketitle}
  {\@toptitlebar}
  {%
    \hbox to \textwidth{%
    
    \includegraphics[height=0.7cm]{figures/aibox-logo.png}%
      \hfill
      \includegraphics[height=0.7cm]{figures/logo2.pdf}%
    }%
    \vspace{0.3em}
    \@toptitlebar
  }
  {}{}
\makeatother

\begin{abstract}
Most contemporary artificial intelligence~(AI) systems are released with frozen parameters and improve through external updates, whereas an ideal system would keep improving from its own experience, a capacity we call \emph{recursive self-improvement} (RSI).
Yet studying RSI requires an environment that is continuously changing, objectively verifiable, and naturally grounded, posing a fundamental challenge for agent evolution.
Here we introduce \textsc{EvoForesight}, a sustainable evolution agent system grounded in future event prediction, in which agents continuously interact with a dynamically evolving environment and improve by internalizing and externalizing the experience gained through these interactions.
A living environment synthesizes verifiable tasks, and a recursive-evolution engine turns the reasoning trajectory behind each prediction back into capability: decoupling search from prediction adapts parameter-level evolution to the contamination-prone live environment, while distilling reusable skills lets the agent evolve from the limited live trajectories such an environment affords.
On the FutureX leaderboard, which evaluates predictions against real-world outcomes and is updated weekly, our self-improved forecaster ranked first across all systems over three consecutive weeks, achieving up to a 195\% improvement in rating. On our synthesized evaluation, it ranked among the top three across four base LLMs, outperforming the strongest baseline by up to 10.7 Bradley--Terry points.
More broadly, these results suggest that a system built for self-improvement is best understood as an agent that co-evolves with a genuinely living, open-ended environment, pointing to a shift in AI from training static models on fixed data toward cultivating agents that grow stronger from their own experience, with such an open world as the indispensable basis of that growth.
Our resources are available at  \url{https://github.com/RUCAIBox/EvoForesight}.
\end{abstract}

\begin{document}
\maketitle

\let\oldthefootnote\thefootnote



\section{Introduction}
\label{sec:introduction}

Intelligence is often measured by what a system can do at a single moment in time~\citep{ouyang2022instructgpt,bai2022constitutional}.
Yet the more consequential question may not be how capable a model is today, but whether it can become more capable through its own experience~\citep{deepseekmath,guo2025deepseekr1,lambert2024tulu3}.
Across the natural and social sciences, the systems we regard as most powerful (\eg biological evolution, the scientific method, human expertise) are precisely those that improve themselves cumulatively, correcting past errors and compounding past gains~\citep{schmidhuber2003godel,clune2019aigas,silver2025era}.
This suggests that an ideal artificial system should be measured less by what it can do at any single moment than by how its capabilities form and grow over time.
A system that improves itself from experience is a prerequisite for AI that acts as an autonomous agent and as a genuine instrument of discovery.
In this work, we study this form of \emph{recursive self-improvement} (RSI) in the concrete and demanding setting of real-world event forecasting.

\begin{figure}[p]
    \centering
    \includegraphics[width=\linewidth]{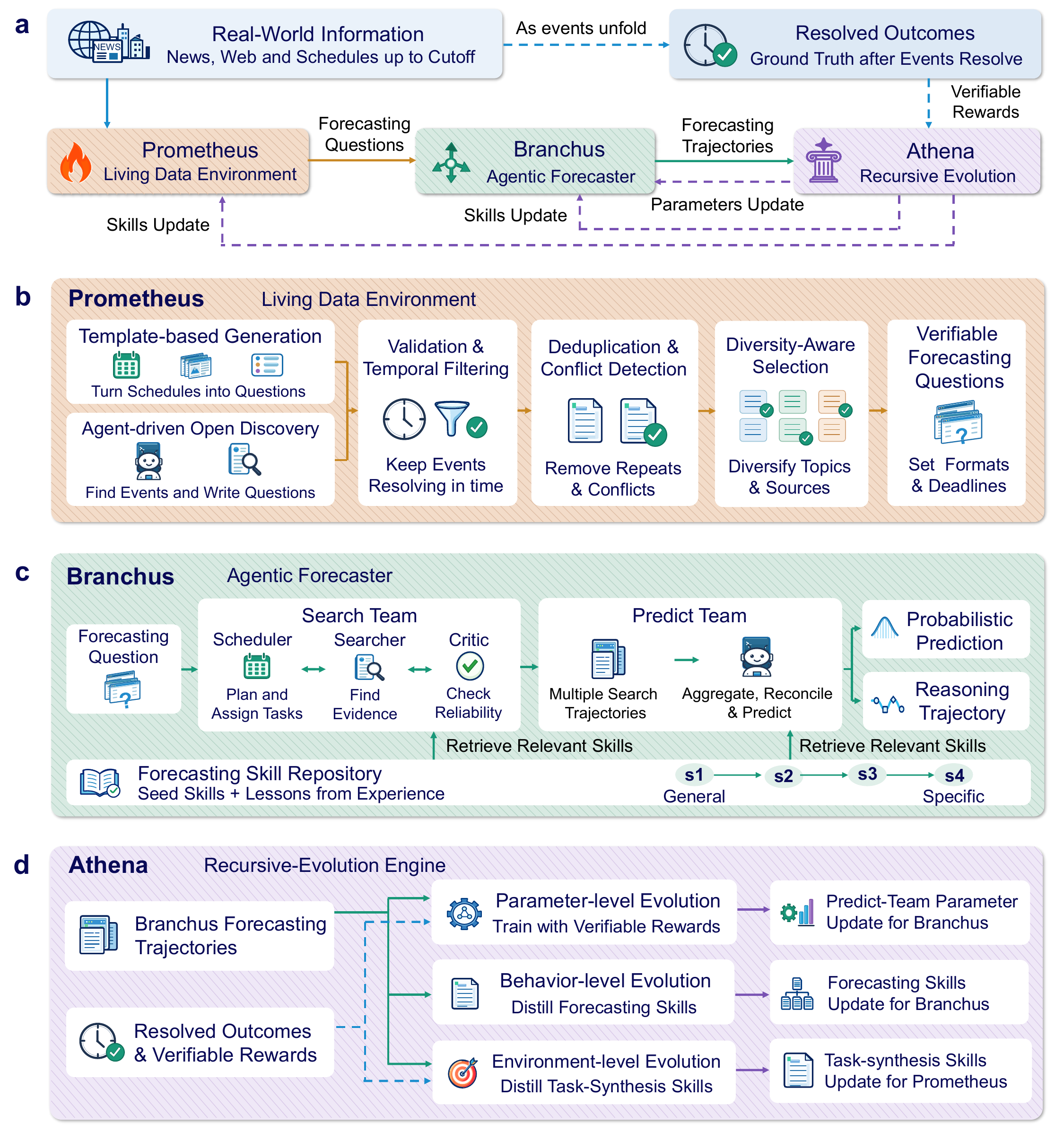}
    \captionsetup{font=small}
    \caption{\textbf{Overview of \textsc{EvoForesight} for recursive self-improvement in real-world event forecasting.}
    \textbf{a,} The closed loop connects question generation, forecasting, and evolution, with resolved outcomes supplying verifiable rewards and updates feeding into subsequent rounds.
    \textbf{b,} \textsc{Prometheus} generates questions through templates and agent-driven discovery, followed by temporal validation, deduplication, conflict detection, and diversity-aware selection.
    \textbf{c,} \textsc{Branchus} separates evidence gathering, by a \emph{search team} of a scheduler, searcher, and critic, from probabilistic prediction, by a \emph{predict team}. The search team is time-sensitive, as its result depends on when it runs, while the predict team is time-invariant, reasoning only over the search team's cached evidence and reasoning trajectories. Both teams retrieve relevant forecasting skills.
    \textbf{d,} \textsc{Athena} updates the predict team's parameters using cached evidence and verifiable rewards, distills search trajectories into reusable forecasting skills, and converts observed weaknesses into task-synthesis skills for \textsc{Prometheus}. Resolved outcomes inform parameter and environment updates but are withheld from the behavior-level judge.}
    \label{fig:overview}
\end{figure}

Studying recursive self-improvement, however, is not merely a matter of method.
It places unusually high demands on the research setting itself.
An agent can only improve recursively if its environment keeps posing fresh challenges and can tell, without human intervention, whether the agent is doing better than before.
What this calls for is an interactive environment, yet the fixed datasets that dominate today's evaluation offer no such thing: inert and unchanging, they neither renew the challenge nor judge progress on their own.
Real-world event forecasting, we argue, is a natural testbed for such study: it is \emph{grounded} in the unfolding world rather than a hand-constructed dataset, \emph{verifiable} through future events whose objective ground truth arrives with time, and \emph{non-stationary} because the world keeps changing. 
Together these properties close the loop that recursion requires: the agent observes the world, makes predictions, receives future outcomes, learns from the resulting experience, and improves the predictions it makes next (Figure~\ref{fig:overview}a).
Whereas prior work on event and time-series forecasting~\citep{zou2022forecasting,halawi2024forecasting} has largely pursued single-shot predictive accuracy, we instead treat forecasting as an environment in which to study how agents improve recursively through continuous interaction with a changing world.

Succeeding in this environment first requires a few foundational capabilities for solving any single forecasting task.
An agent must be able to \emph{gather} task-relevant evidence from a vast information space, as exemplified by search-augmented systems~\citep{song2025r1,yao2022react}, and to \emph{reason}, combining and weighing that evidence to draw multi-step conclusions~\citep{chen2025empirical,bai2026towards}.
Both capabilities are already standard in existing systems, and we treat them as the foundation on which our self-improvement mechanisms build.
Recursive self-improvement goes further, requiring a capability different in kind: an agent that not only solves the task in front of it but also changes its own behavior in light of past successes and failures. Such an agent addresses a distinct question: how its capacity to solve tasks can itself keep improving. 
We take this ability to \emph{evolve} as the central concern of this work, and reserve the name \textsc{EvoForesight} for the framework we propose to realize it (Section~\ref{sec:approach}).

Realizing such an evolving agent runs against a basic limitation of today's AI: it is largely static and dependent on outside intervention to advance, a set of parameters frozen at release rather than a system that keeps improving on its own~\citep{silver2025era,bubeck2023sparks}.
This shows up on two fronts. Evaluation is static, scoring a model on a fixed dataset~\citep{sweb-verified,merrill2026terminalbench} that can be memorized or contaminated; and improvement is passive, with capability gains driven from outside by human-curated data and offline retraining~\citep{hinton2015distilling,chung2024flan} rather than by the model's own experience.
This condition also differs from neighboring lines of work: self-training on self-generated data~\citep{zelikman2022star,huang2023selfimprove} lacks an external, verifiable ground truth and can reinforce a model's own biases; continual learning~\citep{kirkpatrick2017ewc,parisi2019continual} operates over static or artificially partitioned task streams rather than real, non-stationary feedback; and reinforcement learning for reasoning~\citep{deepseekmath,schulman2017proximal} optimizes reward on a task but does not, by itself, target improvement that compounds across time.
By \emph{recursive}, we mean exactly this compounding: the parameters, skills, and environment produced by one round of improvement become the input to the next, forming a self-improvement loop rather than a one-off gain~\citep{schmidhuber2003godel,clune2019aigas}.

Our central argument departs from most work on self-improvement, which pushes on a single lever, be it the model's weights~\citep{huang2023selfimprove}, its prompts~\citep{fernando2024promptbreeder}, or its memory~\citep{shinn2023reflexion}. What makes self-improvement \emph{sustainable}, we argue, is a system that keeps accumulating experience on its own rather than being fed hand-curated data from outside, and our answer is a \emph{living environment} in which the agent continually acts, observes outcomes, and learns. The logic is directional. The environment is the source, posing progressively harder tasks that supply the experience without which no further improvement is possible; the agent's \emph{parameters} and \emph{skills} are the two forms its evolution takes, both shaped from one substrate, the forecasting \emph{trajectories} that are reinforced into parameters, distilled into skills, and mined for the weaknesses that build a harder environment. The three thus couple into a single self-sustaining loop, in which the environment evolves to let the agent keep evolving.

Building on this view, we introduce \textsc{EvoForesight}, an event-forecasting framework for studying recursive self-improvement in a continuously evolving real-world environment, in which the agent and its environment co-evolve.
\textsc{EvoForesight} comprises three modules arranged as a closed loop rather than a pipeline.
\textsc{Prometheus}, a \emph{living data environment}, continually ingests real-world events to form the non-stationary, verifiable world in which forecasting takes place.
\textsc{Branchus}, an \emph{agentic forecaster}, produces predictions together with their full trajectories, generating the learning substrate described above.
\textsc{Athena}, the \emph{recursive-evolution engine}, learns from these trajectories through reinforcement learning and skill accumulation to drive self-improvement.
The loop closes when \textsc{Athena}'s improvements strengthen \textsc{Branchus}'s forecasts, and these stronger forecasts in turn place new demands on \textsc{Prometheus}, pushing the environment to evolve and sustaining an open-ended cycle of agent-environment co-evolution.

This co-evolution produces a strong forecasting system. Across all four LLMs, \textsc{Branchus} ranks among the top three overall and leads its block, and because every result is scored against real-world outcomes, this consistency reflects an advantage intrinsic to the design (BT of $62.2$ on \textsc{Qwen3.8-Max}, $+10.7$ over the strongest baseline in Table~\ref{tab:main_results}). 
More importantly, the gains compound rather than saturate: as the agent's own live trajectories accumulate into a growing pool, the skills distilled from them lift forecasting quality round by round rather than jumping once and plateauing (Figure~\ref{fig:skill_performance}).
Our experiments trace this improvement to how environment, parameters, and skills evolve together, each by its own means.
Reinforcement learning sharpens the \emph{parameters}, tightening probabilistic judgment beyond what accuracy registers (the numeric calibration score improves by $0.31$--$0.38$ while choice accuracy moves by only $1.3$ points in Figure~\ref{fig:rl_results}), yet its reach is bounded by the experience available.
The experience distilled into \emph{skills} lifts capability without touching the weights, learning from only limited trajectories rather than repeated rollouts and so sidestepping the instability that gradient updates can introduce.
The \emph{environment} closes the loop: when it co-evolves with the agent, synthesizing verifiable tasks aimed at the agent's demonstrated weaknesses, these otherwise isolated and bounded gains add up to a larger improvement (Figure~\ref{fig:env_evolution}).
This is evidence that recursive self-improvement arises from the three levers evolving as one.

Taken together, these findings establish recursive self-improvement as the co-evolution of an agent and its living environment. The environment is the abstraction we commit to. Our first contribution is to construct a \emph{real, dynamic, and verifiable} environment for forecasting: grounded in the unfolding world, non-stationary as that world changes, and scored against outcomes that resolve on their own. This environment lets all three levers improve in concert. Its verifiable outcomes become rewards that sharpen \emph{parameters}; its live experience supplies the trajectories that distil into \emph{skills}; and its dynamic form lets it be rewritten toward the agent's exposed weaknesses, so each round feeds the next. Our second contribution turns this opening into method: \textsc{EvoForesight} formalizes the environment, the agent, and the recursive-evolution engine as three modules in a closed loop (Section~\ref{sec:approach}). More broadly, the development of AI may shift from training static models on fixed datasets toward cultivating evolving agents within living environments, and with it the measure of progress: the rate, ceiling, and durability of a system's improvement, which only an evolving environment can reveal.

Forecasting is the concrete setting we build and study, and the way we construct this environment, and the methodology that evolves an agent inside it, carry over to any setting that is grounded, verifiable, and non-stationary, so the same loop can be extended to far more general scenarios.
It also recasts how recursive self-improvement might be pursued, suggesting that progress may come less from any single, ever-larger model than from co-evolving parameters, skills, and the environment that supplies their experience.
Such progress is not without cost: it remains bounded by the quality and quantity of available experience, long-horizon evaluation is expensive, and stability under non-stationary drift is not guaranteed.
Yet even bounded, it reframes the question with which we began. The deepest measure of intelligence may lie less in what a system can do at a single moment than in whether it can grow stronger from its own experience, and our study offers concrete evidence that such growth can be made to occur, and to be measured, in a real and verifiable world.

\clearpage

\section{Results}
\label{sec:experiment}

\subsection{Experimental Design and \textsc{EvoForesight} Framework}


To forecast future events and to test whether an agent can improve itself from doing so, we develop \textsc{EvoForesight}, a framework that turns experience into capability through a closed loop of three components.
\textsc{Prometheus}, the living data environment, gathers daily news and synthesizes it into forecasting tasks whose ground truth is revealed only once the events resolve.
\textsc{Branchus}, the agentic forecaster, gathers evidence with external tools and estimates the probability of each event, producing a full prediction trajectory.
\textsc{Athena}, the recursive-evolution engine, consumes these trajectories and drives improvement along three axes.
It applies reinforcement learning with verifiable rewards (RLVR) to update the parameters of \textsc{Branchus}, distills trajectories into reusable forecasting skills, and evolves adversarial task-synthesis skills in \textsc{Prometheus} to make the environment harder.
Each forecasting trajectory is rich enough to feed all three axes at once: because a future-event prediction forces the agent to gather evidence and weigh it under uncertainty, the same trajectory can be reinforced against its resolved outcome, mined for reusable skills, and analysed for the weaknesses from which a harder environment is built.
Experience is thus fed back to improve the agent's parameters, its skills, and its environment jointly, realizing the recursive self-improvement loop at the heart of this work.
The subsections that follow examine these three axes in turn, after first establishing the strength of the system as a whole.

We draw forecasting tasks from two complementary sources.
The first is FutureX~\citep{zeng2025futurex}, an external benchmark refreshed every week that asks models to predict events in the coming week.
Its tasks span four difficulty levels (L1 to L4), take multiple-choice, ranking, and numerical forms, and cover domains from finance and politics to sports, natural disasters, and technology, so strong performance reflects general forecasting ability.
The second is a higher-throughput evaluation synthesized by \textsc{Prometheus}, used purely as a task generator with its adversarial evolution disabled so the evaluation stays unbiased.
Both sources are genuinely live rather than static or cached.
Each day, \textsc{Prometheus} ingests fresh news from $125$ websites and writes forecasting questions about events that resolve the following day, so \textsc{Branchus} must predict outcomes whose ground truth does not yet exist.
This one-day horizon shortens the verification cycle and lets us evaluate models far more frequently than a weekly benchmark allows.

Every question concerns a future event, so predictions are verified against realized outcomes rather than a fixed answer key.
For the \textsc{Prometheus} evaluation, once an event resolves we task Codex~\citep{openai2026codex} with retrieving the realized outcome from the open web and scoring the prediction against it.
Beyond raw accuracy, we report a \emph{calibration} metric: the model outputs a probability for each event, and we check whether the empirical frequency of those events matches the predicted probabilities.
We also stabilize these per-question scores before aggregating.
On FutureX we adopt its official sliding-window Bradley--Terry (BT) model~\citep{bradley1952rank}, which infers a consistent latent strength from head-to-head weekly results.
On \textsc{Prometheus}, a single day of questions is too small and too variable to rank harnesses reliably, so unless stated otherwise every reported number averages over several consecutive days of live evaluation.

\subsection{Recursive evolution lifts \textsc{EvoForesight} to the strongest forecasting system}
\label{sec:2.3_main_results}


\begin{figure*}[t]
    \centering
    \includegraphics[width=\linewidth]{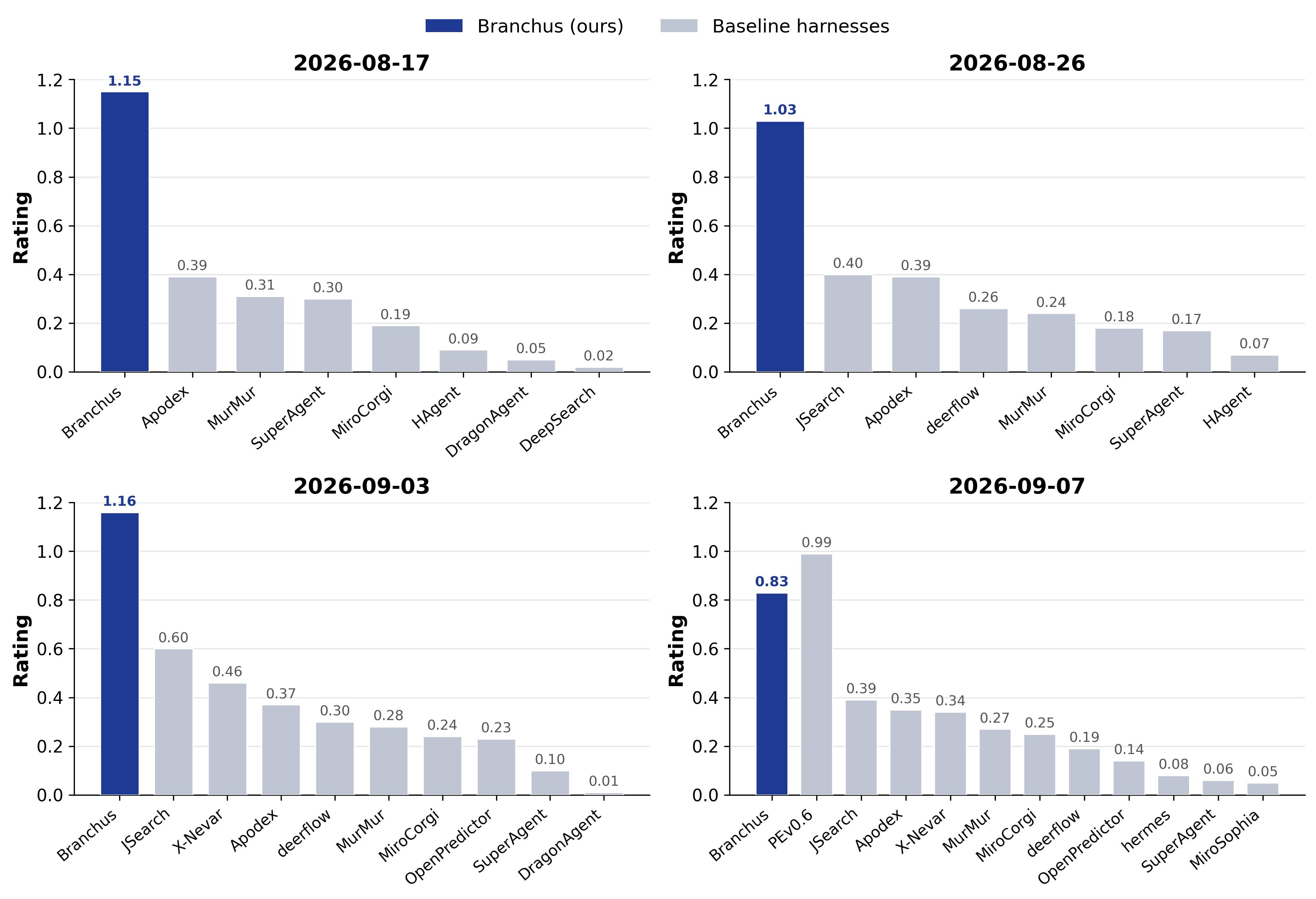}
    \caption{\textbf{Overall FutureX ranking across four weekly evaluations.} Each panel reports the sliding-window Bradley--Terry rating of every harness for one week, with \textsc{Branchus} (ours) shown against the baseline harnesses. The complete official leaderboards are provided in Appendix~\ref{app:futurex_leaderboard} (Figure~\ref{app:fig_futurex_leaderboards}).}
    \label{fig:futurex_overall}
\end{figure*}

\begin{table*}[t]
    \centering
    \small
    \renewcommand{\arraystretch}{0.95}
    \caption{Forecasting performance of \textsc{Branchus} against baseline harnesses across four LLMs on the \textsc{Prometheus}-synthesized evaluation. We report accuracy, Brier, and AUROC for choice questions, relative WIS for numeric questions, and the Bradley--Terry score (BT) with its rank overall. \textit{The rank is a single ordering over all 23 harness--LLM systems in the table, not a within-block rank}. Best values per column within each base-LLM block are in \textbf{bold}.}
      \begin{tabular}{l cccc cc}
        \toprule
        \multirow{2.5}*{\textbf{Harness}} & \multicolumn{3}{c}{\textbf{Choice Questions}} & \multicolumn{1}{c}{\textbf{Numeric Questions}} & \multicolumn{2}{c}{\textbf{Overall Performance}} \\
        \cmidrule(lr){2-4}\cmidrule(lr){5-5}\cmidrule(lr){6-7}
        & Accuracy $\uparrow$ & Brier $\downarrow$ & AUROC $\uparrow$ & Relative WIS $\downarrow$ & BT $\uparrow$ & Rank $\downarrow$ \\
        \midrule
        \multicolumn{7}{c}{\textsc{Qwen3.8-Max}} \\
        \textsc{ReAct} & 70.7 & 0.346 & 0.784 & 0.913 & 47.9 & 16 \\
        \textsc{MAS} & 70.3 & 0.343 & \textbf{0.808} & 0.871 & 51.5 & 11 \\
        \textsc{OpenClaw} & 71.9 & 0.356 & 0.749 & 0.933 & 47.1 & 18 \\
        \textsc{DeerFlow} & 71.9 & 0.354 & 0.748 & 0.927 & 45.5 & 20 \\
        \textsc{MiroFlow} & 70.3 & 0.351 & 0.790 & 0.900 & 50.6 & 13 \\
        \textsc{Branchus} (Ours) & \textbf{72.3} & \textbf{0.336} & 0.787 & \textbf{0.737} & \textbf{62.2} & \textbf{1} \\
        \midrule
        \multicolumn{7}{c}{\textsc{GPT-5.6-Sol}} \\
        \textsc{ReAct} & 70.3 & 0.340 & 0.812 & 0.915 & 53.4 & 7 \\
        \textsc{MAS} & \textbf{72.3} & 0.342 & 0.776 & 0.913 & 55.0 & 5 \\
        \textsc{OpenClaw} & 71.9 & 0.346 & 0.777 & 0.909 & 53.4 & 8 \\
        \textsc{MiroFlow} & 69.5 & 0.370 & 0.765 & 4.152 & 40.7 & 23 \\
        \textsc{Branchus} (Ours) & 71.5 & \textbf{0.332} & \textbf{0.813} & \textbf{0.865} & \textbf{57.9} & \textbf{2} \\
        \midrule
        \multicolumn{7}{c}{\textsc{DeepSeek-V4-Pro}} \\
        \textsc{ReAct} & 71.5 & 0.350 & 0.769 & 0.848 & 53.1 & 10 \\
        \textsc{MAS} & \textbf{72.7} & \textbf{0.340} & 0.773 & \textbf{0.823} & 55.9 & 4 \\
        \textsc{OpenClaw} & 71.5 & 0.363 & 0.733 & 0.920 & 48.2 & 14 \\
        \textsc{DeerFlow} & 70.3 & 0.371 & 0.732 & 0.882 & 47.4 & 17 \\
        \textsc{MiroFlow} & 69.9 & 0.351 & \textbf{0.787} & 0.851 & 53.9 & 6 \\
        \textsc{Branchus} (Ours) & 71.9 & 0.341 & 0.783 & 0.835 & \textbf{56.4} & \textbf{3} \\
        \midrule
        \multicolumn{7}{c}{\textsc{Claude-Opus-4.6}} \\
        \textsc{ReAct} & 72.3 & 0.355 & 0.739 & 1.106 & 42.0 & 21 \\
        \textsc{MAS} & 71.5 & 0.376 & 0.714 & 1.105 & 41.6 & 22 \\
        \textsc{OpenClaw} & \textbf{73.8} & 0.351 & 0.725 & \textbf{0.893} & 51.2 & 12 \\
        \textsc{DeerFlow} & 72.3 & 0.352 & 0.756 & 1.021 & 48.2 & 15 \\
        \textsc{MiroFlow} & 70.7 & 0.360 & 0.751 & 0.959 & 46.0 & 19 \\
        \textsc{Branchus} (Ours) & 73.0 & \textbf{0.331} & \textbf{0.793} & 0.900 & \textbf{53.1} & \textbf{9} \\
        \bottomrule
    \end{tabular}
    \label{tab:main_results}
\end{table*}

As promised in the introduction, \textsc{EvoForesight} attains strong and generalizable forecasting performance. The numbers here report \textsc{Branchus}, the module that performs inference, after the full recursive self-improvement of \textsc{EvoForesight}, so they are a system-level result with parameters, skills, and environment already co-evolved rather than a starting point. We pair the evolved system with four LLMs that differ in size, provider, and reasoning behavior (\textsc{Qwen3.8-Max}, \textsc{GPT-5.6-Sol}, \textsc{DeepSeek-V4-Pro}, and \textsc{Claude-Opus-4.6}) and compare against a diverse suite of baselines on the \textsc{Prometheus}-synthesized evaluation (Table~\ref{tab:main_results}; Section~\ref{sec:models_impl} for details), with FutureX giving the externally refreshed view below (Figure~\ref{fig:futurex_overall}). Regardless of the base model, \textsc{EvoForesight} is strongest: it attains the highest BT score and best rank in every block, and the four \textsc{Branchus} configurations occupy the top three positions overall, led by \textsc{Qwen3.8-Max} at a BT of $62.2$. The margin over the best baseline ranges from $+10.7$ to $+0.5$ BT but never changes sign. Because the evaluation resolves against real-world outcomes rather than a fixed answer key, this cross-model consistency indicates an advantage intrinsic to \textsc{EvoForesight} rather than a fortunate match between one design and one model. The subsections that follow isolate how each axis of self-improvement turns the agent's own experience into this frontier.

Model capability and agent design are complementary rather than substitutable.
A stronger LLM lifts the ceiling for every system, so the strongest result in the table is reached by placing \textsc{Branchus} on the most capable model (\textsc{Qwen3.8-Max}); conversely, \textsc{Branchus} rescues a weaker base, reaching rank~$3$ on \textsc{DeepSeek-V4-Pro} and outscoring every baseline irrespective of the model those baselines are built on.
The two effects compose and neither alone is sufficient to reach the frontier, which suggests that further gains require jointly improving the base model and the agent that scaffolds it, motivating the recursive-evolution mechanisms studied in the remainder of this paper.

Crucially, this advantage reflects well-calibrated judgement rather than over-confident guessing.
Raw accuracy alone does not separate the systems: on choice questions the baselines occasionally edge out \textsc{Branchus} by a few tenths of a point, for example \textsc{MAS} on \textsc{DeepSeek-V4-Pro} and \textsc{OpenClaw} on \textsc{Claude-Opus-4.6}.
The decisive gap instead appears on calibration and numeric sharpness: \textsc{Branchus} attains the best Brier score in three of the four blocks and the best relative WIS on both \textsc{Qwen3.8-Max} ($0.737$) and \textsc{GPT-5.6-Sol} ($0.865$), so its interval forecasts are tighter than the strongest baseline rather than merely competitive.
Because the Brier score and WIS jointly penalize over-confident wrong answers and unjustified hedging, \textsc{Branchus} does not trade calibration for accuracy; its probabilities and intervals faithfully reflect the evidence it gathered, which is what lifts it to the top of the BT ranking even where its top-$1$ accuracy is only matched.

Finally, this advantage is stable over time rather than an artifact of a single evaluation window.
Figure~\ref{fig:futurex_overall} tracks the sliding-window BT rating of every harness across four consecutive weekly evaluations, and \textsc{Branchus} (dark blue) sits at or near the top of the field in each panel, holding its lead as the task distribution is refreshed, whereas standard workflows are both lower and less steady, swinging more sharply between weeks and gaining less from a stronger LLM.
We attribute this gap to two related bottlenecks: these workflows execute fixed, largely one-pass evidence gathering, so additional model capability is spent re-reading the same limited context rather than expanding source coverage; and they lack an explicit mechanism for tracking hypotheses and calibrating confidence across steps, so long chains of reasoning accumulate uncorrected errors and drift toward over-confident predictions.
Together, these limitations explain why standard workflows saturate well below the frontier defined by \textsc{Branchus}, and why simply scaling up the LLM is insufficient to close the gap on long-horizon forecasting tasks.
This frontier is where our agent begins rather than ends: the remaining subsections show that recursively evolving \textsc{Branchus} pushes it beyond this already-leading base form, so that capability grows from the agent's own experience over successive rounds instead of jumping once and plateauing.

\subsection{Search-prediction decoupling adapts RFT and RLVR to contamination-prone live environments}


\begin{figure}[t]
    \centering
    \begin{subfigure}[t]{0.49\linewidth}
        \centering
        \includegraphics[width=\linewidth]{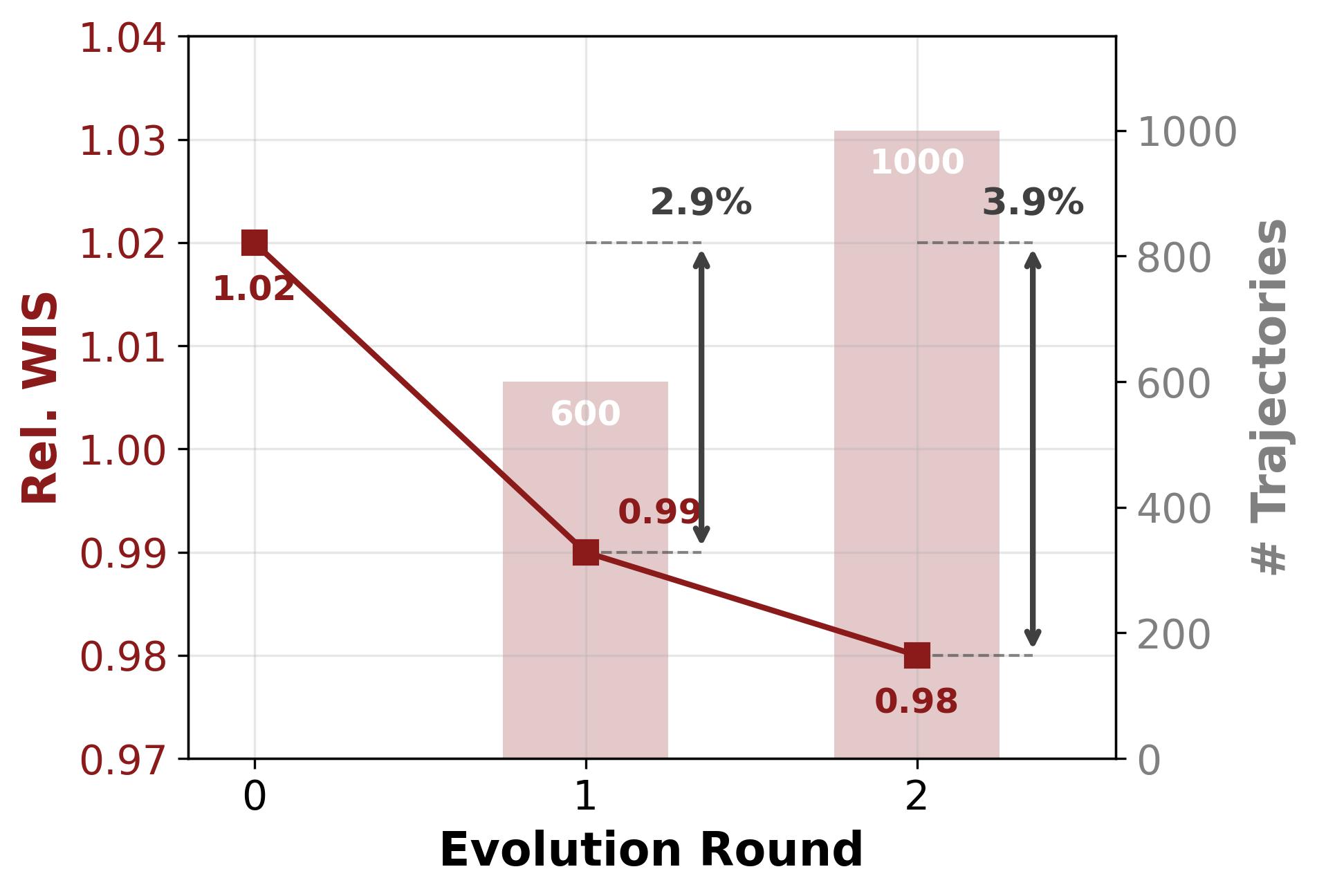}
        \caption{MAPE across evolution rounds.}
        \label{fig:para_performance_wis}
    \end{subfigure}
    \hfill
    \begin{subfigure}[t]{0.49\linewidth}
        \centering
        \includegraphics[width=\linewidth]{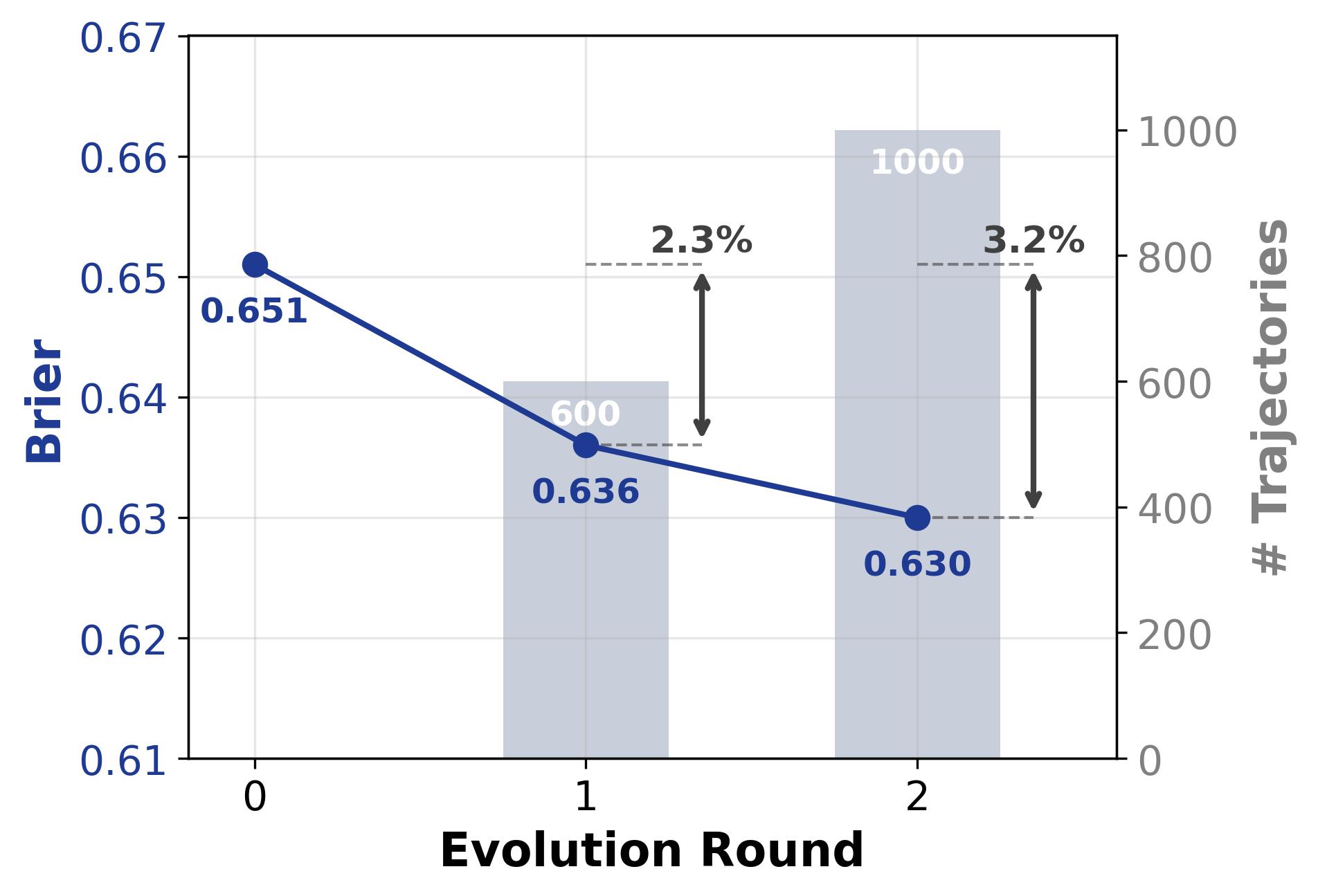}
        \caption{Brier score across evolution rounds.}
        \label{fig:para_performance_brier}
    \end{subfigure}
    \caption{\textbf{Parameter evolution improves forecasting quality continuously.} Across two evolution rounds, the training set grows from 600 to 1,000 examples, accompanied by consistent improvements in forecasting performance. \textbf{Left}: the performance of relative WIS. \textbf{Right}: calibration, measured by the Brier score. \emph{For both metrics, lower is better.}}
    \label{fig:param_rsi}
\end{figure}

\begin{figure}[t]
    \centering
    \includegraphics[width=\linewidth]{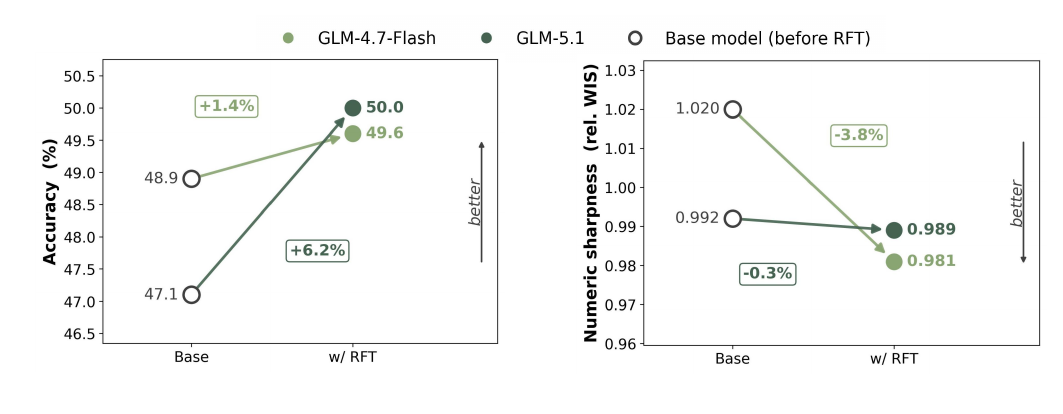}
    \caption{\textbf{RFT delivers consistent, calibration-led gains over the base model.} Each panel traces the Base\,$\rightarrow$\,RFT transition for two LLMs on one axis of forecasting quality, with every axis oriented so that upward movement is an improvement. Fine-tuning on successful trajectories lifts every axis, and the improvement is led by calibration rather than raw accuracy. \textbf{Left}: choice accuracy. \textbf{Right}: numeric interval forecasts, measured by relative WIS.}
    \label{fig:rft_improvement}
\end{figure}

\begin{figure}[t]
    \centering
    \begin{subfigure}[t]{0.49\linewidth}
        \centering
        \includegraphics[width=\linewidth]{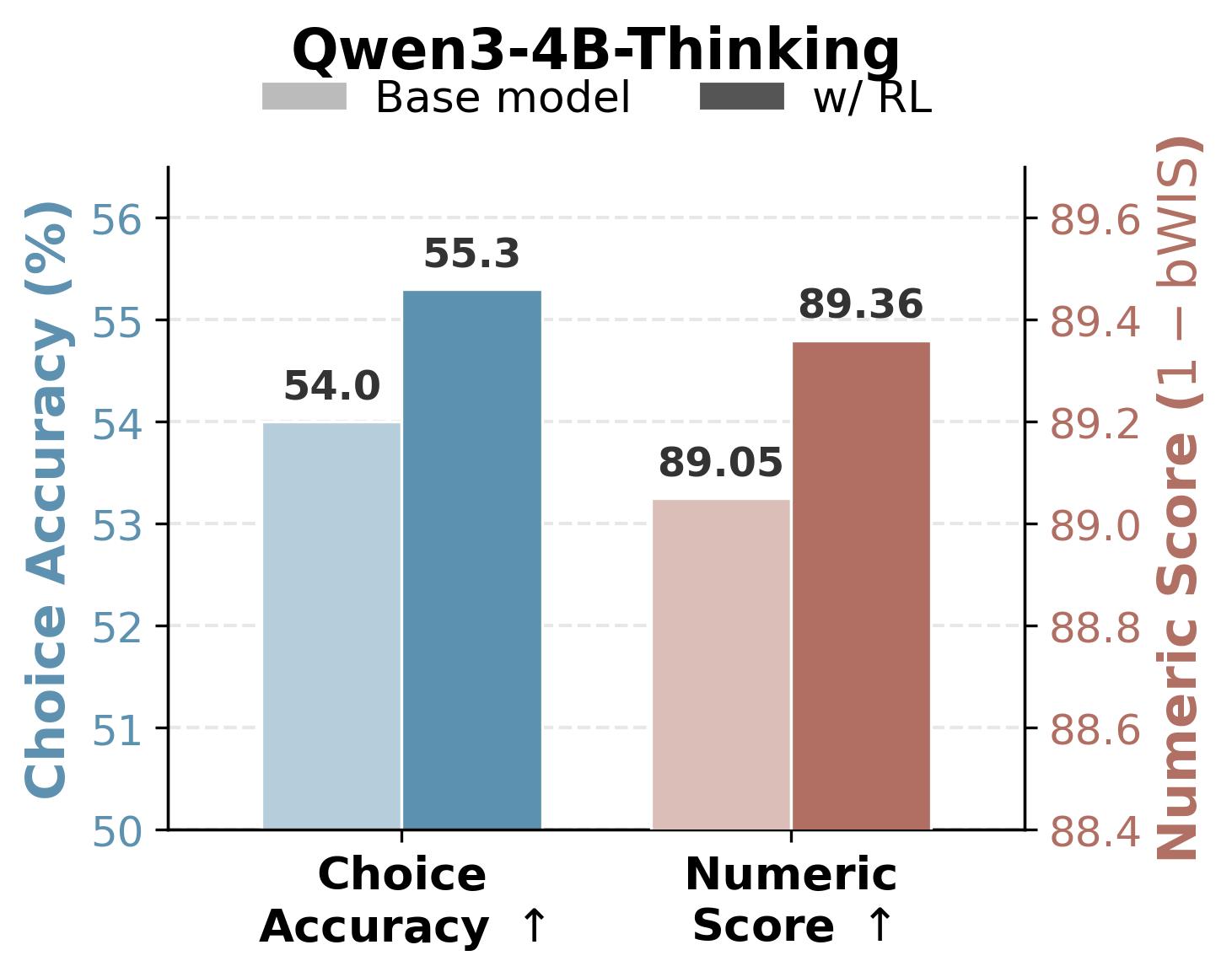}
        \caption{Qwen3-4B-Thinking.}
        \label{fig:rl_results_4b}
    \end{subfigure}
    \hfill
    \begin{subfigure}[t]{0.49\linewidth}
        \centering
        \includegraphics[width=\linewidth]{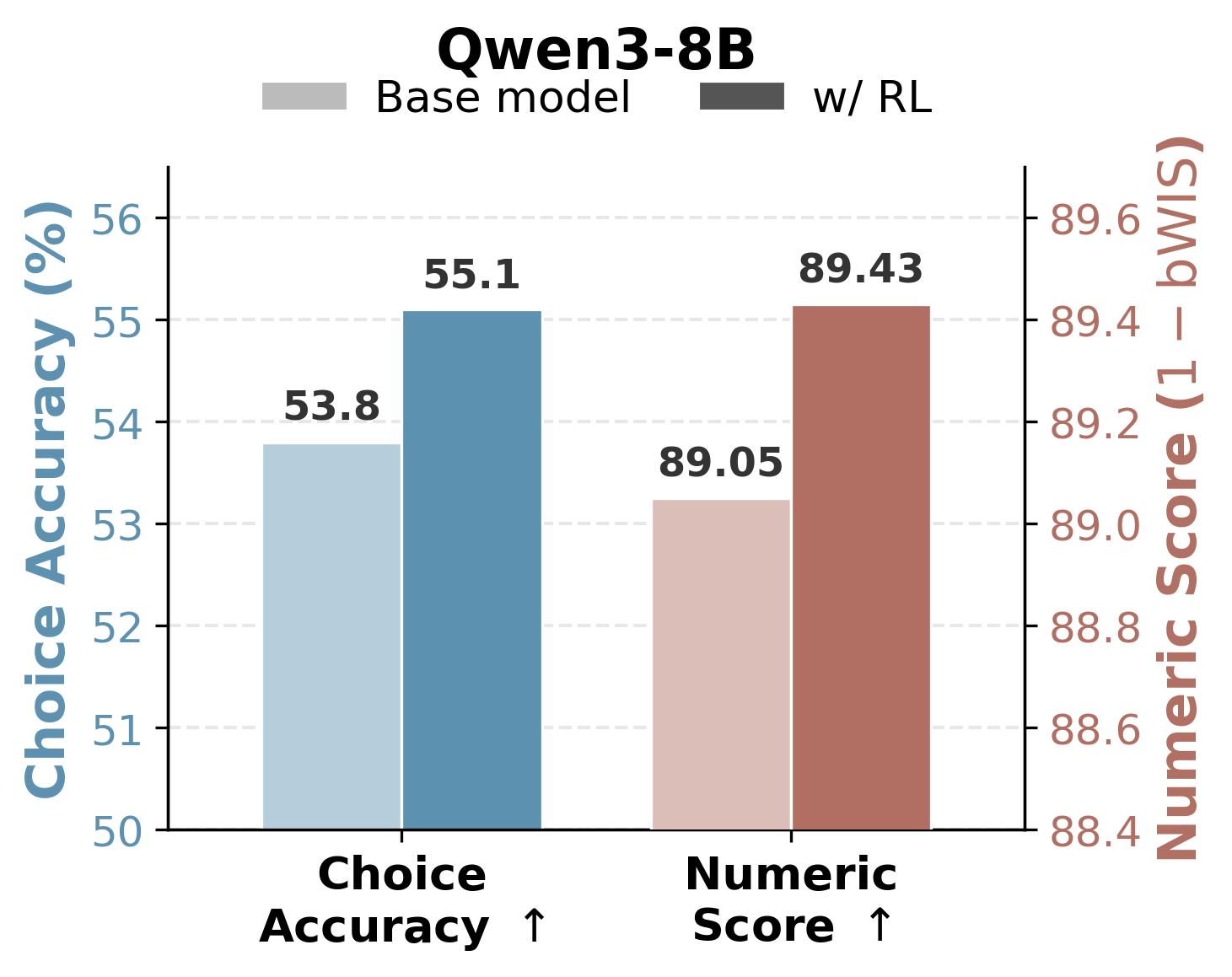}
        \caption{Qwen3-8B.}
        \label{fig:rl_results_8b}
    \end{subfigure}
    \caption{\textbf{RLVR sharpens both forecasting-quality metrics across LLMs.} For each LLM, we compare the base model against its RLVR-tuned counterpart on choice accuracy (left blue axis) and the numeric score (right red axis, $1-\mathrm{bWIS}$). Each metric shows two side-by-side bars, the base model and its RLVR-tuned counterpart, as absolute values. RLVR improves both metrics on both models.}
    \label{fig:rl_results}
\end{figure}

\begin{figure}[t]
    \centering
    \begin{subfigure}[t]{0.49\linewidth}
        \centering
        \includegraphics[width=\linewidth]{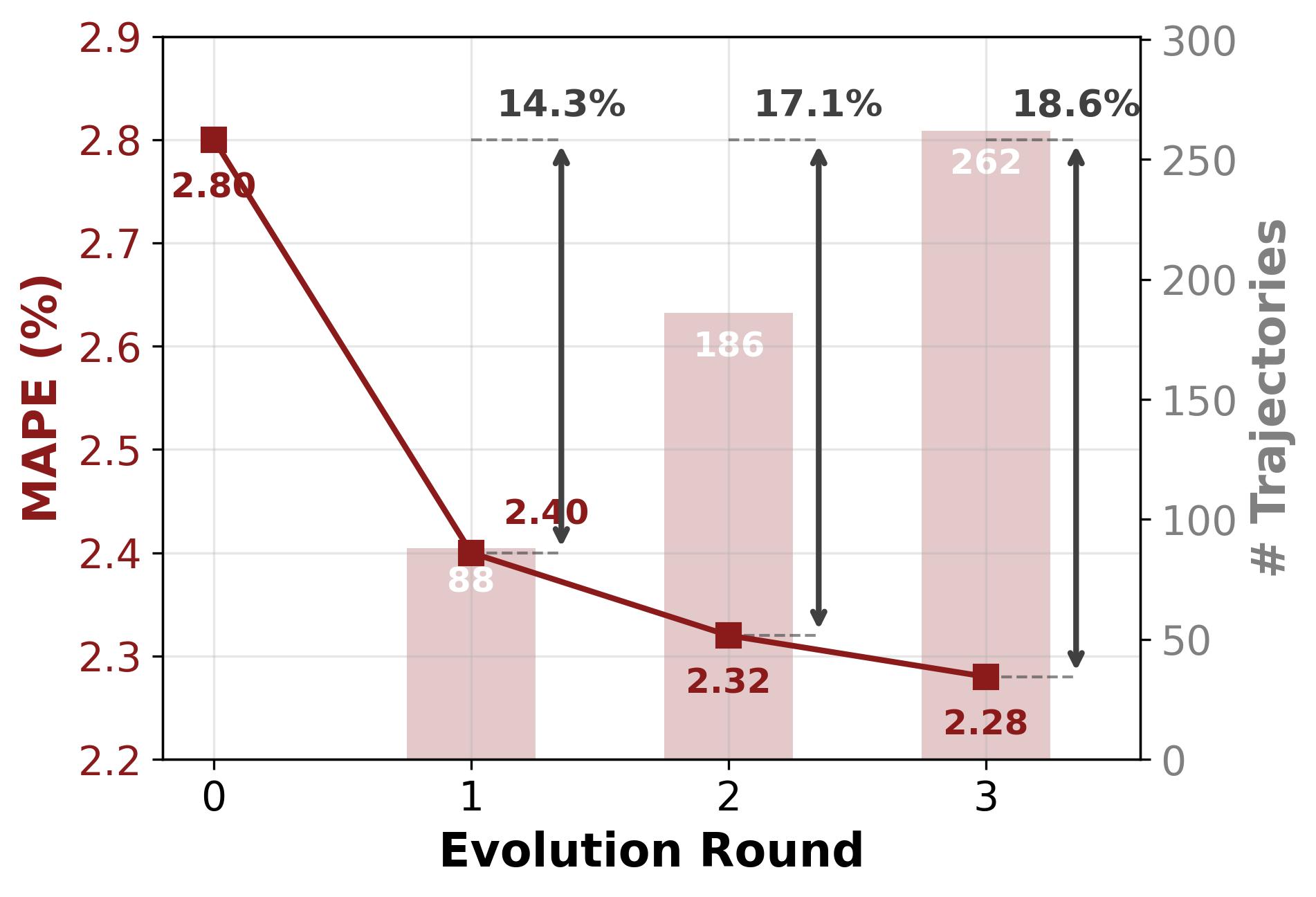}
        \caption{MAPE across evolution rounds.}
        \label{fig:skill_performance_mape}
    \end{subfigure}
    \hfill
    \begin{subfigure}[t]{0.49\linewidth}
        \centering
        \includegraphics[width=\linewidth]{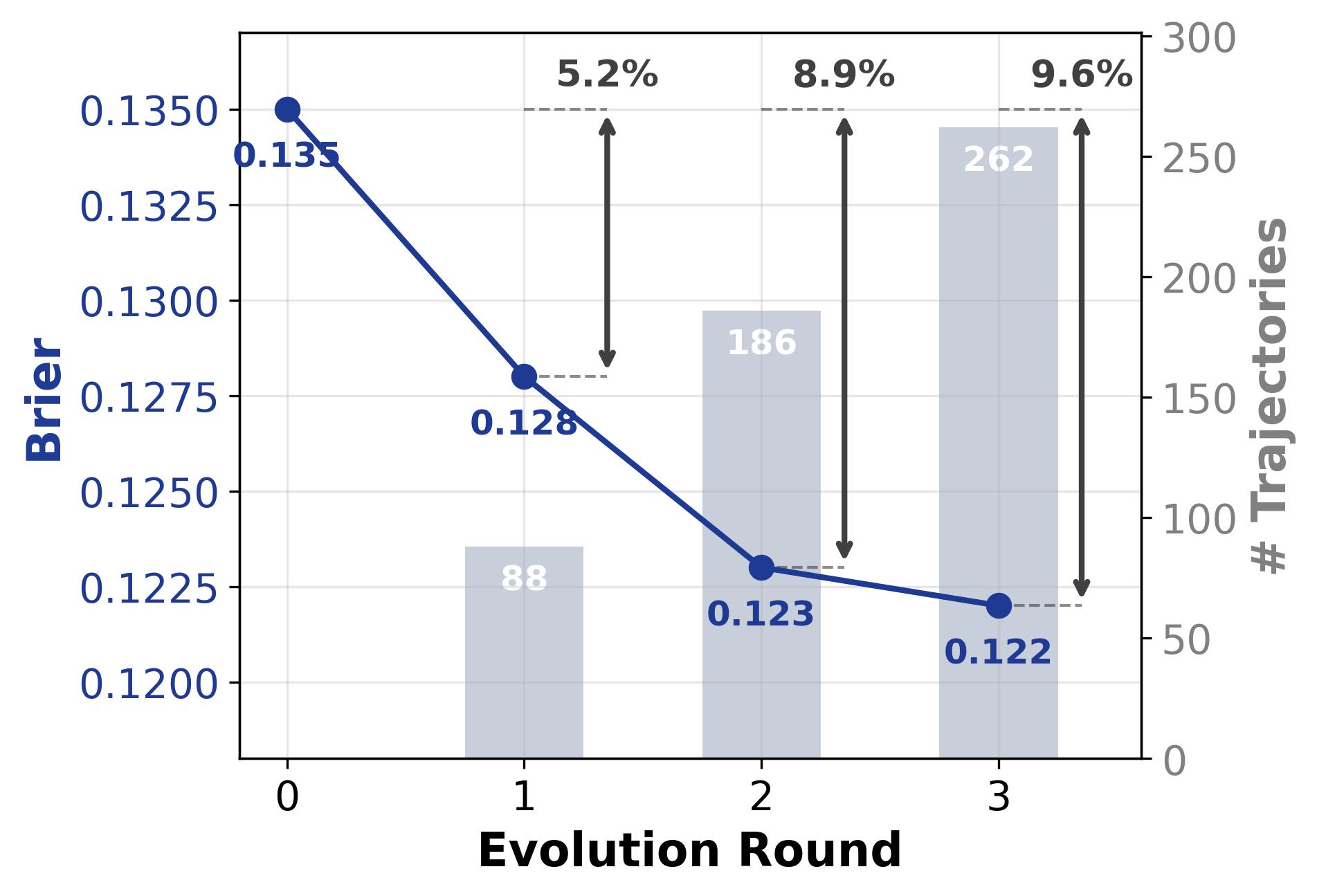}
        \caption{Brier score across evolution rounds.}
        \label{fig:skill_performance_brier}
    \end{subfigure}
    \caption{\textbf{Evolving skills improve forecasting quality continuously.} Distilled skills, inserted at inference without touching the weights, sharpen \textsc{Branchus}'s forecasts on both axes of forecasting quality. Because each round adds its live trajectories to the pool (nested pools of $88$, $186$, and $262$ trajectories), the skills \textsc{Athena} extracts over successive evolution rounds lift forecasting quality round by round rather than in a one-off gain. \textbf{Left}: numeric accuracy, measured by MAPE. \textbf{Right}: calibration, measured by the Brier score. \emph{For both metrics, lower is better.}}
    \label{fig:skill_performance}
\end{figure}

The first of the three axes is the model's parameters. The obstacle to internalizing forecasting experience is that the environment is live and easily contaminated: once an event resolves, its answer becomes retrievable from the web, so the task can no longer train a model to predict rather than look up. The clean window is short, yet RFT and RLVR must roll out and refine over the same tasks repeatedly, which leaves too little uncontaminated, verifiable experience to support training at all. The \emph{function-oriented} design of \textsc{Branchus} removes this obstacle by splitting the incompatible needs across teams organized by function: a time-sensitive search team gathers evidence while a question is still open, and a time-insensitive predict team reasons over it into a forecast. We cache the evidence and both teams' trajectories while the task is still fresh, so the many RFT/RLVR rounds that follow read from this frozen pool rather than from a web that has since revealed the answer. Because the predict team is trained only on fixed, pre-cached evidence, it cannot influence what it sees, which also closes the on-policy shortcut in which RLVR would manufacture answer-entailing evidence; any gain reflects better reasoning over the same information. Once enough cached trajectories resolve into verifiable labels, we train the predict team with two complementary paradigms: RLVR optimizes against each resolved outcome, and RFT fine-tunes on successful trajectories.

Under this function-oriented protocol, RLVR yields consistent gains across both LLMs (Figure~\ref{fig:rl_results}): choice accuracy rises by $1.3$ points on each model, while the numeric score ($1-\mathrm{bWIS}$) improves by $0.31$ and $0.38$. These clean, consistent gains from a reward computed against resolved outcomes are themselves evidence that the design has kept contamination out of the training signal, since a leaked answer would have collapsed the reward long before it could sharpen the model. The two kinds of metric respond at different rates, and we attribute this to the metrics themselves. Accuracy is binary and insensitive to how belief mass shifts, whereas calibration is continuous and reflects any move toward the true outcome probabilities. Calibration thus surfaces improvements too subtle for accuracy to register, acting as an early indicator that RLVR is tightening the model's probabilistic judgment even when the top-1 answer is unchanged.

RFT, under the same protocol, also improves every axis across two LLMs, again led by calibration (Figure~\ref{fig:rft_improvement}): accuracy rises by $0.7$ to $2.9$ points, while Brier and relative WIS improve consistently.
These gains compound over rounds rather than being spent in one pass: training GLM-4.7-Flash with RFT on each round's cached trajectories, the first round already sharpens forecasts (Brier $0.651\!\to\!0.636$, relative WIS $1.02\!\to\!0.99$), and the second round keeps pushing them down (Brier $0.630$, relative WIS $0.98$) as the task pool grows from $600$ to $1000$ (Figure~\ref{fig:param_rsi}). This is the multi-round self-improvement at work: because the environment is live, every round brings a fresh batch of newly resolved events, and the new experience distilled from them carries the forecasting quality a step further. The live stream of information is thus what sustains continual evolution, so the model never runs out of fresh, verifiable experience to learn from. Evolving the environment alongside the model amplifies this effect, since tasks that keep targeting the agent's current weaknesses turn this incoming information into richer experience for the next round and drive still larger gains, which we discuss in Sec.~\ref{sec:co_evolution}.

\subsection{Evolving skills unlock recursive self-improvement from fleeting live trajectories}


\begin{figure*}[t]
    \centering
    \begin{subfigure}[t]{0.48\linewidth}
        \centering
        \includegraphics[width=\linewidth]{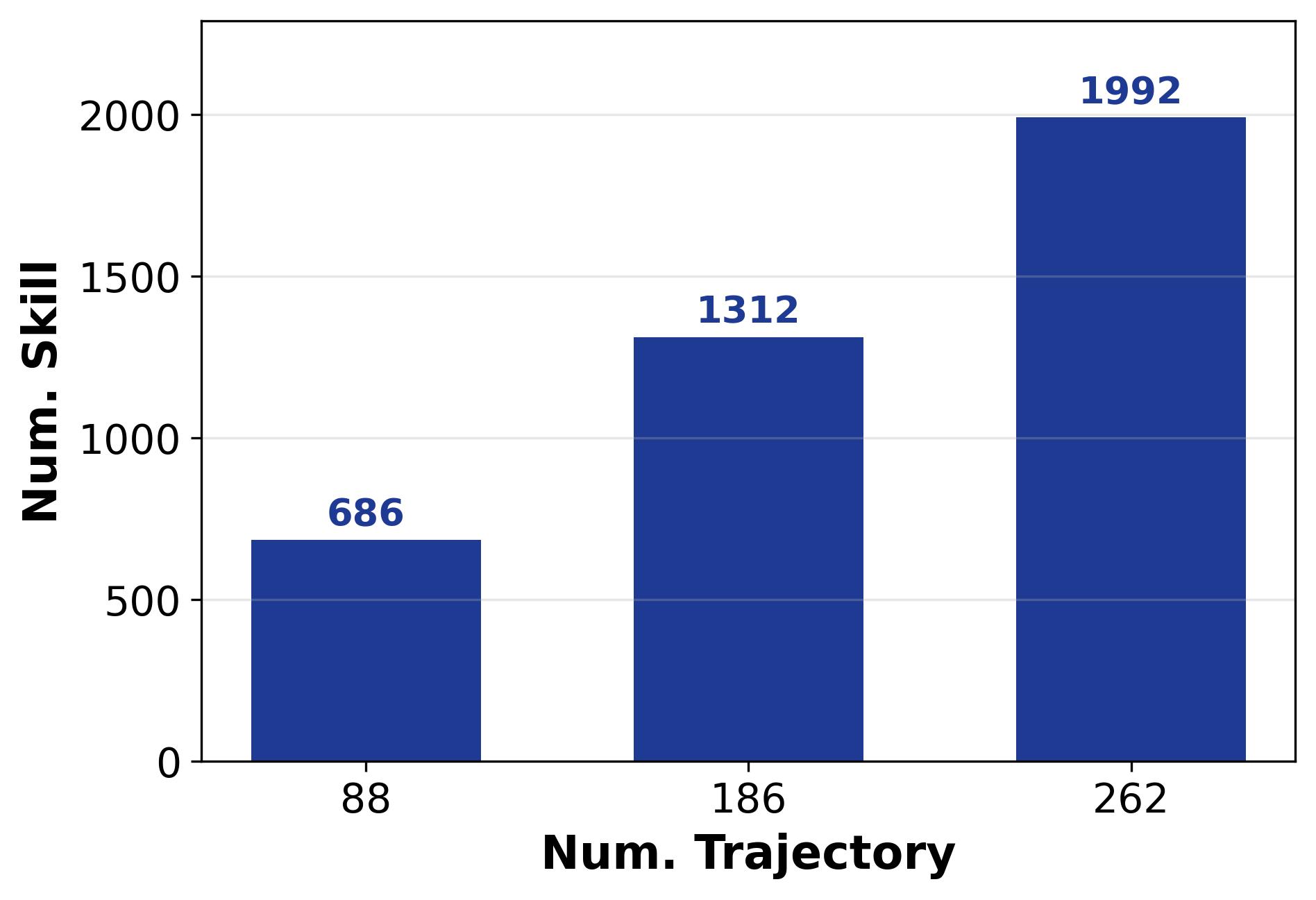}
        \caption{Number of distilled skills.}
        \label{fig:skill_count}
    \end{subfigure}
    \hfill
    \begin{subfigure}[t]{0.48\linewidth}
        \centering
        \includegraphics[width=\linewidth]{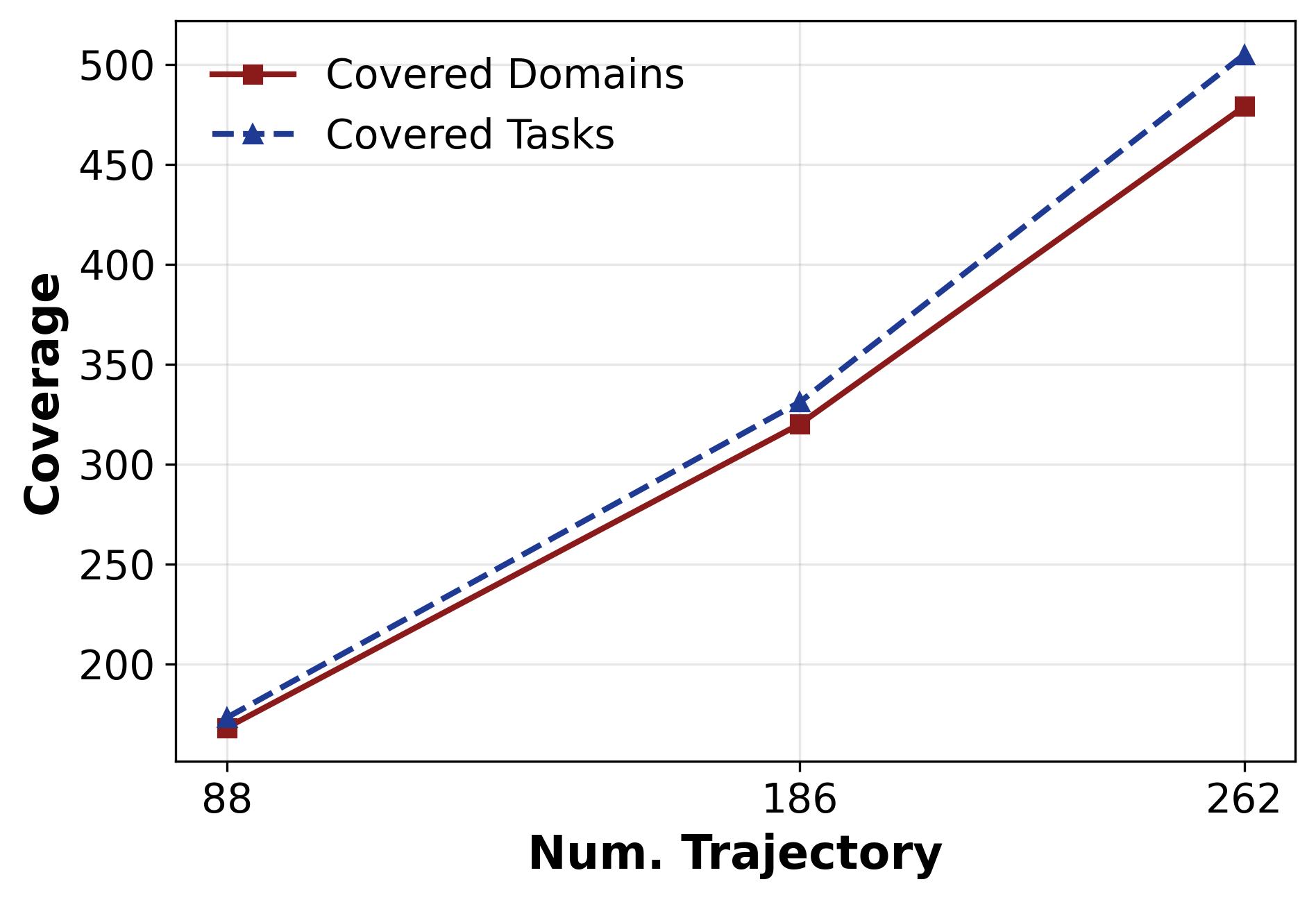}
        \caption{Topical coverage.}
        \label{fig:skill_coverage}
    \end{subfigure}
    \caption{\textbf{Distilled skills scale across evolution rounds.} As the collected trajectories grow round by round, \textsc{Athena} keeps distilling new skills from them, so the library expands rather than saturating, and later skills extend into domains seen only rarely before. \textbf{Left}: the number of distilled skills at each evolution round. \textbf{Right}: the breadth of topical coverage, measured by the number of the domains and concrete tasks that the skills cover.}
    \label{fig:skill_results}
\end{figure*}

The live trajectories that the function-oriented design cached for RLVR can instead be read for the lessons they contain, internalizing the same scarce experience in a lighter, more transparent form: reusable forecasting \emph{skills} that sit outside the model weights and are invoked at inference time. A single live trajectory thus feeds more than one lever, as RLVR turns it into a parameter update while skill distillation turns the same record into an explicit, inspectable procedure learned without any gradient update. The stakes here are sharper than for parameters, because fresh trajectories are collected only within each task's short open window, so the axis is useful only if a mere handful of them can already lift the system. We let \textsc{Athena} distill the lessons from \textsc{Branchus}'s trajectories into skill entries and build libraries over three evolution rounds, from nested pools of $88$, $186$ and $262$ live trajectories, each round extending the previous one as it accumulates more trajectories, giving \textsc{Branchus} the current library at inference so that any change in performance is attributable to the skills alone.

A handful of live trajectories already unlocks clear improvements on both axes of forecasting quality (Figure~\ref{fig:skill_performance}). Relative to the skill-free \textsc{Branchus} baseline, the Brier score drops from $0.135$ to $0.122$ ($0.013$ lower), and the same pattern reappears on the numeric side of the benchmark, where the MAPE falls from $2.80\%$ to $2.28\%$ ($0.52$ points lower).
Two features of the result are worth emphasising.
First, the skills are inserted without touching the model weights, so the gain is a direct measurement of the value of externalised live experience rather than an interaction with the underlying policy.
Second, the joint improvement in the Brier score and the numeric MAPE mirrors the pattern observed under RLVR, indicating that distilled skills and reinforcement learning act along the same axis of forecasting competence and can, in principle, be composed.

As each round accumulates more live trajectories into the growing pool, forecasting quality improves round by round, though at a slowing pace. Each pool adds gain on top of the last, with the Brier score reaching $0.007$, $0.012$ and $0.013$ below the baseline across the three rounds and the MAPE descending from $2.80\%$ to $2.40\%$, $2.32\%$ and $2.28\%$ (Figure~\ref{fig:skill_performance}), so the skills distilled at a later round help \textsc{Branchus} more than those from an earlier one. Later rounds move the aggregate metric less but are far from idle, for what they add is breadth: topical coverage broadens roughly threefold, from $168$ domains and $173$ tasks at round $1$ ($88$ trajectories) to $479$ domains and $505$ tasks at round $3$ ($262$ trajectories) (Figure~\ref{fig:skill_coverage}), as later skills reach domains that appeared only rarely before. The slowdown therefore reflects a shift in what each new live trajectory buys rather than saturation of the mechanism: the dominant forecasting patterns are captured early, so even a small pool internalizes the core modes of reasoning, after which added skills extend coverage into domains the evaluation set rarely samples and their benefit is diluted in the overall metric. Having now internalized the same trajectories into both parameters and skills, we next compare where each axis helps and at what cost, examining how the benefit of skills depends on task structure, before turning to the environment that supplies the experience.

\subsection{Skill evolution guides evidence gathering while parameter evolution sharpens evidence use}


\begin{figure}[t]
    \centering
    \begin{subfigure}[t]{0.48\linewidth}
        \centering
        \includegraphics[width=\linewidth]{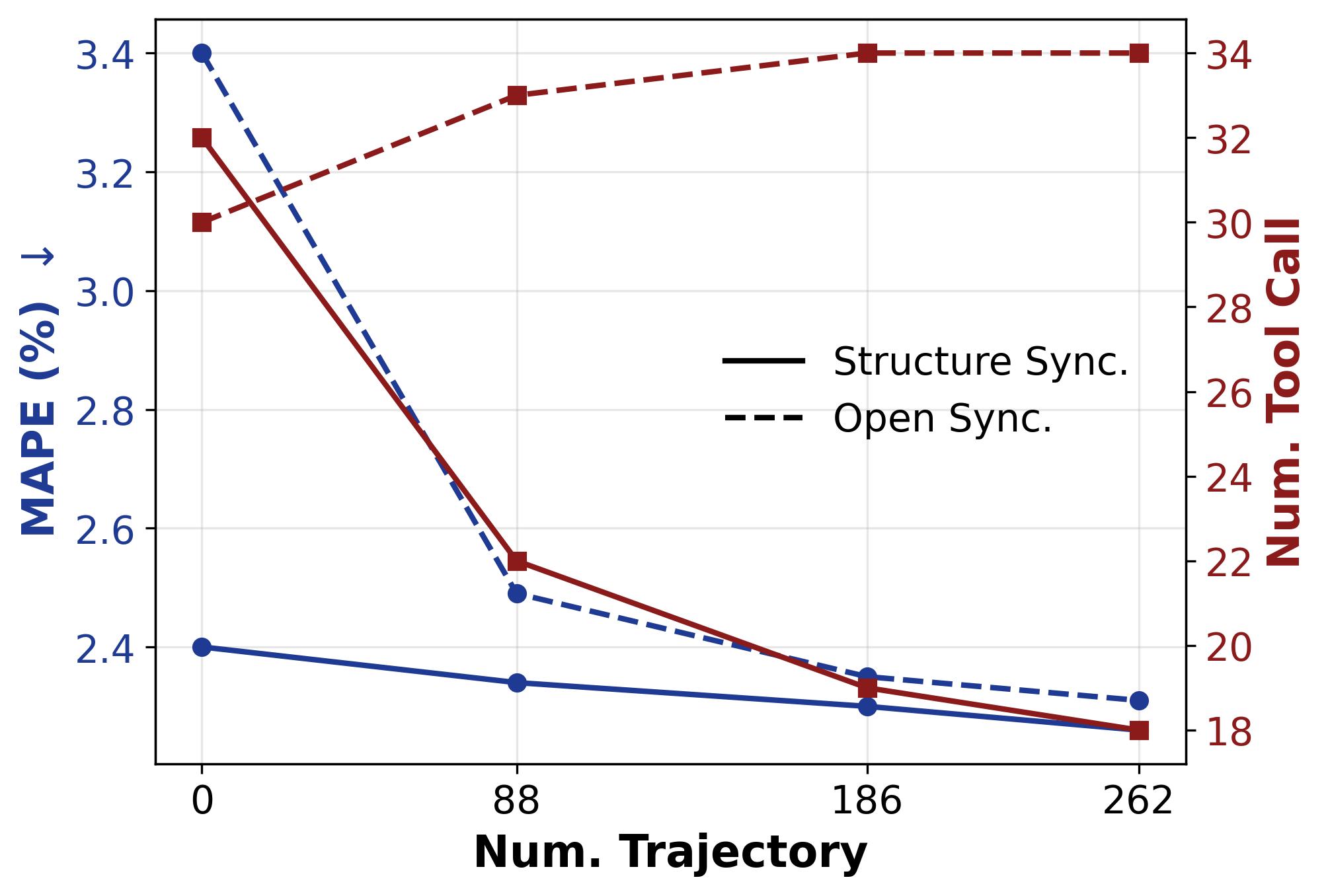}
        \caption{Where skills help, by task structure.}
        \label{fig:skill_efficiency_sub}
    \end{subfigure}
    \hfill
    \begin{subfigure}[t]{0.48\linewidth}
        \centering
        \includegraphics[width=\linewidth]{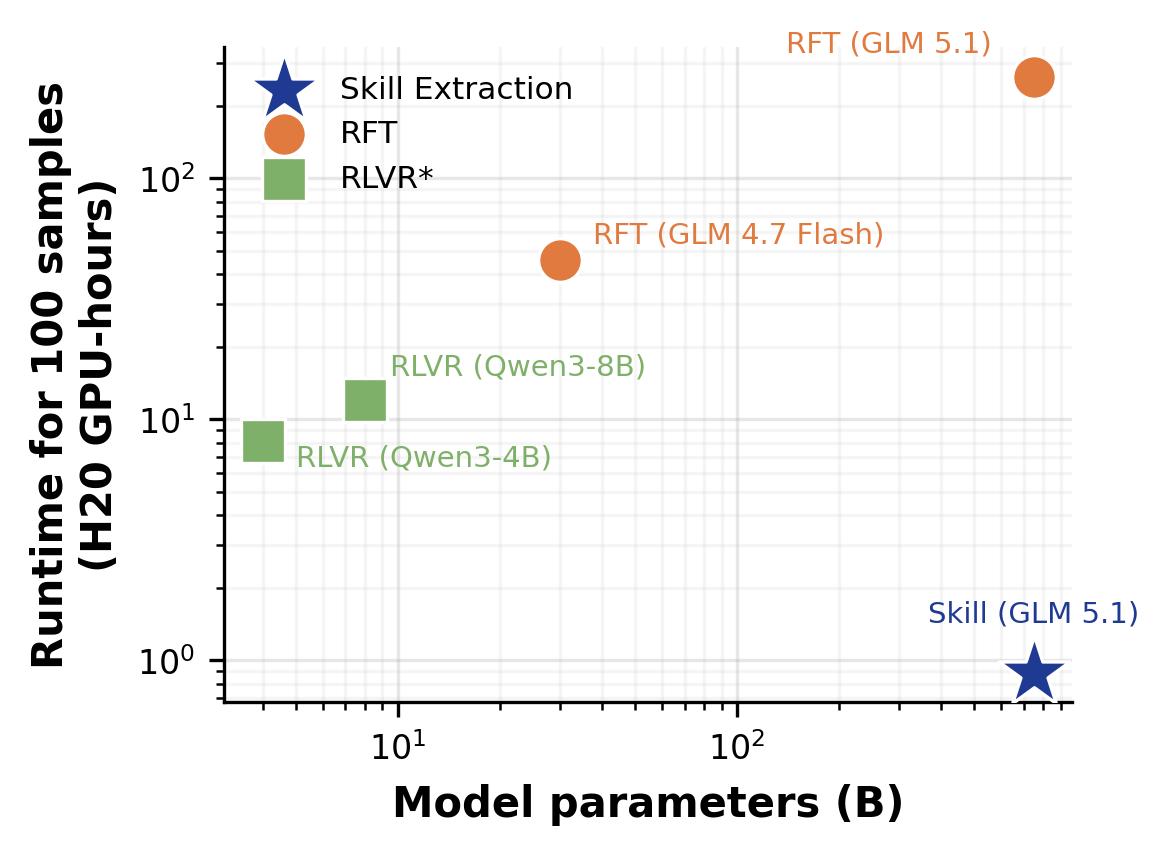}
        \caption{What skills cost, by model size.}
        \label{fig:cost_comparison}
    \end{subfigure}
    \caption{\textbf{Skills versus parameter updates: where they help and what they cost.} Externalizing experience into skills and internalizing it into parameters act at different stages and carry different costs, so the two axes are complementary rather than interchangeable. \textbf{Left}: forecasting accuracy and the agent's tool-call budget as the skill library grows, separated into structure synthesis problems that reuse recurring patterns and open synthesis problems from unfamiliar domains. \textbf{Right}: the compute cost of skill extraction against parameter updates as the base model scales up.}
    \label{fig:skill_vs_params}
\end{figure}

\begin{table}[t]
    \centering
    \small
    \caption{Token cost of distilling skills from one trajectory, by stage. Values are averages per trajectory; output tokens include reasoning tokens.}
    \begin{tabular}{llrrrr}
        \toprule
        \textbf{Stage} & \textbf{Step} & \textbf{Calls} & \textbf{Input} & \textbf{Output} & \textbf{Total} \\
        \midrule
        \multirow{3}*{Evaluation} & Step scoring (searcher, critic) & 6.49 & 100,781 & 8,476 & 109,257 \\
         & Scheduler scoring & 1.45 & 34,487 & 2,259 & 36,745 \\
         & Trajectory diagnosis & 1.05 & 52,396 & 1,098 & 53,494 \\
        \midrule
        \multirow{2}*{Extraction} & Experience extraction & 1.02 & 29,363 & 1,372 & 30,735 \\
         & Experience filtering & 0.91 & 29,520 & 508 & 30,029 \\
        \midrule
        \multirow{2}*{Generalization} & Four-level generalization & 2.02 & 18,409 & 2,354 & 20,763 \\
         & Generalization review & 2.02 & 19,813 & 2,299 & 22,112 \\
        \midrule
        \multicolumn{2}{l}{\textbf{Total}} & \textbf{14.96} & \textbf{284,769} & \textbf{18,366} & \textbf{303,135} \\
        \bottomrule
    \end{tabular}
    \label{tab:distill_cost}
\end{table}

The same forecasting experience can be turned into capability in two ways. Parameter evolution \emph{internalizes} it, folding the lessons of past trajectories into the model weights, whereas skill evolution \emph{externalizes} it, writing those lessons out as explicit, reusable procedures that sit beside the weights. The two forms act on different stages of forecasting: externalized skills shape how the agent \emph{explores} for evidence, while internalized parameters shape how well it \emph{uses} the evidence it has gathered, and a practical system must know when each is worth its cost. To expose where skills help, we partition the questions synthesized by \textsc{Prometheus} by how they were generated: \emph{structure synthesis problems}, scraped from fixed fields of a designated website and instantiated over a hand-designed template, so they share a stereotyped structure and a few recurring reasoning patterns; and \emph{open synthesis problems}, freely proposed by \textsc{Codex} from the day's news, so they are heterogeneous in topic, format, and required evidence. Alongside accuracy and calibration, we track the agent's \emph{tool-call budget}, which reveals whether skills make the agent do more work per question or the same work more directly.

Where skills help turns out to depend sharply on how the problem was synthesized (Figure~\ref{fig:skill_efficiency_sub}). On structure synthesis problems, which reuse a few recurring patterns, recursively evolved skills cut the agent's tool use: as the library grows, tool calls drop by \textbf{44\%} (from $32$ to $18$ per question) while MAPE stays at least as strong as the skill-free baseline. One encounter is often enough for \textsc{Athena} to distil a skill that captures the whole family, letting \textsc{Branchus} skip exploratory search and act on a compiled shortcut. On open synthesis problems drawn from domains the skills were never designed for, the signature reverses: the agent issues \textbf{13\%} more tool calls (from $30$ to $34$ per question) and investigates more broadly, and here the quality gains are the largest in the plot, with MAPE falling by \textbf{32\%} (from $3.40\%$ to $2.31\%$) and closing almost the entire gap to the accuracy on structure synthesis problems. On these novel problems a skill cannot supply the exact procedure, so it instead acts as a high-level plan that opens up broader lines of search than the agent would pursue unaided. Externalizing experience therefore improves the agent's \emph{exploration}: it makes evidence gathering either more direct, when a matching skill compiles the task away, or more far-reaching, when a loosely related skill widens the search over an unfamiliar domain.

Internalizing the same experience into parameters acts on a different stage and leaves this exploration behaviour untouched. Because \textsc{Athena} updates only the predict team, which reasons over cached evidence, and never the search team that issues tool calls, RLVR changes neither the number of searches nor what evidence is gathered; what it changes is how well that fixed evidence is \emph{used}, as the predict team learns to weigh and integrate the same observations into a sharper, better-calibrated forecast. The two forms also differ sharply in cost. Externalization is pure inference, with no rollout and no backward pass, a gap that only widens with model size (Figure~\ref{fig:cost_comparison} and Table~\ref{tab:distill_cost}). This cheapness, however, comes with a ceiling: a few trajectories, on the order of $88$, already yield most of the achievable gain, because externalization can only reorganise patterns some trajectory already exhibits and cannot discover strategies outside \textsc{Branchus}'s current repertoire. Internalization spends its larger budget precisely on behaviours the model does not yet exhibit, and keeps improving past the point where externalization plateaus.

Externalization and internalization are therefore complementary rather than substitutes: externalizing experience into skills gives rapid, cheap gains in exploration up to a well-defined ceiling, and internalizing it into parameters gives slower, costlier gains in evidence use that push beyond it. This suggests a simple division of labour for the recursive loop, in which a compute-efficient system externalizes experience into skills every round and reserves parameter updates for occasional longer bursts, recovering most of the capability of pure RLVR at a fraction of the compute and letting the same budget drive many more turns of the agent and environment loop. The compute saved this way is best spent on the third axis, the environment itself, since, as we show next, an environment that keeps posing new weakness-targeted tasks is what decides whether recursive improvement continues to compound or eventually flattens.

\subsection{An evolving environment continuously turns the agent's own failures into adversarial challenges}


The two agent-side axes share a common ceiling: whether experience is internalized into parameters or externalized into skills, both can only reorganise what the agent has already encountered, so each is bounded by the distribution of tasks the environment poses, and neither can lift this ceiling because neither controls the experience it feeds on. The third axis does. Rather than draw tasks from a fixed distribution, we let the \emph{environment} evolve in response to the agent, which is the engine of the living environment set out in the introduction: the agent's own forecasts expose its weaknesses, and those weaknesses drive the environment to pose harder tasks. We operationalise it by coupling \textsc{Athena} to the task synthesis of \textsc{Prometheus}: \textsc{Athena} analyses each batch of trajectories not for what worked but for what failed, distilling each recurring failure mode into a \emph{task-synthesis skill} that \textsc{Prometheus} uses to generate the next round, so that the environment continuously turns the agent's own failures into the adversarial challenges it must next overcome. The forecasting trajectory thus feeds its third and final lever, the same record reinforced into parameters and distilled into skills now mined for the failures from which a harder environment is built, closing the recursive loop in which all three axes evolve from one shared substrate.

The questions synthesized under the evolved skill are measurably harder for the current agent, the signature a weakness-targeting environment should produce. Evaluating the same \textsc{Branchus} checkpoint on questions from the original and the evolved skill, selection accuracy drops from $0.60$ to $0.53$ and the Brier score worsens from $0.52$ to $0.58$ (Figure~\ref{fig:skill_difficulty}). This added difficulty is not noise: if the evolved questions were mere coin flips, the agent's confidence would stop separating correct from incorrect answers, and AUROC would collapse toward $0.5$. Instead AUROC is essentially unchanged, edging up from $0.678$ to $0.68$, so the agent's uncertainty remains informative. By mining the agent's own trajectories for its failure modes, the environment manufactures genuinely harder yet answerable questions, which is the raw material the next round of self-improvement needs.

A worked example makes concrete what the statistics summarise (Appendix~\ref{app:case_prometheus}). Running the synthesis pipeline twice over an identical corpus, once with the baseline generator and once after injecting a task-synthesis skill from $\mathcal{S}_{\mathrm{env}}$, isolates the skill's effect while the source material is held fixed. For a Women's 77\,kg weightlifting final, the baseline keeps the nearly-decided predicate ``who wins the gold'', whereas the skill-steered generator reframes the same event around a precise numeric quantity, the gold medallist's total in kilograms, with the 268\,kg standard as a salient threshold, giving a question that remains objective and resolvable from one official page yet is genuinely uncertain. The same sharpening recurs across domains, as a coarse air-quality category band becomes a request for the exact index value. These \emph{numerical-forecast} and \emph{threshold-judgment} categories are where \textsc{Branchus}'s error is highest, and the skill-steered batch concentrates its questions there. The environment-level skill thus holds each question's objectivity and reproducible resolvability fixed and moves only its \emph{difficulty} into the dimension where the agent is weakest, the mechanism behind the aggregate drop in accuracy and Brier at preserved AUROC reported above.

\subsection{A targeted and challenging environment drives sustained system evolution}
\label{sec:co_evolution}


\begin{figure}[t]
    \centering
    \begin{subfigure}[t]{0.49\linewidth}
        \centering
        \includegraphics[width=\linewidth]{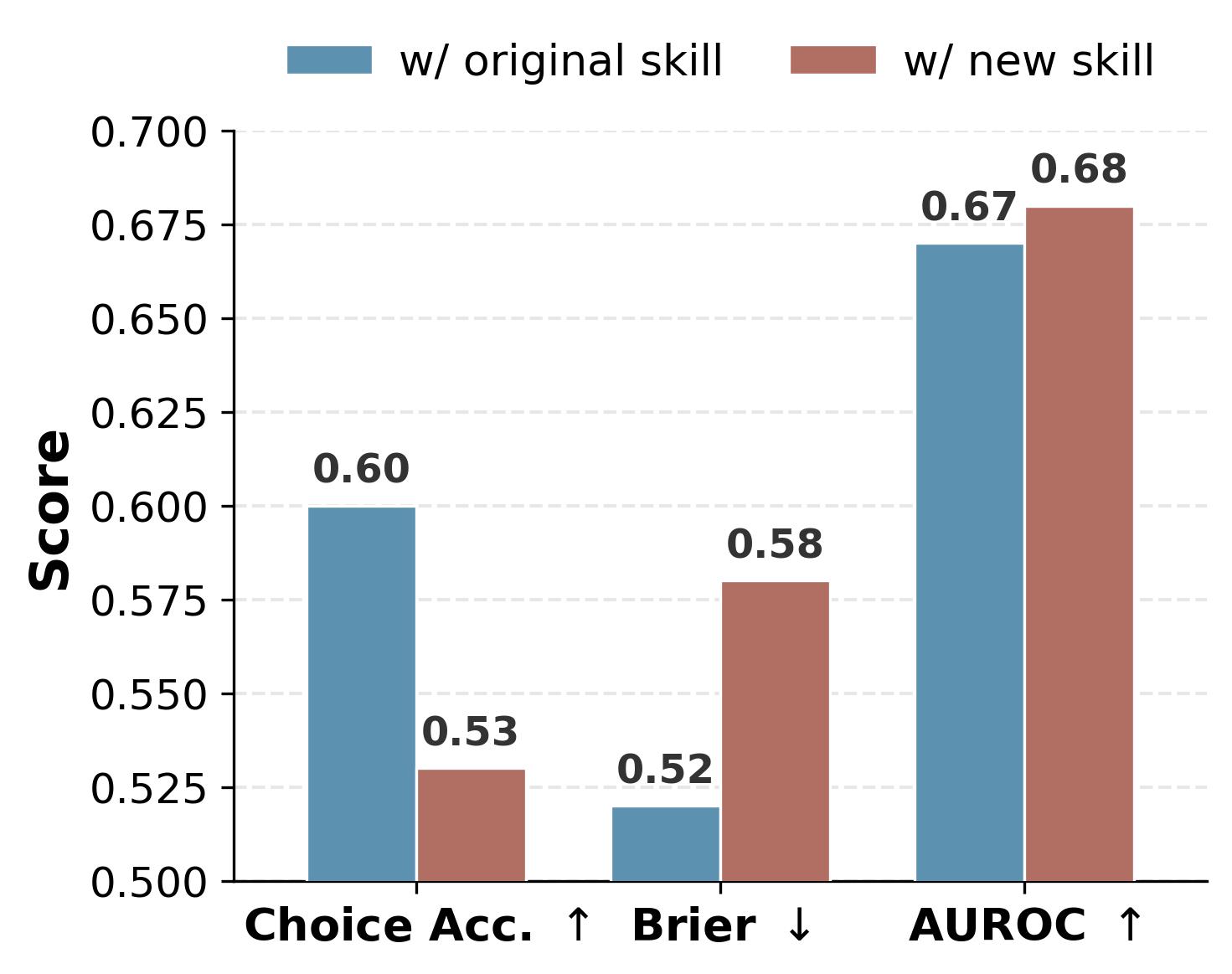}
        \caption{Difficulty of synthesized questions.}
        \label{fig:skill_difficulty}
    \end{subfigure}
    \hfill
    \begin{subfigure}[t]{0.49\linewidth}
        \centering
        \includegraphics[width=\linewidth]{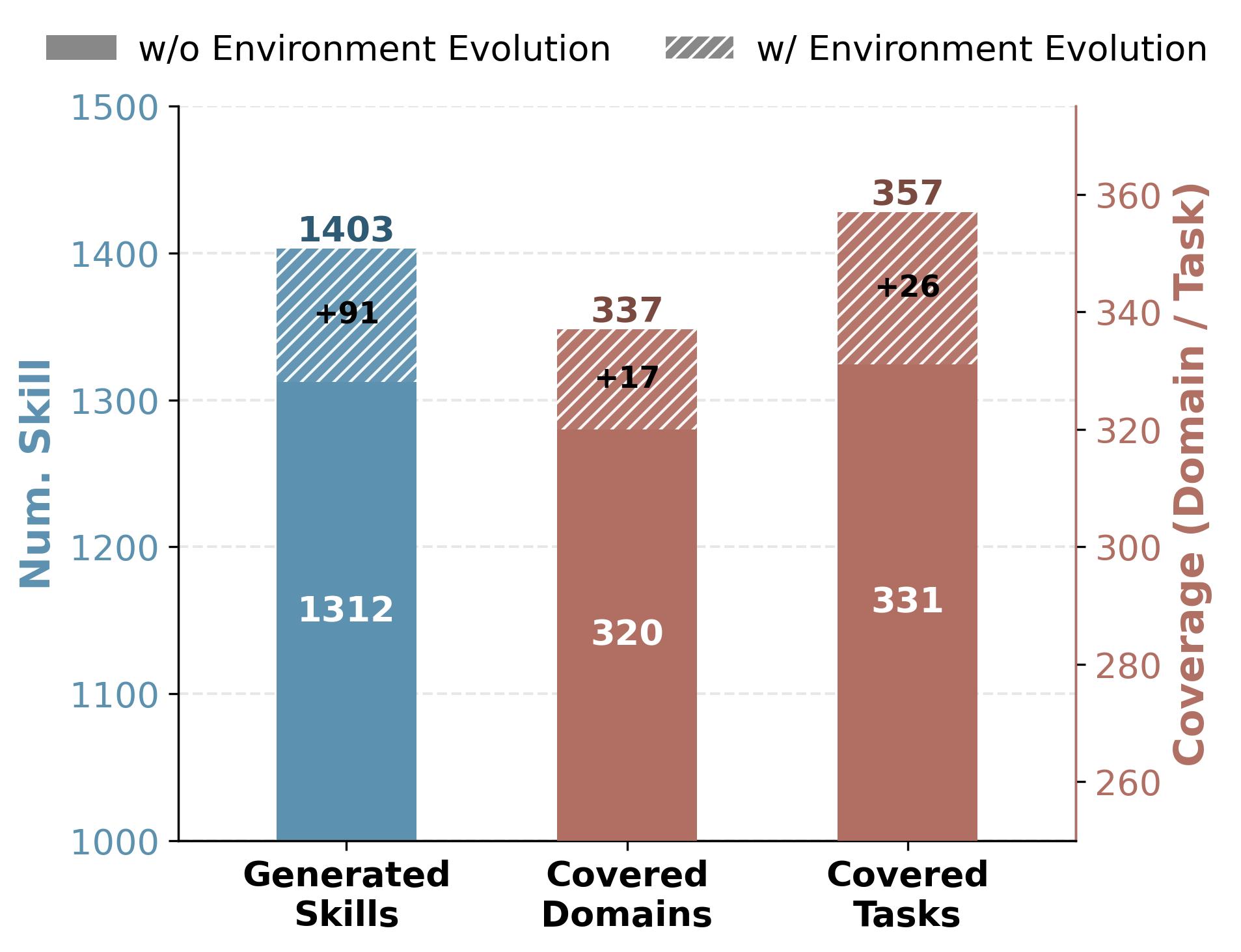}
        \caption{Breadth of distilled experience.}
        \label{fig:env_coverage}
    \end{subfigure}
    \caption{\textbf{Environment evolution targets genuine weaknesses and enriches the resulting experience.} Coupling task synthesis to the agent's own failure modes turns its weaknesses into harder challenges, and the richer trajectories these produce feed back into the system's evolution. \textbf{Left}: the difficulty of the synthesized questions under the original versus the evolved skill, showing that evolution raises difficulty while keeping each question well-posed and resolvable. \textbf{Right}: the distilled experience obtained from the same trajectory pool, comparing a weakness-targeting environment against a fixed one in the number of skills and the breadth of their coverage.}
    \label{fig:env_evolution}
\end{figure}

\begin{figure}[t]
    \centering
    \includegraphics[width=\linewidth]{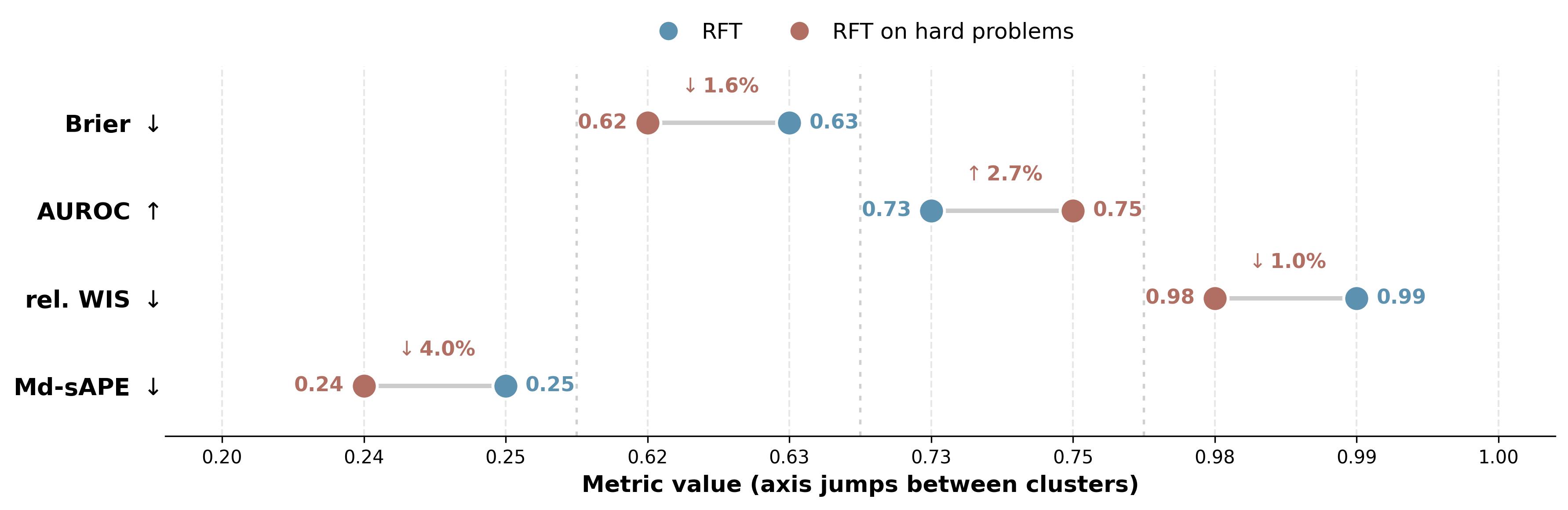}
    \caption{\textbf{Fine-tuning on the harder, evolved questions pays off.} Across every metric, RFT on the adversarially synthesized hard problems improves over plain RFT. Each row is drawn on its own scale.}
    \label{fig:hard_prob_improvement}
\end{figure}

The harder questions are valuable not in themselves but because the richer experience they generate flows back into both of \textsc{Branchus}'s evolution axes, so a weakness-targeting environment improves the system more than a fixed one does. For the skill axis, the evolved questions probe parts of the task space the agent has not mastered, and the experience distilled from them covers substantially more ground. From the same pool of $262$ trajectories, coupling task synthesis to the agent's weaknesses raises the number of distilled skills from $1312$ to $1403$ and broadens coverage from $320$ to $337$ domains and from $331$ to $357$ concrete tasks (Figure~\ref{fig:env_coverage}). The gain outlasts the current round, since a skill library spanning more domains is more likely to match the questions the agent meets next, making experience accumulated under an evolving environment more broadly reusable than experience accumulated under a static one.

The parameter axis benefits in the same direction. Relative to plain RFT, RFT on the adversarially synthesized hard problems improves every forecasting metric we track, lowering the Brier score by $1.6\%$, raising AUROC by $2.7\%$, and tightening numeric forecasts by $1.0\%$ in relative WIS and $4.0\%$ in Md-sAPE (Figure~\ref{fig:hard_prob_improvement}). Because both axes draw on the same enriched experience, the evolved environment lifts the system through the skills it distils and the parameters it updates alike.
The agent and its environment thus enter a productive cycle: each round of environment evolution exposes a weakness, and each round of agent evolution repairs it, so capability compounds rather than saturates. This is our clearest evidence that an evolving environment is an essential part of recursive self-improvement rather than an optional add-on, since a fixed distribution of tasks eventually exhausts what it can teach whereas one that keeps distilling challenges from the agent's own mistakes holds the loop open.

Taken together, the preceding subsections trace a single trajectory as it is turned, in turn, into updated parameters, distilled skills, and a harder environment. No one of these axes is sufficient on its own: parameters and skills internalize only the experience already available, and it is the evolving environment that keeps that experience expanding. It is their joint evolution, grounded in the information-rich trajectories that event forecasting naturally produces, that turns a strong forecaster into one whose capability compounds across rounds, exactly the form of recursive self-improvement set out in the introduction.

\section{Discussion}
\label{sec:discussion}

This study presents \textsc{EvoForesight}, a recursive self-improvement framework that treats real-world event forecasting as a living environment the agent learns inside, rather than a static task it is scored against.
Across a diverse suite of agent baselines spanning model families and domains, \textsc{EvoForesight} attains the strongest forecasting performance, matching or surpassing every alternative with clear gains in calibration (Section~\ref{sec:2.3_main_results}).
Beyond this static advantage, the system keeps improving from its own experience.
Skill evolution shows this most clearly: as each round adds its live trajectories to the pool, the skills distilled from the growing pool lift forecasting quality step by step rather than in a single jump.
These results show that an evolving agent inside an evolving environment can sustain improvement where prior approaches saturate, reframing progress from training a static model on a fixed dataset to cultivating a system whose capability accrues from its own experience.

The central contribution of \textsc{EvoForesight} is a real, dynamic, and verifiable environment that turns forecasting into a usable training signal.
This environment couples three modules: a living data source (\textsc{Prometheus}), an agentic forecaster (\textsc{Branchus}), and a recursive-evolution engine (\textsc{Athena}).
Such an environment is difficult to train on directly, because forecasting data expires once an event resolves, and a question answerable only before resolution loses its training signal afterward.
\textsc{EvoForesight} addresses this by separating evidence gathering from prediction: a time-sensitive search team caches the evidence available at forecast time, and a time-invariant predict team learns over that frozen context, so resolved outcomes supply verifiable rewards while the agent sees only the information available to it at forecast time.
Within this environment the agent retains experience through two complementary routes, compressing it into parameters by reinforcement learning on verifiable rewards and externalizing it into an explicit, reusable skill library distilled from forecasting trajectories.
Combining the two yields the fast, inspectable gains of skill accumulation together with the deeper gains that only parameter updates reach, a combination that surpasses any single-lever training.

These two routes, one internal and one external, strengthen different capabilities of the agents.
Parameter evolution writes experience into the weights: reinforcement learning trains the predict team on cached evidence, so the model draws fuller and better-calibrated conclusions from the information in hand.
Skill evolution writes experience into an explicit, reusable library that guides the search team, so the agent gathers evidence more accurately and comprehensively in the first place.
The two are complementary rather than redundant: one sharpens reasoning over fixed evidence, the other improves the evidence that reasoning receives.
Learning along both routes at once is itself a key technical enabler of recursive self-improvement, because they differ not only in what they strengthen but in what they demand of the experience that drives them (Section~\ref{sec:experiment}).
Internal learning needs a large volume of trajectories to evolve stably, since reinforcement learning must roll out and refine over the same tasks repeatedly, which makes it hard to apply in a live, contamination-prone environment whose clean, verifiable window is short.
External learning is far more stable, as a mere handful of live trajectories already lifts the system within each task's narrow open window, exactly the regime in which parameter updates run short of data.
Running both in tandem thus lets the system keep improving from scarce fresh experience, so the strongest configurations pair a parameter-trained predict team with a mature skill library, and the effective capability of the system is defined jointly by its weights and its accumulated skills.

Both weights and skills draw on one source, the experience the environment supplies, which makes the environment the governing term of the system, the one on which the other two depend.
A fixed environment fails this role in two ways.
First, it saturates: both routes exhaust the same task distribution, as the predict team resamples evidence it has already mastered and the skill library finds only patterns it has already distilled, so added compute relearns the same regularity and further agent-side training struggles to drive the curve higher.
Second, it is prone to contamination: once the outcomes of past events become publicly retrievable, the model can exploit a reasoning shortcut, recalling a resolved answer instead of reasoning toward a forecast.
To sustain improvement, the environment itself must therefore evolve, and evolve purposefully: it must stay fresh rather than reuse resolved events, and it must grow harder in a targeted way, posing new tasks at the current edge of the agent's competence.
The dynamic environment we construct meets both demands by drawing new events and scoring them against outcomes that resolve only in the future, which provides a signal that cannot have been seen in advance, to force the agent to genuinely forecast rather than recall and to keep the experience that drives recursive self-improvement renewing round after round.

This evolution is driven by the cooperation between \textsc{Athena} and \textsc{Prometheus}. \textsc{Athena} distills the failures it finds in the agent's trajectories into task-synthesis skills, and \textsc{Prometheus} uses them to generate the next round of tasks, so the environment acts as a teacher whose difficulty tracks the agent's current weaknesses: as the agent improves and its weaknesses shift, the tasks shift with them, and both sides keep changing.
Two boundaries constrain this teacher.
First, it is only as discerning as its diagnoses: a wrong prediction is an explicit failure that can be turned into a harder task, whereas a silent weakness such as overconfidence stays hidden from this signal, so the environment improves the agent most where its errors are already visible.
Second, its pace is set by the world, because a round resolves only after the events it concerns have occurred, so questions that take months to settle make each round of improvement slow and costly.

Several limitations of the present system point to future directions, each bearing on how far recursive self-improvement can run.
The first concerns the experience the environment supplies, its fuel: where resolvable, informative events are rare or strongly culture-dependent, trajectories stay thin, so broadening the sources of experience in both quantity and quality is a natural way to raise the ceiling of the recursive loop. The second concerns the skill library as the world drifts, since a skill distilled under one regime may apply differently once the underlying regularity shifts; forgetting, time-weighting, and failure detection would keep the externalized experience reliable, so that improvement accumulates genuine capability. The third concerns the evolution operators themselves: richer forms of parameter evolution that let updates compound round by round as skill accumulation already does, and more expressive forms of skill evolution in how experience is distilled, organized, and retrieved, both offer room to turn the same trajectories into larger gains. Pursuing these directions together is a natural way to strengthen the recursive loop.

The lasting product of this work is a methodology, a technical route for exploring recursive self-improvement, portable to any domain that is grounded, verifiable, and non-stationary. Carrying it elsewhere asks only for the three terms forecasting supplies here: a living source of tasks whose outcomes resolve over time, an agent that records its trajectories, and an evolution engine that turns those trajectories into updated weights, reusable skills, and harder tasks.
Scientific discovery, software engineering, and strategic planning each meet this description. In every such domain the environment commands the process, as an agent climbs only as far as its environment raises the bar, so building environments that keep evolving is the central problem recursive self-improvement must solve.
In place of ever larger static models on ever larger static datasets, the alternative is to cultivate evolving agents within evolving environments, setting the ceiling on progress by the openness of the environment rather than the size of a snapshot. \textsc{EvoForesight} is an early demonstration that this path can be walked, moving the measure of intelligence from what a system can do at a single moment to whether it can grow stronger through its own experience, observably and reproducibly in a real and verifiable world.
\section{Approach}
\label{sec:approach}

\subsection{Overview}
\textsc{EvoForesight} is a closed loop in which an agent and its environment improve one another over time, realized by three modules (Figure~\ref{fig:overview}), rather than a feed-forward pipeline from input to prediction.
\textsc{Prometheus} (Section~\ref{sec:prometheus}), the living data environment, turns real-world news into forecasting tasks whose ground truth is revealed only once the events resolve, supplying both the tasks and an objective, time-delivered reward.
\textsc{Branchus} (Section~\ref{sec:branchus}), the agentic forecaster, gathers evidence with external tools and reasons over it to issue a probabilistic prediction, producing both an answer and a full \emph{trajectory}.
\textsc{Athena} (Section~\ref{sec:athena}), the recursive-evolution engine, converts these trajectories back into capability: it updates \textsc{Branchus}'s parameters, distills its reasoning into reusable skills, and analyzes its failures into task-synthesis skills that make \textsc{Prometheus} pose harder, weakness-targeted questions.
The three thus form a single evolving system, in which a stronger \textsc{Branchus} meets a steadily more demanding \textsc{Prometheus}, while the world remains the external ground of every reward.

\paratitle{Notation and Problem Setup.}
We make this loop precise before instantiating its three modules.
\textsc{Branchus} is a policy $\pi_\theta$ with parameters $\theta$ that, given a forecasting question $q$ and a repository of forecasting skills $\mathcal{S}$, produces a probabilistic prediction together with the reasoning trajectory that supports it,
\begin{equation}
\label{eq:branchus}
(\hat{p},\,\tau) \;\sim\; \pi_\theta(\,\cdot \mid q,\,\mathcal{S}\,),
\end{equation}
where $\hat{p}$ is a distribution over the possible outcomes of $q$ and $\tau$ records the evidence gathered, the hypotheses entertained, and the intermediate decisions taken.
\textsc{Prometheus} is the environment $\mathcal{E}$: at round $t$ it emits a batch of questions $\mathcal{Q}^{(t)}$, and once the underlying events resolve it reveals, for each $q$, an objective outcome $y_q$ that defines a verifiable reward
\begin{equation}
\label{eq:reward}
r(q,\hat{p}) \;=\; \mathbf{1}\!\left[\,\arg\max_{o}\hat{p}(o)=y_q\,\right] \in \{0,1\},
\end{equation}
which checks whether the outcome the model deems most likely matches the resolved outcome $y_q$, so that time rather than human annotation supplies supervision.
We summarize the state of the system at round $t$ by the triple
\begin{equation}
\label{eq:state}
\Sigma^{(t)} \;=\; \big(\theta^{(t)},\,\mathcal{S}^{(t)},\,\mathcal{E}^{(t)}\big),
\end{equation}
whose three components are the agent's parameters, its accumulated skill library, and the environment that supplies its tasks.
\textsc{Athena} advances this state through three evolution operators, each learning from the trajectory pool $\mathcal{T}^{(t)}=\{\tau\}$ collected over $\mathcal{Q}^{(t)}$: a parameter update $\Phi_{\mathrm{rl}}$, a skill update $\Phi_{\mathrm{skill}}$, and an environment update $\Phi_{\mathrm{env}}$ (Section~\ref{sec:athena}).
Writing the rollout of $\pi_{\theta^{(t)}}$ on $\mathcal{Q}^{(t)}$ as $\mathcal{T}^{(t)}$, one round of recursive self-improvement is the update
\begin{equation}
\label{eq:rsi}
\theta^{(t+1)} = \Phi_{\mathrm{rl}}\!\big(\theta^{(t)},\mathcal{T}^{(t)}\big),\quad
\mathcal{S}^{(t+1)} = \Phi_{\mathrm{skill}}\!\big(\mathcal{S}^{(t)},\mathcal{T}^{(t)}\big),\quad
\mathcal{E}^{(t+1)} = \Phi_{\mathrm{env}}\!\big(\mathcal{E}^{(t)},\mathcal{T}^{(t)}\big),
\end{equation}
which maps $\Sigma^{(t)}\!\mapsto\!\Sigma^{(t+1)}$ and makes the products of one round the raw materials of the next (Algorithm~\ref{alg:EvoForesight}).
The word \emph{recursive} refers to this composition: the triple co-evolves rather than converging to a fixed set of weights, because $\Phi_{\mathrm{env}}$ keeps generating a next batch $\mathcal{Q}^{(t+1)}$ matched to the capability of the updated agent $\pi_{\theta^{(t+1)}}$.

\begin{algorithm}[t]
\caption{\textsc{EvoForesight}: one evolution round}
\label{alg:EvoForesight}
\begin{algorithmic}[1]
\Require state $\Sigma^{(t)}=(\theta^{(t)},\mathcal{S}^{(t)},\mathcal{E}^{(t)})$, forecast horizon $H$
\Ensure updated state $\Sigma^{(t+1)}$
\algsection{Task synthesis (\textsc{Prometheus})}
\State $\mathcal{D}^{(t)} \gets \textsc{Crawl}(\mathcal{E}^{(t)})$ \algcommentgray{crawl web pages: calendars, feeds, developing news}
\State $\tilde{\mathcal{Q}}^{(t)} \gets \textsc{Gen}(\mathcal{D}^{(t)},\mathcal{S}_{\mathrm{env}}^{(t)})$ \algcommentgray{synthesize question drafts $q=(e,o,f,t_{\mathrm{cut}},t_{\mathrm{dl}})$}
\State $\mathcal{Q}^{(t)} \gets \textsc{Select}\circ\textsc{Filter}(\tilde{\mathcal{Q}}^{(t)})$ \algcommentgray{temporal/dedup/conflict filtering, diversity-aware selection}
\algsection{Rollout (\textsc{Branchus})}
\State $\mathcal{T}^{(t)} \gets \{\, (\hat{p},\tau)\sim\pi_{\theta^{(t)}}(\cdot\mid q,\mathcal{S}^{(t)}) : q\in\mathcal{Q}^{(t)} \,\}$
\State observe outcomes $y_q$ and rewards $r(q,\hat{p})$ as events resolve within $H$
\algsection{Evolution (\textsc{Athena})}
\State $\theta^{(t+1)} \gets \Phi_{\mathrm{rl}}(\theta^{(t)},\mathcal{T}^{(t)})$ \algcommentgray{parameter level: RLVR}
\State $\mathcal{S}^{(t+1)} \gets \Phi_{\mathrm{skill}}(\mathcal{S}^{(t)},\mathcal{T}^{(t)})$ \algcommentgray{behavior level: skill distillation}
\State $\mathcal{E}^{(t+1)} \gets \Phi_{\mathrm{env}}(\mathcal{E}^{(t)},\mathcal{T}^{(t)})$ \algcommentgray{environment level: adversarial synthesis}
\State \Return $\Sigma^{(t+1)}=(\theta^{(t+1)},\mathcal{S}^{(t+1)},\mathcal{E}^{(t+1)})$
\end{algorithmic}
\end{algorithm}

\subsection{Prometheus: A Living Data Environment}
\label{sec:prometheus}

\textsc{Prometheus} constructs batches of objective, short-horizon prediction questions through two complementary mechanisms. In \emph{template-based generation}, source-specific adapters retrieve public calendars, schedules, feeds, and structured pages, identify upcoming events, and instantiate predefined question templates. In \emph{agent-driven open discovery}, \textsc{Codex}~\citep{openai2026codex} gathers information from relevant websites, interprets scheduled events and developing stories, and constructs structured question drafts. Its sources span many domains, including sports, economics, monetary policy, energy, weather, public health, technology, government affairs, and cryptocurrency. Candidates from both routes enter a shared construction and publication pipeline, which we formalize next.
Formally, each candidate question is a tuple
\begin{equation}
\label{eq:question}
q \;=\; \big(e,\,o,\,f,\,t_{\mathrm{cut}},\,t_{\mathrm{dl}}\big),
\end{equation}
pairing an identifier $e$ for the underlying real-world event with an identifier $o$ for the specific outcome being asked about (such as the winner of a match or its total score), an answer format $f$, a generation time $t_{\mathrm{cut}}$ that marks when the question is created, and a forecast deadline $t_{\mathrm{dl}}$.
The pipeline proceeds in three stages. It first crawls the web pages exposed by the current environment, $\mathcal{D}^{(t)} = \textsc{Crawl}(\mathcal{E}^{(t)})$, yielding a corpus $\mathcal{D}^{(t)}$ of calendars, feeds, and developing news from which questions can be posed. A generator then turns this corpus into a set of candidate drafts, $\tilde{\mathcal{Q}}^{(t)} = \textsc{Gen}(\mathcal{D}^{(t)},\mathcal{S}_{\mathrm{env}}^{(t)})$, where each draft is a tuple of the form in Equation~\eqref{eq:question} and $\mathcal{S}_{\mathrm{env}}^{(t)}$ is the library of task-synthesis skills that \textsc{Athena} feeds back to steer generation toward \textsc{Branchus}'s weaknesses (Section~\ref{sec:athena}). A filter-and-select map finally yields the published batch, so that the pipeline is the composition
\begin{equation}
\label{eq:prometheus-pipeline}
\mathcal{Q}^{(t)} = \textsc{Select}\circ\textsc{Filter}\big(\textsc{Gen}(\mathcal{D}^{(t)},\mathcal{S}_{\mathrm{env}}^{(t)})\big),
\end{equation}
whose stages we now describe.

\paratitle{Validation and Temporal Filtering.}
The pipeline first identifies events whose outcomes are expected to become observable within the requested time horizon $H$, measured from the start of question generation. Once the generation time $t_{\mathrm{cut}}$ is recorded, each candidate is reassessed for temporal eligibility and is retained only if its deadline and expected resolution time $t_{\mathrm{res}}$ satisfy
\begin{equation}
\label{eq:temporal}
t_{\mathrm{cut}} \;<\; t_{\mathrm{dl}} \qquad\text{and}\qquad t_{\mathrm{cut}} \;<\; t_{\mathrm{res}}(q) \;\le\; t_{\mathrm{cut}} + H,
\end{equation}
so that the answer remains open at generation time and resolves within the requested horizon.

\paratitle{Deduplication and Conflict Detection.}
The remaining drafts undergo duplicate removal and conflict detection. Each
draft receives a reproducible identifier $\mathrm{id}(q)$ derived from the underlying event $e$,
the result $o$ being predicted, the answer format $f$, and the relevant time
boundaries. Drafts whose $\mathrm{id}(q)$ already appears in the question bank are
excluded. Within the current batch, drafts that share the same $(e,o)$ are compared: when their structured attributes agree,
one representative is retained despite differences in wording; when they
disagree, the group is discarded to keep the published versions
of a question consistent.

\paratitle{Diversity-Aware Selection.}
The remaining candidates are selected to promote diversity across events, domains, sources, and question types. The
selector first favors events not yet represented in the batch. It then
considers a second question for an already represented event, up to a maximum of two questions per event.
Among otherwise comparable candidates, it favors domains and sources with
fewer questions in the historical bank and current batch, followed by
question types and answer formats that occur less frequently in the current
batch, and then earlier forecast deadlines. This procedure promotes a diverse set of questions, though the final distribution remains constrained by the number of eligible questions
available from each domain and source.

\subsection{Branchus: Agentic Forecasting System}
\label{sec:branchus}

\textsc{Branchus} is the agentic forecaster that answers the questions posed by \textsc{Prometheus}: given a question, it gathers evidence from the outside world, reasons over it, and issues a probabilistic prediction, recording the full trajectory that leads to each answer.
Two mechanisms underpin its design. A \emph{multi-agent team} organization separates collecting evidence from turning evidence into a prediction, so each part can be scaled and improved independently. A \emph{skill management} mechanism equips the agents with an evolving repository of forecasting knowledge, from hand-authored seeds to knowledge distilled from past experience.
Together these make \textsc{Branchus} both a strong forecaster and a producer of the structured trajectories that \textsc{Athena} later learns from.

\paratitle{Multi-Agent Teams.}
\textsc{Branchus} factors the policy of Equation~\eqref{eq:branchus} into two cooperating teams, a \emph{search team} $\pi^{\mathrm{s}}_\theta$ and a \emph{predict team} $\pi^{\mathrm{p}}_\theta$, organized by whether their work depends on when it is done.
The \emph{search team} performs search: it collects and vets evidence relevant to a question, and consists of a \emph{searcher}, a \emph{scheduler}, and a \emph{critic}.
Its work is time-sensitive, because the available evidence changes as an event approaches and, once the event resolves, its answer becomes directly retrievable, so the same search yields a different result depending on when it is run.
The scheduler decomposes the question and dispatches sub-agents, the searcher retrieves information from external sources, and the critic assesses the quality, relevance, and reliability of what has been gathered. The model invokes these sub-agents on its own as needed, iterating between retrieval and judgement until the evidence is sufficient, and every step is retained as the search trajectory $\tau_{\mathrm{s}}\sim\pi^{\mathrm{s}}_\theta(\cdot\mid q,\mathcal{S})$.
The \emph{predict team} performs prediction, turning this evidence into a forecast.
Its work is time-invariant, because it reasons only over evidence already gathered: given the same evidence, it produces the same forecast regardless of when it runs.
Rather than re-searching, it calls a local tool to access the search team's cached reasoning trajectories, which lets it aggregate a set $\{\tau_{\mathrm{s}}^{(j)}\}_{j=1}^{m}$ of such trajectories into a single view of the evidence and reconcile the partial conclusions they contain.
From this aggregated evidence it produces the final probabilistic answer,
\begin{equation}
\label{eq:twoteam}
\hat{p} \;\sim\; \pi^{\mathrm{p}}_\theta\!\big(\,\cdot \mid q,\,\{\tau_{\mathrm{s}}^{(j)}\}_{j=1}^{m},\,\mathcal{S}\,\big),
\end{equation}
which recovers Equation~\eqref{eq:branchus} with $\tau=(\{\tau_{\mathrm{s}}^{(j)}\}_j,\tau_{\mathrm{p}})$.
This division of labour keeps evidence gathering and prediction loosely coupled: the search team explores broadly and in parallel, while the predict team concentrates the resulting trajectories into a calibrated judgement.

\paratitle{Skill Management.}
\textsc{Branchus} draws on a repository of forecasting \emph{skills} $\mathcal{S}$, compact pieces of procedural knowledge that guide how a question is approached. Skills are of two kinds.
\emph{Task-specific skills} are authored by hand and form a seed set $\mathcal{S}_0$: each encodes knowledge about a particular class of question, including important source websites to consult and prompts that steer the reasoning, giving the agent a useful starting competence before any experience is accumulated.
\emph{Experience-distilled skills}, by contrast, are consolidated from \textsc{Branchus}'s own trajectories by \textsc{Athena} and grow as the system evolves, so that $\mathcal{S}^{(t)}=\mathcal{S}_0\cup\bigcup_{t'<t}\Phi_{\mathrm{skill}}(\mathcal{S}^{(t')},\mathcal{T}^{(t')})$.
To make skills easy to organize, retrieve, and reuse, we structure each skill as a four-level tuple $s=(s_1,s_2,s_3,s_4)$ ordered from the abstract to the concrete, so that a general forecasting principle $s_1$ at the top specializes progressively down to a concrete procedure or source $s_4$ at the bottom.
At inference time, an agent assigned role $a$ to question $q$ retrieves a relevant subset and keeps, for each matched skill, the most specific level that still applies,
\begin{equation}
\label{eq:retrieve}
\textsc{Retrieve}(q,a,\mathcal{S}) \;=\; \big\{\, s_{\ell^\star(s)} : s\in\mathcal{S},\; \textsc{match}(s,q,a) \,\big\},\qquad
\ell^\star(s)=\max\{\ell : s_\ell \text{ applies to } q\}.
\end{equation}
This layered representation lets \textsc{Branchus} match a new question to skills at the appropriate level of generality, so that a question without an exact precedent still benefits from a broader level, and lets the repository hold both broad heuristics and highly specific recipes within a single consistent schema.

\subsection{Athena: Recursive Evolution System}
\label{sec:athena}

\textsc{Athena} is the recursive-evolution engine of \textsc{EvoForesight}: it extracts experience from the trajectory pool $\mathcal{T}^{(t)}$ that \textsc{Branchus} produces and turns it into improvement through the three operators of Equation~\eqref{eq:rsi}.
At the parameter level, $\Phi_{\mathrm{rl}}$ applies reinforcement learning to \textsc{Branchus}'s predictions to update its weights $\theta$. At the behaviour level, $\Phi_{\mathrm{skill}}$ distils reusable skills from its trajectories to strengthen the agent while leaving those weights untouched. At the environment level, $\Phi_{\mathrm{env}}$ synthesizes adversarial skills from its failures to make \textsc{Prometheus} generate harder, weakness-targeted tasks.
The first two operators evolve the agent and the third evolves the environment, so that both improve from the same stream of experience.

\paratitle{Parameter-Level Evolution: Reinforcement Learning on Branchus's Predictions.}
At the parameter level, $\Phi_{\mathrm{rl}}$ improves \textsc{Branchus} by reinforcement learning, using Group Relative Policy Optimization (GRPO)~\citep{deepseekmath} on questions synthesized by \textsc{Prometheus}.
Training proceeds in two stages that separate evidence gathering from parameter updates.
First, the search team runs over these questions and its reasoning trajectories $\{\tau_{\mathrm{s}}\}$ are cached; once enough have accumulated, we perform reinforcement learning with verifiable rewards (RLVR) on the predict team alone, using the cached evidence.
Training only the predict team on fixed, pre-collected trajectories rules out answer leakage, since the model can no longer shortcut the task by re-collecting evidence that trivially entails the answer; yet because the evidence rather than the answer is fixed, the predict team still explores freely, sampling $G$ candidate predictions $\{\hat{p}_i\}_{i=1}^{G}$ from the same evidence and updating on their outcomes.
Concretely, each sampled prediction receives the verifiable reward of Equation~\eqref{eq:reward}, $r_i=r(q,\hat{p}_i)\in\{0,1\}$, which is turned into a group-relative advantage by normalizing within the sampled group,
\begin{equation}
\label{eq:advantage}
A_i \;=\; \frac{r_i - \mathrm{mean}\big(\{r_j\}_{j=1}^{G}\big)}{\mathrm{std}\big(\{r_j\}_{j=1}^{G}\big)},
\end{equation}
and the predict team is updated by maximizing the clipped GRPO objective over the fixed evidence,
\begin{equation}
\label{eq:grpo}
\Phi_{\mathrm{rl}}:\;\;
\theta^{(t+1)} = \arg\max_{\theta}\;
\mathbb{E}_{q\sim\mathcal{Q}^{(t)},\,\{\hat{p}_i\}\sim\pi^{\mathrm{p}}_{\theta^{(t)}}}
\!\left[\frac{1}{G}\sum_{i=1}^{G}\min\!\Big(w_i(\theta)\,A_i,\;\mathrm{clip}\big(w_i(\theta),1\!-\!\epsilon,1\!+\!\epsilon\big)A_i\Big)\right],
\end{equation}
where $w_i(\theta)=\pi^{\mathrm{p}}_\theta(\hat{p}_i\mid q,\{\tau_{\mathrm{s}}^{(j)}\},\mathcal{S})/\pi^{\mathrm{p}}_{\theta^{(t)}}(\hat{p}_i\mid\cdot)$ is the importance ratio and $\epsilon$ the clipping range.
The resolved ground truth $y_q$ from \textsc{Prometheus} supplies the reward signal in Equation~\eqref{eq:reward} without human annotation. This regime motivated the multi-agent design: by isolating prediction from search, the two-team organization makes it possible to reinforce the predict team on cached evidence, so that the measured improvement reflects better reasoning over the same information rather than opportunistic re-collection.

\paratitle{Behavior-Level Evolution: Distilling Reusable Skills from Branchus's Trajectories.}
At the behavior level, $\Phi_{\mathrm{skill}}$ turns \textsc{Branchus}'s reasoning trajectories into reusable skills.
These skills guide the agent at inference time and improve how it works while leaving its weights fixed.
Formally, $\Phi_{\mathrm{skill}}$ maps the trajectory pool to a set of new skills, $\Phi_{\mathrm{skill}}(\mathcal{S}^{(t)},\mathcal{T}^{(t)})=\mathcal{S}^{(t)}\cup\Delta\mathcal{S}^{(t)}$, through the composition $\Delta\mathcal{S}^{(t)}=(\textsc{Organize}\circ\textsc{Filter}\circ\textsc{Extract}\circ\textsc{Eval})(\mathcal{T}^{(t)})$.
In the first stage, a judge model $\textsc{Eval}$ scores the scheduler, searcher, and critic of the search team against role-specific rubrics.
The resolved outcome $y_q$ is withheld from the judge, because forecast correctness is an imperfect proxy for reasoning quality.
In the second stage, $\textsc{Extract}$ uses these scores to identify positive experiences from helpful but non-obvious actions, and negative experiences from recurring mistakes or failure modes.
Separating evaluation from extraction lets each step focus on a single judgment.
A $\textsc{Filter}$ then discards weak experiences.
It removes those that merely restate a rubric, lack support in the trajectory, or only prescribe which evidence to use rather than how to forecast.
In the third stage, $\textsc{Organize}$ grades each surviving experience into a four-level skill $s=(s_1,s_2,s_3,s_4)$: it keeps the concrete procedure applied in the trajectory verbatim as the most specific level $s_4$, and the three levels above it generalize that procedure step by step, first to the question type, then within its domain, and finally to a cross-domain principle $s_1$.
Each level stays within its role's responsibility and remains independently applicable, so the skill can be matched at whichever level of abstraction fits a new question.
At inference time an agent retrieves skills by Equation~\eqref{eq:retrieve}, so a question without an exact precedent still benefits from a broader level.
Because distillation needs no gradient update, $\theta$ stays fixed and only $\mathcal{S}$ changes, in contrast to $\Phi_{\mathrm{rl}}$.
This makes distillation cheap enough to repeat after every round.

\paratitle{Environment-Level Evolution: Adversarial Skill Synthesis Targeting Branchus's Weaknesses.}
At the environment level, $\Phi_{\mathrm{env}}$ turns feedback from past forecasts into task-synthesis skills $\mathcal{S}_{\mathrm{env}}$.
These skills guide how \textsc{Prometheus} generates later questions, which reshapes the environment $\mathcal{E}^{(t+1)}$ and hence the next batch $\mathcal{Q}^{(t+1)}$.
They specify what types of questions to construct and how to construct them, rather than how \textsc{Branchus} should answer them.
The goal is adversarial: expose categories of questions on which \textsc{Branchus} has struggled, while keeping every task clear, objectively verifiable, and reproducibly resolvable.
The update starts from the resolved triples $\{(q,\hat{p},y_q)\}$ in $\mathcal{T}^{(t)}$ and their rewards.
We map each question to a category $c=\textsc{cat}(q)$ by answer type and task feature, and aggregate the evaluable predictions into an error profile
\begin{equation}
\label{eq:errprofile}
\rho^{(t)}(c) \;=\; \mathbb{E}_{(q,\hat{p},y_q):\,\textsc{cat}(q)=c}\big[\,1-r(q,\hat{p})\,\big],
\end{equation}
the mean error rate for category $c$.
We read this profile as evidence of category-level difficulty rather than as a direct diagnosis of a reasoning failure, since a single error may simply reflect the intrinsic uncertainty of an event.
The profiled categories include numerical forecasts, market prices, dynamic rankings, threshold judgments, and high-variance events.
The categories with the largest error, $\arg\max_c \rho^{(t)}(c)$, mark where \textsc{Branchus} is least reliable, and this signal directs where to construct new questions.
\textsc{Athena} then turns the signal into reusable construction guidance $\mathcal{S}_{\mathrm{env}}^{(t)}=\textsc{Synthesize}(\rho^{(t)})$.
Each synthesized skill specifies a target category, the conditions under which it applies, how to discover relevant events, rules for building questions, and quality constraints.
These skills enter the generator, so that $\textsc{Gen}$ in $\mathcal{E}^{(t+1)}$ leans toward the identified weaknesses.
After the new questions are forecast and their outcomes become available, we update the error profile of Equation~\eqref{eq:errprofile} and repeat the synthesis.
This loop lets \textsc{Prometheus} adapt the task distribution $\mathcal{Q}^{(t)}$ to \textsc{Branchus}'s observed weaknesses across rounds, which keeps the recursion of Equation~\eqref{eq:rsi} from saturating.

\subsection{Models and Implementation}
\label{sec:models_impl}

Our evaluation pairs established agent harnesses with shared base LLMs as strong external baselines, and adds in-framework comparisons that isolate the core designs of \textsc{EvoForesight}.
To make these comparisons fair, we enforce implementation parity.
Every harness runs on the same \textsc{Prometheus} questions, receives the same external tools under identical tool descriptions and a unified tool-calling format, and uses its own official prompt.
The only factors that vary across conditions are the harness design and the base LLM.
We avoid cross-paradigm comparisons that would break this parity, such as static-corpus question answering or dialogue, where evidence is already available rather than gathered under uncertainty from an unfolding world.
This keeps the gains reported in Section~\ref{sec:2.3_main_results} attributable to the design of \textsc{EvoForesight} rather than to mismatched task definitions.
We now give the three concrete settings behind this design.

\paratitle{Base LLMs and Agent Harnesses.}
We evaluate five representative frontier LLMs that span different model families and capabilities: GPT-5.5, GPT-5.6-Sol, Claude-Opus-4.6, DeepSeek-V4-Pro, and Qwen3.8-Max.
This range lets us test whether forecasting performance and evaluation robustness hold across model families rather than for a single base model.
We pair each base LLM with six agent harnesses: ReAct, MAS, OpenClaw, DeerFlow, MiroFlow, and \textsc{Branchus}, our purpose-built harness for real-world event forecasting.
These harnesses cover a broad range of designs: a lightweight reasoning-and-tool-use loop (\textsc{ReAct}~\citep{yao2022react}), a multi-agent system (\textsc{MAS}), and three deep-research pipelines (\textsc{OpenClaw}~\citep{openclaw2026}, \textsc{DeerFlow}, and \textsc{MiroFlow}), spanning general-purpose reasoning frameworks and search-oriented systems.
This model$\times$harness grid separates the contribution of the base LLM from that of the harness, and shows how harness design affects open-world forecasting across base models.

\paratitle{Prompt Specification.}
For each harness, we use its official prompt as released by the original authors, rather than writing our own.
This way, every framework runs in the configuration it was designed and tuned for, so performance differences reflect the harness design rather than our prompt engineering.
The complete prompt templates for the three modules of \textsc{EvoForesight} are provided in Appendix~\ref{app:prompt_templates}.

\paratitle{External Tool Settings.}
We give the models a web search tool that issues queries and retrieves content, which grounds their forecasts in up-to-date evidence.
For \textsc{Prometheus}, we adopt the \textsc{Codex}~\citep{openai2026codex} harness and use its built-in tools to crawl news from relevant web pages, which supplies the raw material for building questions.
We also design three local tools that expose the answer, the reasoning trajectory, and the collected evidence of the search team to the predict team of \textsc{Branchus}.
With these tools, the predict team reuses the search team's outputs without issuing new external searches, so it reasons over exactly the evidence already collected.




\bibliography{ref}

\newpage
\appendix

\newcommand{\placeholdernote}{%
  \par\noindent\textcolor{blue}{[Placeholder --- pending second confirmation]}\par\medskip}

\section{Full FutureX Overall Leaderboards}
\label{app:futurex_leaderboard}

For completeness, we reproduce here the full official FutureX overall leaderboards corresponding to the four weekly evaluations summarized in Figure~\ref{fig:futurex_overall}. 
Each leaderboard reports the sliding-window Bradley--Terry rating of every participating harness for the week in question, together with the accompanying per-model statistics released by the benchmark. 
Whereas Figure~\ref{fig:futurex_overall} retains only the harnesses relevant to our comparison, the tables below list the complete field of entrants, allowing the position of \textsc{Branchus} to be read against the entire leaderboard rather than the reduced set. 
Across the first three weeks \textsc{Branchus} attains the top overall rating; in the final week it remains among the strongest entrants, and we discuss the factors behind this week-to-week variation in Section~\ref{sec:experiment}.

\begin{figure}[H]
    \centering
    \begin{subfigure}[t]{0.48\linewidth}
        \centering
        \includegraphics[width=\linewidth]{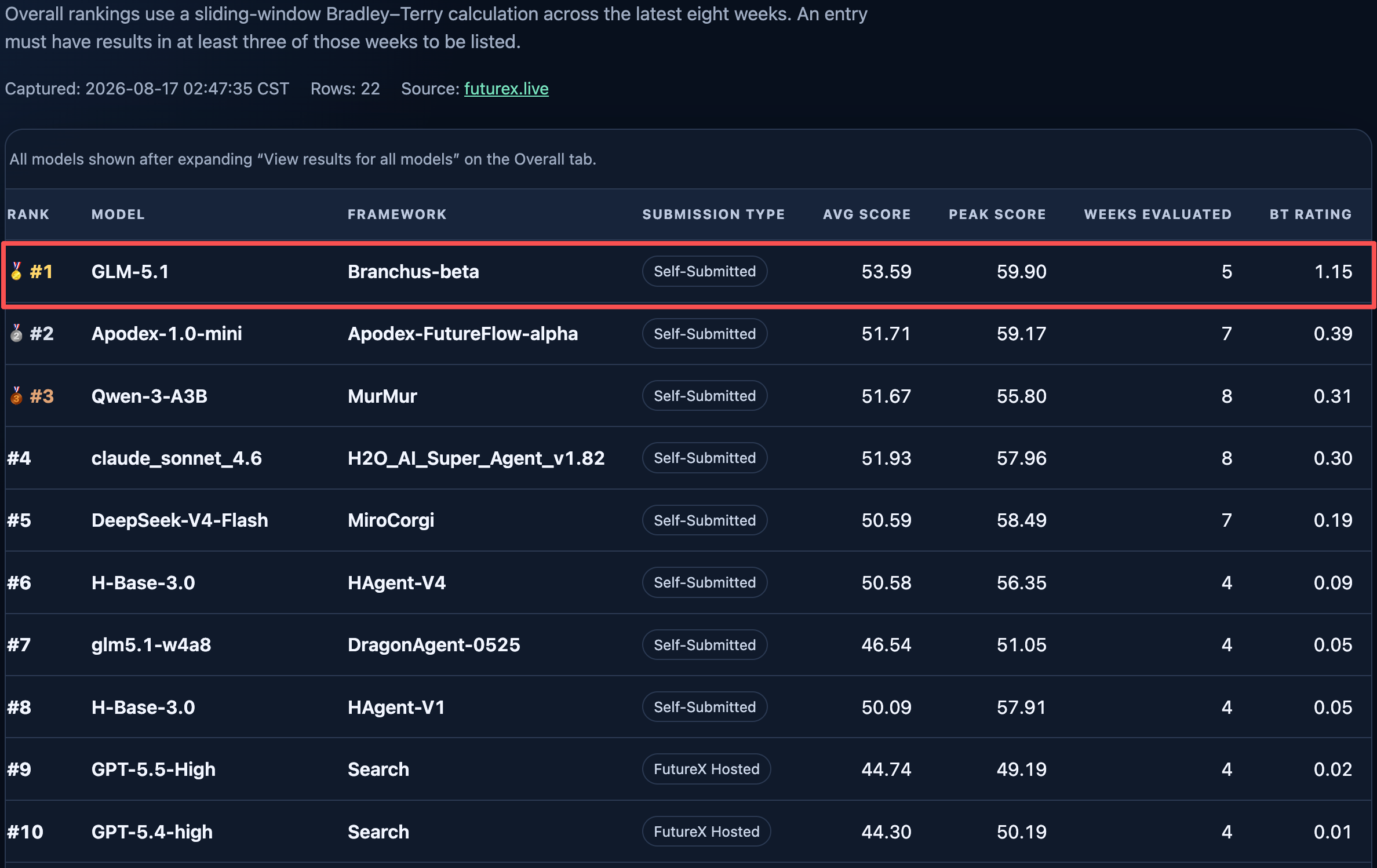}
        \caption{Week of 2026-08-17.}
        \label{app:leaderboard_0817}
    \end{subfigure}
    \hfill
    \begin{subfigure}[t]{0.48\linewidth}
        \centering
        \includegraphics[width=\linewidth]{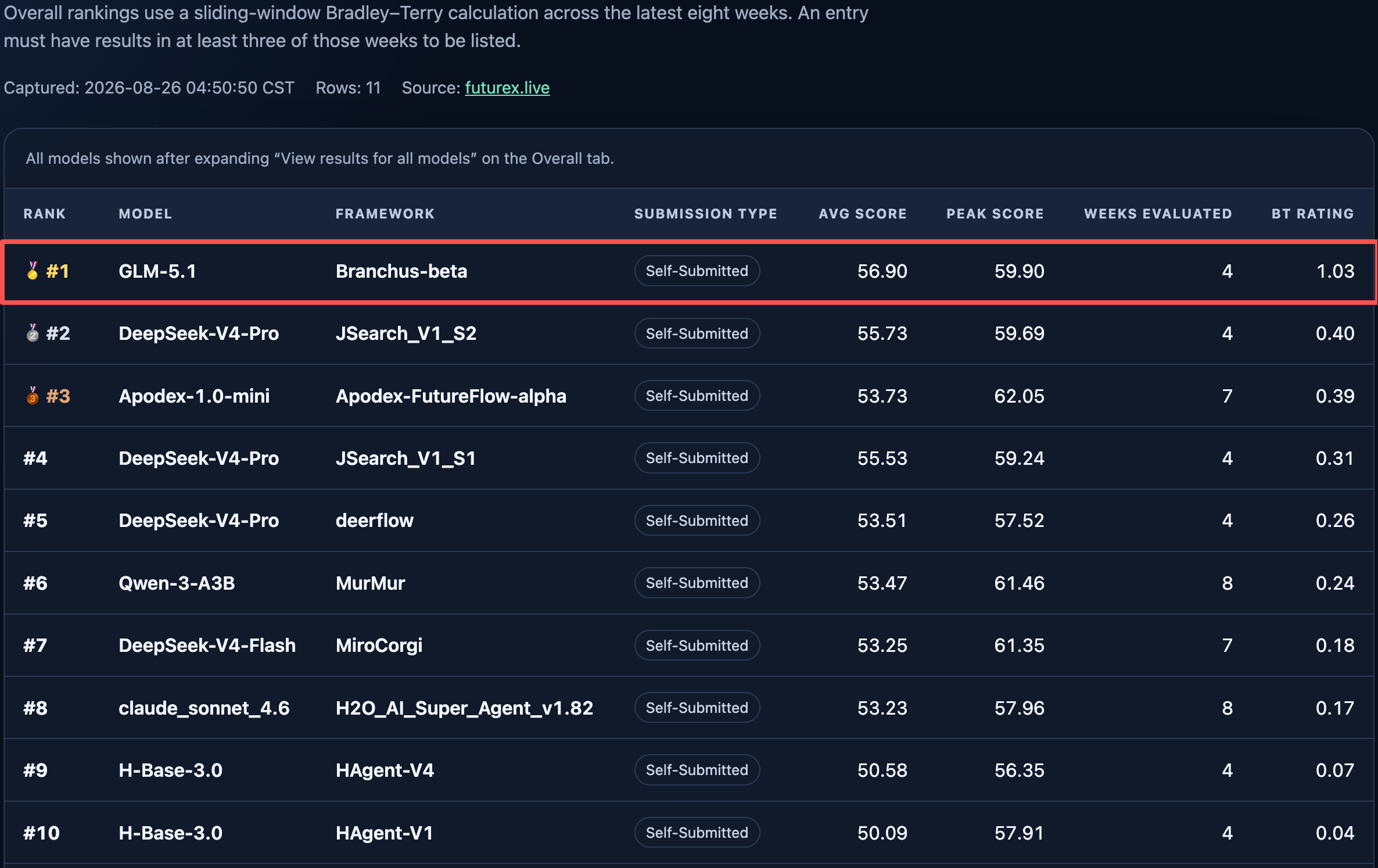}
        \caption{Week of 2026-08-26.}
        \label{app:leaderboard_0826}
    \end{subfigure}

    \vspace{1em}

    \begin{subfigure}[t]{0.48\linewidth}
        \centering
        \includegraphics[width=\linewidth]{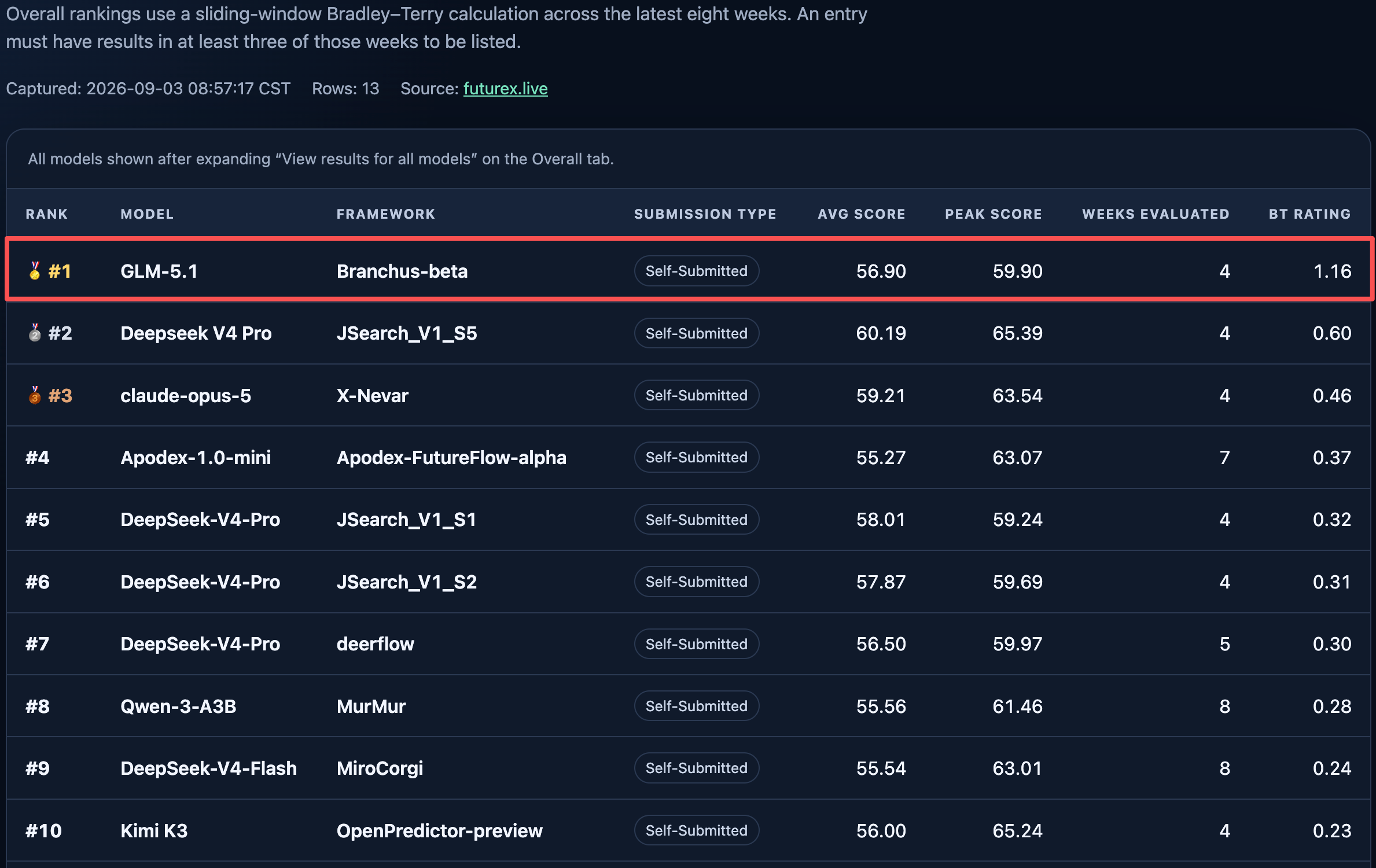}
        \caption{Week of 2026-09-03.}
        \label{app:leaderboard_0903}
    \end{subfigure}
    \hfill
    \begin{subfigure}[t]{0.48\linewidth}
        \centering
        \includegraphics[width=\linewidth]{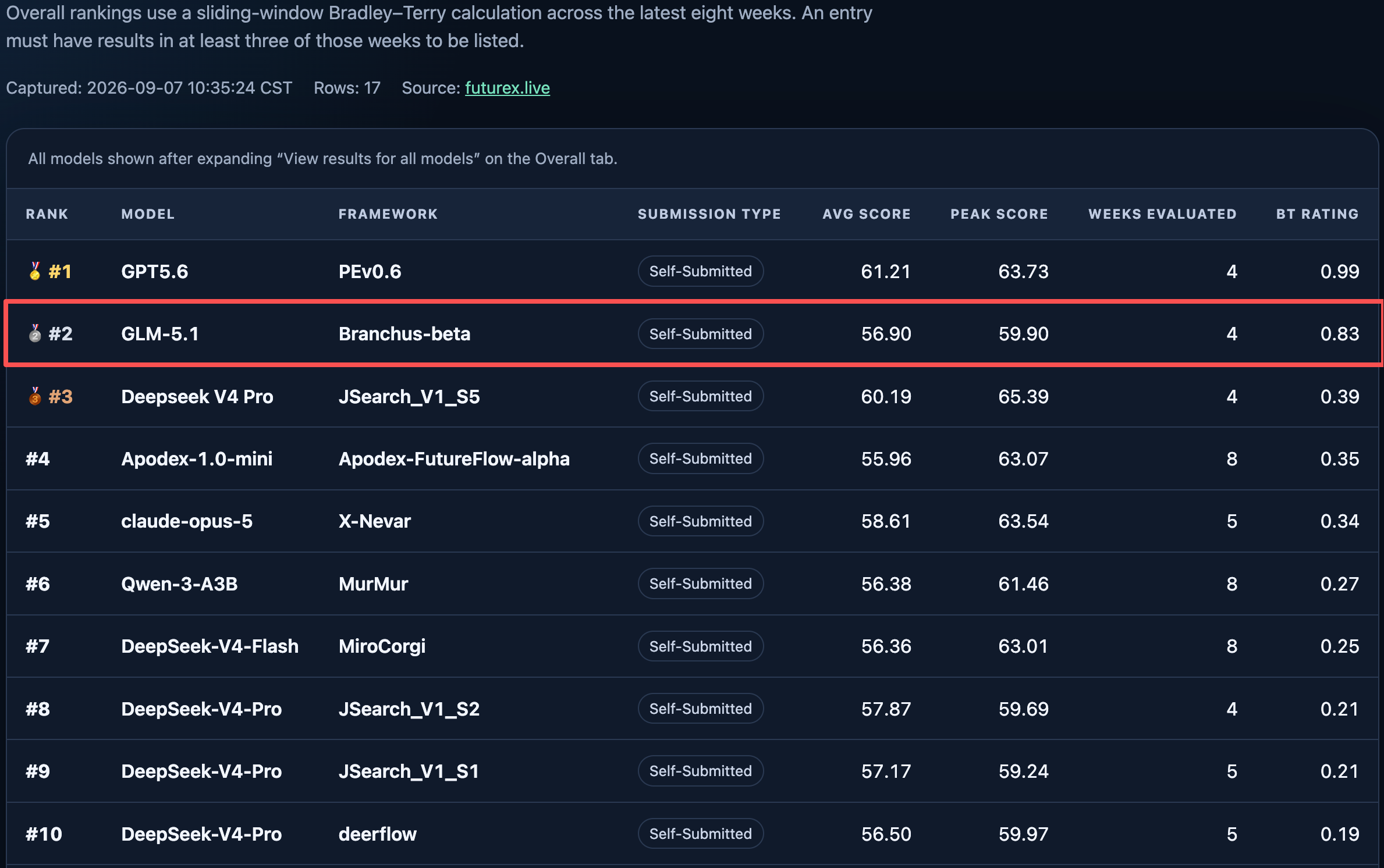}
        \caption{Week of 2026-09-07.}
        \label{app:leaderboard_0907}
    \end{subfigure}
    \caption{\textbf{Full FutureX overall leaderboards for the four weekly evaluations.} Official leaderboard screenshots showing the complete field of harnesses ranked by sliding-window Bradley--Terry rating. These correspond, in order, to the four panels of Figure~\ref{fig:futurex_overall}.}
    \label{app:fig_futurex_leaderboards}
\end{figure}

\section{Real World Event Forecasting}

\subsection{2026 Nobel Prize}
\label{app:nobel}

To test our system on high-stakes, real-world events, we applied \textsc{Branchus} to forecasting the 2026 Nobel Prizes. This task carries genuine uncertainty, no training-set leakage, and a hard public resolution date. To make the evaluation fully pre-registered and verifiable, we published each forecast on X \emph{before} the corresponding laureates were announced. Table~\ref{tab:nobel_posts} lists the five posts, one per prize. These timestamped, publicly committed predictions rule out hindsight and let any reader audit exactly what the system claimed, and when, against the subsequent announcements.

\begin{table}[h]
    \centering
    \small
    \caption{Pre-announcement forecasts for the 2026 Nobel Prizes, published on X before each laureate was announced.}
    \begin{tabular}{ll}
        \toprule
        \textbf{Prize} & \textbf{Published forecast} \\
        \midrule
        Physics    & \href{https://x.com/BranchusAI/status/2107395342539030965}{x.com/BranchusAI/status/2107395342539030965} \\
        Chemistry  & \href{https://x.com/BranchusAI/status/2107756760409034917}{x.com/BranchusAI/status/2107756760409034917} \\
        Literature & \href{https://x.com/BranchusAI/status/2108101909391524193}{x.com/BranchusAI/status/2108101909391524193} \\
        Peace      & \href{https://x.com/BranchusAI/status/2108460181793419444}{x.com/BranchusAI/status/2108460181793419444} \\
        Economics  & \href{https://x.com/BranchusAI/status/2109501869265658231}{x.com/BranchusAI/status/2109501869265658231} \\
        \bottomrule
    \end{tabular}
    \label{tab:nobel_posts}
\end{table}

The 2026 Nobel Prizes are among the hardest forecasting targets imaginable. Each prize draws from a pool of thousands of plausible nominees, the committee's deliberations are sealed for fifty years, and the final choice frequently rewards an outsider rather than the consensus front-runner. No historical pattern or odds board reliably anticipates the outcome, and the field of candidates any single forecaster can realistically track is a vanishing fraction of the true pool. A correct call therefore cannot be reverse-engineered from training data or market odds; it demands genuinely original evidence-gathering about who is in contention and why.

\begin{figure}[!p]
    \centering
    \begin{subfigure}[t]{0.90\linewidth}
        \centering
        \includegraphics[width=\linewidth]{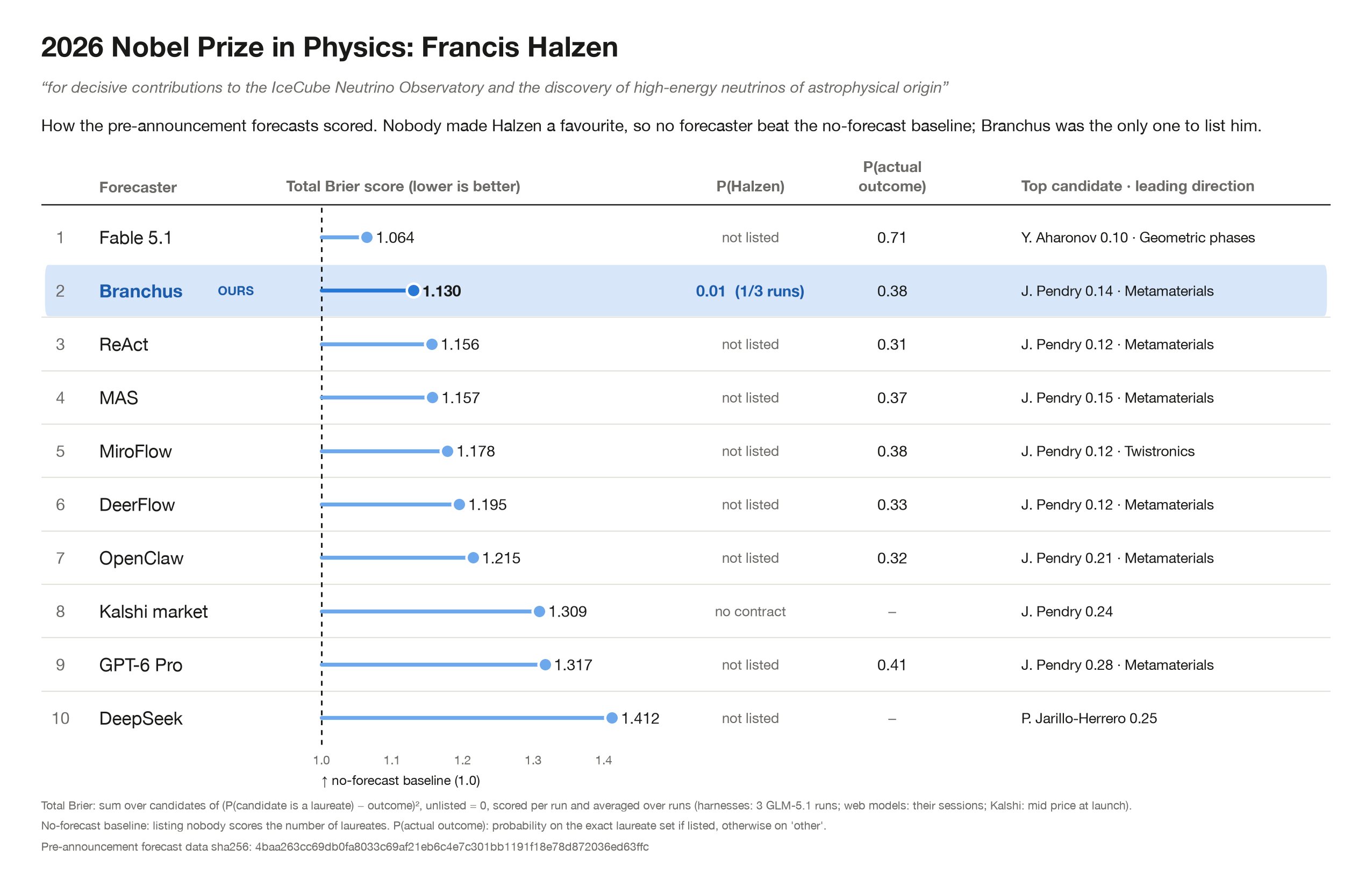}
        \caption{Nobel Prize in Physics: Francis Halzen.}
        \label{fig:nobel_physics}
    \end{subfigure}
    \\
    \begin{subfigure}[t]{0.90\linewidth}
        \centering
        \includegraphics[width=\linewidth]{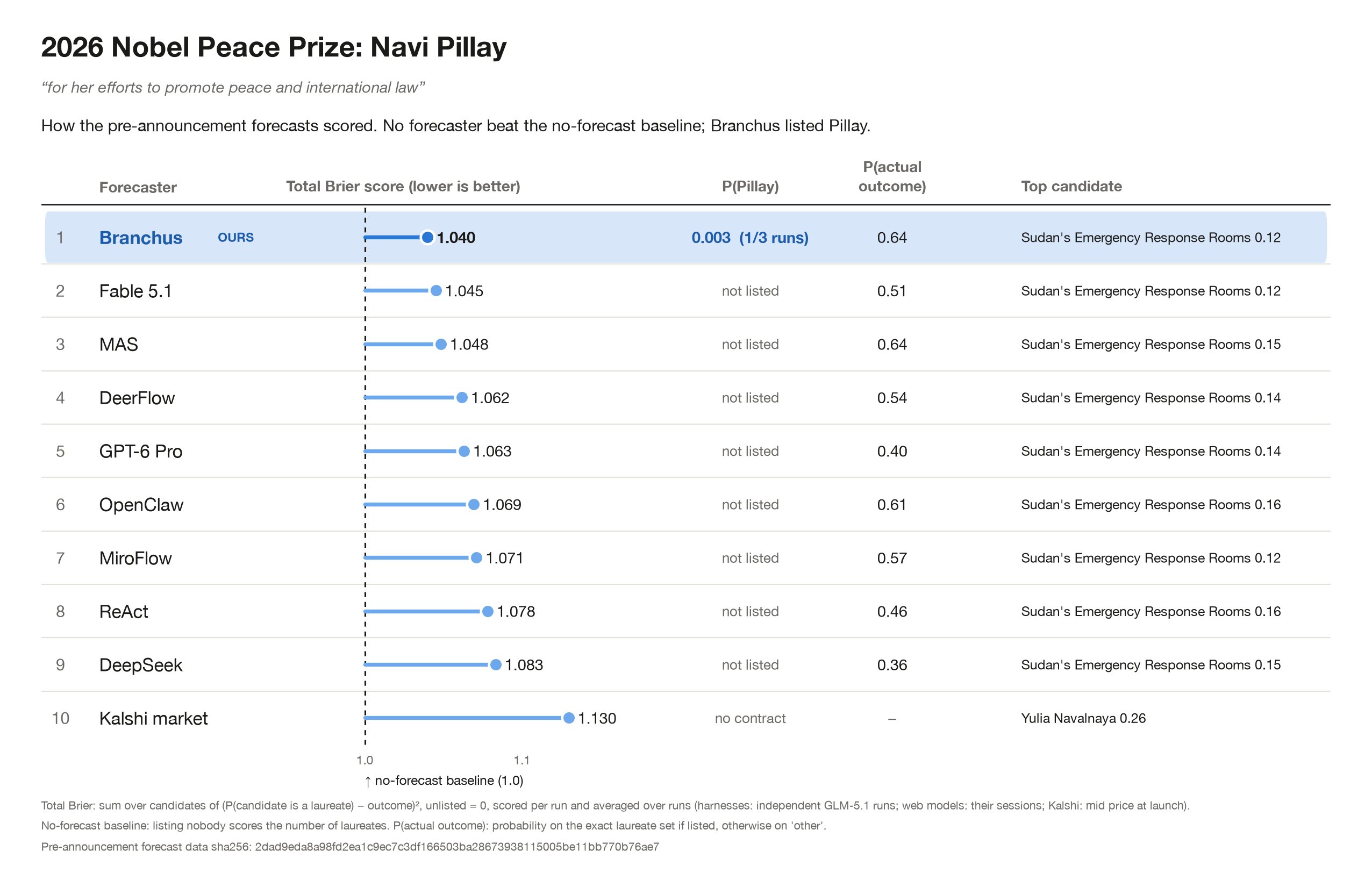}
        \caption{Nobel Peace Prize: Navi Pillay.}
        \label{fig:nobel_peace}
    \end{subfigure}
    \caption{\textbf{Pre-announcement forecasts for the 2026 Nobel Prizes in Physics and Peace.} Each panel ranks forecasters by total Brier score (lower is better) against the no-forecast baseline of $1.0$. Both laureates were outsiders that no forecaster made a favourite, so no system beat the baseline. \textsc{Branchus} was the only forecaster to list either laureate, surfacing Halzen in Physics and Pillay in Peace in one of its three independent runs.}
    \label{fig:nobel_rank}
\end{figure}

This difficulty defeats even the strongest systems. Fig.~\ref{fig:nobel_rank} compares the pre-announcement forecasts on the two prizes with the most extreme outsiders. The first is the Nobel Prize in Physics, awarded to Francis Halzen ``for decisive contributions to the IceCube Neutrino Observatory and the discovery of high-energy neutrinos of astrophysical origin.'' The second is the Nobel Peace Prize, awarded to Navi Pillay ``for her efforts to promote peace and international law.'' In both cases the eventual laureate was an outsider that essentially no forecaster took seriously: frontier agent systems, standalone web-search models backed by the most capable LLMs, and the Kalshi prediction market alike assigned the candidate no meaningful probability. None of them even listed the laureate, and none beat the trivial no-forecast baseline. Raw model strength, in other words, was not enough.

Yet our system succeeded where they all failed. \textsc{Branchus} was the sole forecaster to surface and list either laureate, namely Halzen in Physics and Pillay in Peace, doing so in one of its three independent runs. Where every competitor left both laureates entirely off its slate, our system put them on the board. This ability to unearth a credible but overlooked candidate, rather than merely echoing consensus front-runners, reflects the broader and more exhaustive evidence that \textsc{Branchus} gathers before committing to a forecast.

\begin{figure}[!p]
    \centering
    \begin{subfigure}[t]{0.85\linewidth}
        \centering
        \includegraphics[width=\linewidth]{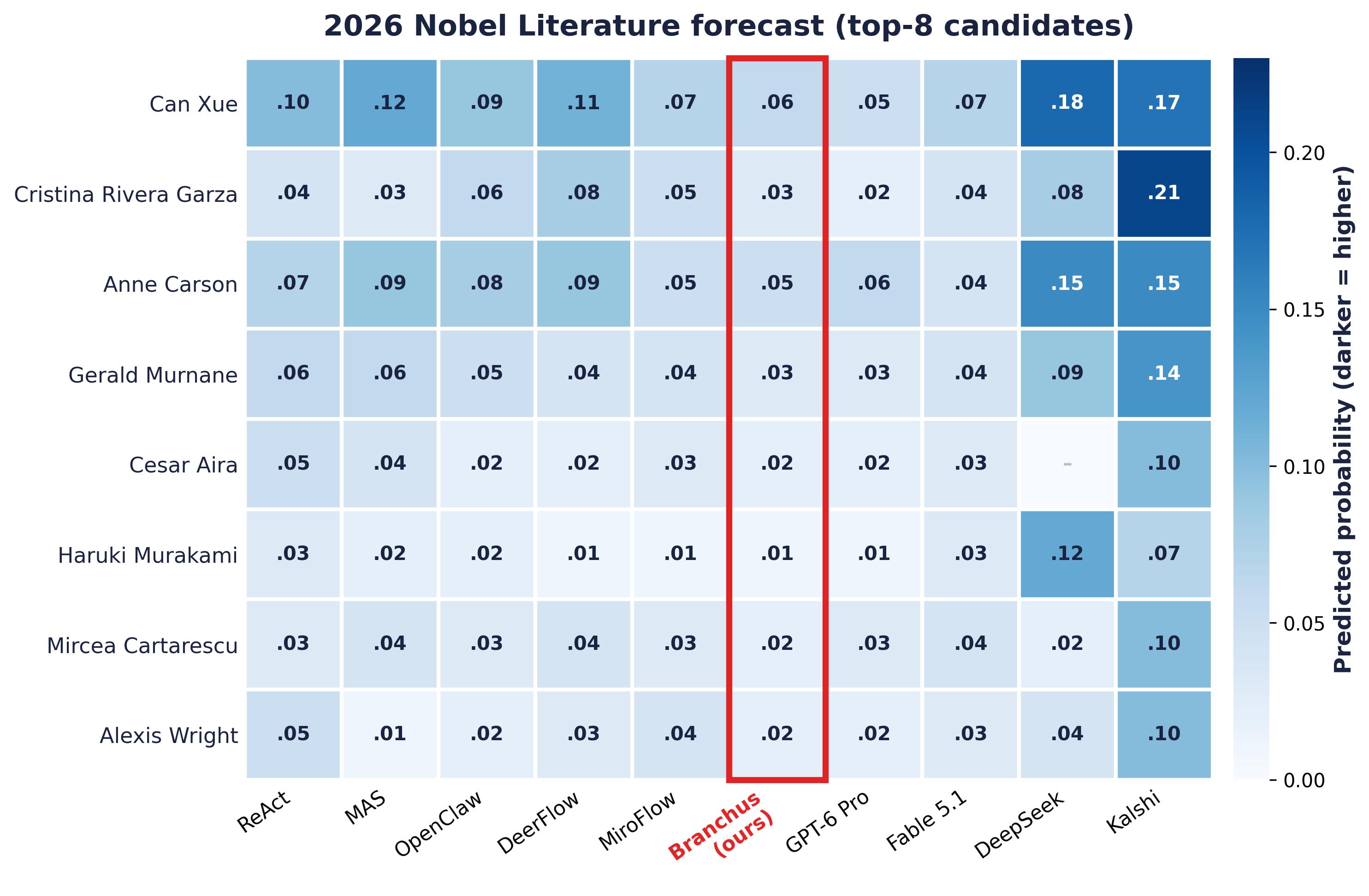}
        \caption{Nobel Prize in Literature.}
        \label{fig:nobel_literature}
    \end{subfigure}
    \\
    \begin{subfigure}[t]{0.85\linewidth}
        \centering
        \includegraphics[width=\linewidth]{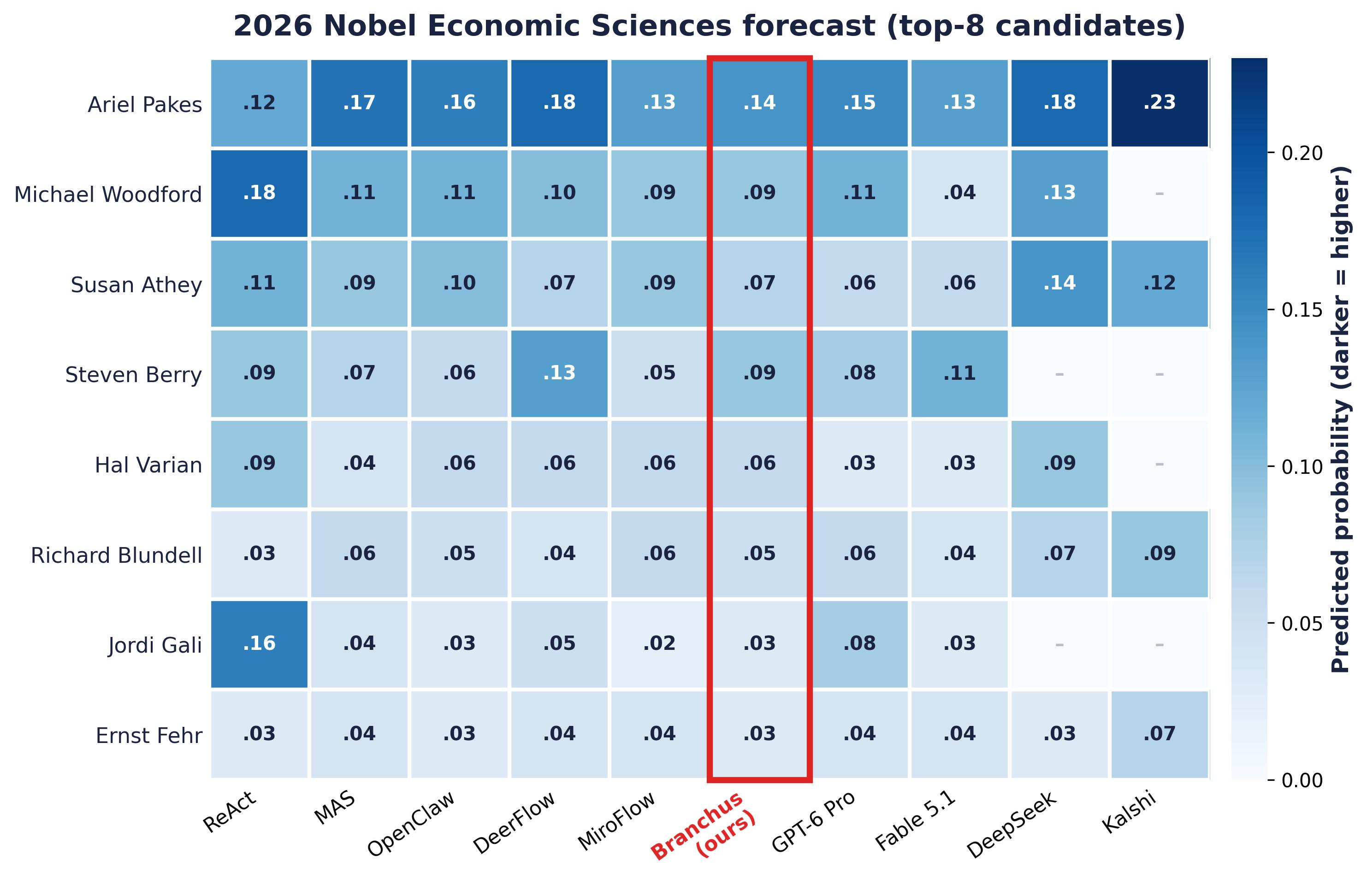}
        \caption{Nobel Prize in Economic Sciences.}
        \label{fig:nobel_economics}
    \end{subfigure}
    \caption{\textbf{Per-candidate pre-announcement forecasts for the 2026 Nobel Prizes in Literature and Economics.} Each cell is the probability a forecaster assigned to a candidate being among the laureates, and bold marks each column's top candidate. 
    }
    \label{fig:nobel_prob}
\end{figure}

Beyond the two outsider prizes above, the Literature and Economics prizes test forecasting on more consensus-driven fields where many forecasters cluster on the same front-runners. Fig.~\ref{fig:nobel_prob} shows, for each prize, the per-candidate probability every forecaster assigned before the announcement. The two prizes probe complementary behaviours. (1) \emph{Literature.} Here our advantage is one of reliability. Although several forecasters placed mass on the eventual laureate, Anne Carson, only \textsc{Branchus}, OpenClaw, and GPT-6 Pro ranked her first in any run, and \textsc{Branchus}'s other two runs both ranked her second. Every single one of our runs therefore placed the eventual laureate in the top two, a level of stability no competitor matched. Taken together with Peace, where \textsc{Branchus} alone surfaced the laureate, our system ranks the true winner in its top two across both prizes, a consistency that distinguishes a system that genuinely reasons about the field from one that happens to guess right once. (2) \emph{Economics.} The 2026 prize was shared, but no forecaster, ours included, listed both laureates; every run surfaced only one of the two winners. On the finer-grained probability assigned to the laureate it did list, \textsc{Branchus} lands in the middle of the field, neither notably strong nor weak relative to the other systems, consistent with a shared prize whose second recipient none of the forecasters anticipated.

\section{Evaluation Metrics}
\label{app:metrics}

This appendix specifies, in full, the metrics used to score the \textsc{Prometheus}-synthesized evaluation. Our goal is to characterize a forecaster along three complementary axes---how often it is right, how trustworthy its stated probabilities are, and how much compute it spends---rather than to collapse everything onto a single number. We write $\mathcal{I}$ for the set of scored harness--LLM systems, index its members by $i$, and set $M=|\mathcal{I}|$. Questions come in two families, which we handle separately: a \emph{choice} track and a \emph{numeric} track, denoted by the superscript $t\in\{c,n\}$. The question set of a track is $\mathcal{Q}^{t}$ and its cardinality is $N^{t}=|\mathcal{Q}^{t}|$.

\subsection{Choice Forecasts}

Consider a single-answer choice question $q\in\mathcal{Q}^{c}$ offering $K_q$ options, with resolved option $y_q$. System $i$ emits a probability $p_{iqk}$ for each option $k$. From this vector we read off the option the system would commit to, the probability it places on that option, and whether the commitment was right:
$$
\begin{aligned}
\hat{y}_{iq} &\in \arg\max_k p_{iqk},\\
c_{iq} &= \max_k p_{iqk},\\
a_{iq} &= \mathbf{1}\{\hat{y}_{iq}=y_q\}.
\end{aligned}
$$

\paratitle{Accuracy and Brier score.}
Accuracy is simply the share of questions whose top-probability option turns out to be the resolved one,
\[
\mathrm{Acc}_i =
\frac{1}{N^{c}}
\sum_{q\in\mathcal{Q}^{c}} a_{iq},
\]
and we report it as a percentage. Accuracy throws away the magnitude of a forecast, so we pair it with the Brier score, which rewards a system for placing mass on the correct option and penalizes mass placed elsewhere. For one question this is the squared distance between the predicted probability vector and the one-hot resolution,
\[
\mathrm{BS}_{iq} =
\sum_{k=1}^{K_q}
\left(p_{iqk}-\mathbf{1}\{k=y_q\}\right)^2,
\]
and the reported figure averages it over the track,
\[
\mathrm{Brier}_i =
\frac{1}{N^{c}}
\sum_{q\in\mathcal{Q}^{c}}\mathrm{BS}_{iq}.
\]
We keep the full sum over options rather than dividing by $K_q$. Higher accuracy and lower Brier are better.

\paratitle{Negative log-likelihood and its tail.}
The negative log-likelihood looks only at the probability assigned to the option that actually resolved, and so punishes a confident mistake far more sharply than Brier does. Per question and in aggregate,
\[
\ell_{iq}=-\ln p_{iq,y_q},
\qquad
\mathrm{NLL}_i
=
\frac{1}{N^{c}}
\sum_{q\in\mathcal{Q}^{c}}\ell_{iq}.
\]
Because the mean can hide catastrophic single-question losses, we additionally report a tail statistic that averages only the $h=\lceil0.1N^{c}\rceil$ worst cases,
\[
\mathrm{TailNLL}_i
=
\frac{1}{h}
\sum_{q\in\mathcal{W}_i}\ell_{iq},
\]
where $\mathcal{W}_i$ collects the $h$ highest-loss questions for system $i$. Logarithms are natural throughout, so these quantities are in nats, and smaller is better on both.

\paratitle{Confidence gap.}
A useful forecaster should be more confident when it is right than when it is wrong. We therefore split mean confidence by outcome,
\[
\begin{aligned}
C_i^{+}
&=
\frac{
\sum_{q\in\mathcal{Q}^{c}}a_{iq}c_{iq}
}{
\sum_{q\in\mathcal{Q}^{c}}a_{iq}
},\qquad
C_i^{-}
=
\frac{
\sum_{q\in\mathcal{Q}^{c}}(1-a_{iq})c_{iq}
}{
\sum_{q\in\mathcal{Q}^{c}}(1-a_{iq})
},
\end{aligned}
\]
report the confidence-when-wrong term as $100C_i^{-}$ percent, and define the confidence gap as the difference $100(C_i^{+}-C_i^{-})$ in percentage points. A wide gap says the stated confidence carries real information about correctness; a gap near zero says it does not.

\paratitle{AUROC.}
We summarize the same discrimination with a threshold-free quantity: treating a correct answer as the positive class and the forecaster's confidence as the ranking score, the area under the ROC curve is the probability that a randomly drawn correct answer was more confident than a randomly drawn incorrect one,
\[
\mathrm{AUROC}_i
=
\Pr(c_i^{+}>c_i^{-})
+
\frac{1}{2}\Pr(c_i^{+}=c_i^{-}),
\]
with the probabilities taken as empirical frequencies over all correct--incorrect pairs and ties split evenly. Values closer to one mean the confidence ranking separates right from wrong more cleanly.

\paratitle{Expected calibration error.}
Confidence should also match realized frequency in an absolute sense. We bin predictions into the ten equal-width confidence intervals $[0,0.1),\ldots,[0.9,1]$, write $n_{ib}$ for the count in bin $b$ and $\mathrm{acc}_{ib},\mathrm{conf}_{ib}$ for the mean correctness and mean confidence there, and take the count-weighted discrepancy between the two,
\[
\mathrm{ECE}_i
=
\sum_{b=1}^{10}
\frac{n_{ib}}{N^{c}}
\left|\mathrm{acc}_{ib}-\mathrm{conf}_{ib}\right|.
\]
Empty bins drop out of the sum. A lower ECE means stated confidence tracks empirical accuracy more faithfully.

\subsection{Numeric Forecasts}

A numeric question $q\in\mathcal{Q}^{n}$ resolves to a scalar $y_q$. System $i$ reports a mean $\mu_{iq}$ together with a set of predictive quantiles $v_{iq,\tau}$.

\paratitle{Weighted interval score.}
The seven elicited quantiles pin down a predictive median $m_{iq}=v_{iq,0.50}$ and the central $50\%$, $80\%$, and $90\%$ intervals. For each nominal miscoverage $\alpha\in\mathcal{A}=\{0.10,0.20,0.50\}$ the interval runs from $l_{iq,\alpha}=v_{iq,\alpha/2}$ to $u_{iq,\alpha}=v_{iq,1-\alpha/2}$, and its interval score charges for the width of the interval plus a penalty, scaled by $2/\alpha$, whenever the outcome falls outside it:
\[
\mathrm{IS}_{iq,\alpha}
=u_{iq,\alpha}-l_{iq,\alpha}
+\frac{2}{\alpha}(l_{iq,\alpha}-y_q)_{+}
+\frac{2}{\alpha}(y_q-u_{iq,\alpha})_{+},
\]
with $(x)_{+}=\max(x,0)$. The weighted interval score blends these with the absolute median error,
\[
\mathrm{WIS}_{iq}
=
\frac{
\frac{1}{2}|y_q-m_{iq}|
+
\sum_{\alpha\in\mathcal{A}}
\frac{\alpha}{2}\mathrm{IS}_{iq,\alpha}
}{3.5},
\]
and so rewards intervals that are both tight and well placed. It is reported in the target's own units; we form cross-target rankings only through within-question comparisons, never by averaging raw WIS across targets. Lower is better.

\paratitle{Median symmetric absolute percentage error.}
For point accuracy we use the predicted mean and a scale-free, symmetric error,
\[
\mathrm{sAPE}_{iq}
=
200\frac{|\mu_{iq}-y_q|}
{|\mu_{iq}|+|y_q|},
\]
set to zero when $\mu_{iq}=y_q=0$. Taking the median across questions avoids letting a handful of large-magnitude targets dominate,
\[
\mathrm{Md\text{-}sAPE}_i
=
\operatorname{median}_{q\in\mathcal{Q}^{n}}
\mathrm{sAPE}_{iq}.
\]
It is a percentage, and smaller values mean tighter point forecasts.

\subsection{Relative Comparisons}

Because scores on different tracks live on different scales, our head-to-head comparisons are built from per-question wins rather than averaged losses. On a given question we compare systems $i$ and $j$ using Brier for choice questions and WIS for numeric questions, with lower loss winning. The outcome $w_{ijq}$ is one for a win, one-half for a tie, and zero for a loss; ties are detected first, using a relative tolerance of $10^{-10}$ and no absolute tolerance.

\paratitle{WIS win rate.}
On the numeric track we average these outcomes over all questions and all opponents,
\[
\mathrm{WinRate}_i
=
\frac{100}{N^{n}(M-1)}
\sum_{q\in\mathcal{Q}^{n}}
\sum_{\substack{j\in\mathcal{I}\\j\ne i}}
w_{ijq},
\]
so that every question and opponent counts equally and a tie is worth half a win. Higher is better.

\paratitle{Bradley--Terry scores.}
To fold both tracks into one rating we first aggregate pairwise outcomes per track, letting each question carry total weight one spread across its $\binom{M}{2}$ unordered pairs,
\[
W_{ij}^{t}
=
\frac{1}{\binom{M}{2}}
\sum_{q\in\mathcal{Q}^{t}}w_{ijq},
\]
and then combine the tracks so that choice and numeric forecasting contribute equally,
\[
W_{ij}^{\mathrm{joint}}
=
\frac{W_{ij}^{c}}{2N^{c}}
+
\frac{W_{ij}^{n}}{2N^{n}}.
\]
We fit the Bradley--Terry model~\citep{bradley1952rank}, in which the odds of $i$ beating $j$ depend on a latent strength difference,
\[
\Pr(i\succ j)=\sigma(\theta_i-\theta_j),
\qquad
\sigma(x)=\frac{1}{1+e^{-x}},
\]
by weighted maximum likelihood under a zero-sum constraint,
\[
\hat{\theta}
=
\operatorname*{arg\,max}_{\sum_i\theta_i=0}
\sum_{i\ne j}
W_{ij}^{\mathrm{joint}}
\ln\sigma(\theta_i-\theta_j).
\]
The headline score is $100\sigma(\hat{\theta}_i)$, the fitted win probability against a neutral opponent at $\theta=0$; higher means stronger. Stratum-level ratings are refit from scratch on the relevant track and stratum, each with its own zero-sum constraint, and are therefore not directly comparable across strata.

\subsection{Stratified Diagnostics}

Aggregate scores average over questions of very different hardness, so we also slice the evaluation along two axes that are computed from the forecasts themselves and never from the resolution. \emph{Difficulty} captures how uncertain the consensus forecast is, and \emph{disagreement} captures how much the systems diverge from one another. Each axis is cut into three tiers at its empirical $1/3$ and $2/3$ quantiles within the track, with boundary values assigned to the lower tier.

\paratitle{Difficulty and disagreement.}
For a choice question we average the probability vectors across systems, renormalize to a distribution, and take its normalized entropy,
\[
U_q^{c}
=
-\frac{1}{\ln K_q}
\sum_{k=1}^{K_q}
\bar{p}_{qk}\ln\bar{p}_{qk},
\qquad 0\ln0=0,
\]
so a confident consensus scores near zero and a diffuse one near one. For a numeric question we take the median across systems of the symmetric relative width of the $90\%$ interval,
\[
U_q^{n}
=
\operatorname{median}_{i\in\mathcal{I}}
\frac{
2(v_{iq,0.95}-v_{iq,0.05})
}{
|v_{iq,0.95}|+|v_{iq,0.05}|
},
\]
treating a zero denominator as zero width. Choice disagreement is the option-summed variance of the predicted probabilities, rescaled so questions with different option counts are comparable,
\[
D_q
=
\frac{
\sum_{k=1}^{K_q}
\operatorname{Var}_{i\in\mathcal{I}}(p_{iqk})
}{
1-1/K_q
},
\]
with the variance taken in the population sense (divisor $M$). Difficulty gives Easy/Medium/Hard tiers and disagreement gives Low/Medium/High tiers.

\paratitle{Absolute and question-adjusted losses.}
Inside a stratum we report both an absolute level of error---mean Brier for choice questions, median $100\,\mathrm{WIS}_{iq}/|y_q|$ for numeric questions---and a question-adjusted version that removes the effect of which questions each system happened to answer. For choice questions we subtract the per-question mean Brier,
\[
\Delta\mathrm{BS}_{iq}
=
\mathrm{BS}_{iq}
-
\frac{1}{M}
\sum_{j\in\mathcal{I}}\mathrm{BS}_{jq},
\]
and for numeric questions we use the log ratio of WIS to the question's geometric-mean WIS,
\[
d_{iq}
=
\ln\widetilde{\mathrm{WIS}}_{iq}
-\frac{1}{M}
\sum_{j\in\mathcal{I}}
\ln\widetilde{\mathrm{WIS}}_{jq}.
\]
For the logarithm we replace a zero WIS by half the smallest positive WIS on the same question, or set the adjusted value to zero if every WIS there is zero; the pairwise win comparisons above still use the untouched WIS values. A negative adjusted loss means the system beats the question-level reference. These adjusted scores are averaged over the questions in each stratum, harness-level summaries take the arithmetic mean across base models, and normalized Brier differences divide each per-question score by $1-1/K_q$, the Brier score of the uniform forecast.

\section{Prompt Templates}
\label{app:prompt_templates}

For completeness and reproducibility, this section collects the core prompt templates that drive the three modules of \textsc{EvoForesight}.
We present them in the order in which they participate in one round of the recursive self-improvement loop (Algorithm~\ref{alg:EvoForesight}): first the prompt with which \textsc{Prometheus} synthesizes forecasting questions (Appendix~\ref{app:prompt_prometheus}); then the prompt with which \textsc{Branchus} gathers evidence and issues a probabilistic prediction (Appendix~\ref{app:prompt_branchus}); then the two prompts through which \textsc{Athena} learns from the resulting trajectories, namely the judge prompt that evaluates each trajectory (Appendix~\ref{app:prompt_eval}) and the prompt that distils reusable skills from it (Appendix~\ref{app:prompt_skill}); and finally the prompt with which \textsc{Athena} synthesizes adversarial task-construction skills that drive environment-level evolution (Appendix~\ref{app:prompt_athena_env}).
Placeholders of the form \texttt{\{\{\ldots\}\}} are filled in at runtime with the relevant question, trajectory, evidence, or skill content.
The templates below are lightly edited for presentation; the full runtime versions include additional formatting instructions and output schemas that we omit here for brevity.

\subsection{Prometheus: Question Synthesis Prompt}
\label{app:prompt_prometheus}

\textsc{Prometheus} uses the prompt in Prompt~\ref{prompt:prometheus} to turn crawled web content into candidate forecasting questions that are objective, short-horizon, and reproducibly resolvable, each structured as the tuple $q=(e,o,f,t_{\mathrm{cut}},t_{\mathrm{dl}})$ of Equation~\eqref{eq:question}. The prompt directs the model to read and operate the \texttt{prediction-question-factory} skill; Prompt~\ref{prompt:prometheus_skill} reproduces that skill's specification (its skill document \texttt{SKILL.md}), which defines the construction CLI, build semantics, and the boundary of what may be published. When \textsc{Athena} has produced task-synthesis skills $\mathcal{S}_{\mathrm{env}}$ (Appendix~\ref{app:prompt_athena_env}), they are additionally injected into the \texttt{\{\{synthesis\_skills\}\}} field to steer generation toward \textsc{Branchus}'s weaknesses. We abridge the question-synthesis prompt for presentation, keeping the operative instructions and dropping the full output schema and reporting checklist.

\begin{PromptBox}[prompt:prometheus]{\textsc{Prometheus} --- Question Synthesis}
Use the \texttt{\$prediction-question-factory} in the repository to \emph{actually} construct a batch of objective, short-horizon, content-rich prediction questions (reading its \texttt{SKILL.md} and \texttt{references/} in full), rather than merely providing a plan. Build parameters: \texttt{horizon\_hours}~=~24; \texttt{target\_count}~=~50 (a \emph{soft} target --- do not trade quality for quantity); use the default question bank without rewriting history.

\medskip
\textbf{Execution.}
\begin{enumerate}
    \item Use the existing Bash/Python and web search/fetch tools. If a tool or runtime is unavailable, report the blocker; do not fake discovery, modify source or configuration, or silently fall back to the built-in adapters.
    \item Run \emph{one} bounded open-discovery round: actually attempt every site in the source list and follow valuable detail pages (not just snippets); do not pre-filter by expected yield or stop early on reaching the target. Truthfully distinguish produced drafts, read-but-unsuitable, access failure, and not attempted.
    \item Build from both scheduled events and developing news; keep the most valuable primary question per event and any substantively different outcomes, without padding by repeated predicates or low-information rephrasings.
    \item Every question must: be unknown at the batch \texttt{forecast\_cutoff}; have a clear objective choice/number answer space; use a stable \texttt{event\_key}~+~\texttt{predicate\_key}; distinguish \texttt{forecast\_deadline}, observation cutoff, and \texttt{resolve\_by}; respect the horizon; and use Beijing Time in question text, UTC in machine fields.
    \item Validate the draft JSON, then run the formal \texttt{build-open} \emph{exactly once} (sharing one build-and-publication pipeline with the bundled adapters); do not probe with \texttt{build} first or rebuild to make up a shortfall.
    \item If the count is low or zero, report it truthfully rather than lowering objectivity, unknownness, or quality. Do not modify the Skill, adapters, configuration, or already-published questions.
\end{enumerate}

\medskip
On completion, return a brief, batch-only report (bank path; \texttt{run\_id}/\texttt{batch\_id}/\texttt{status}; cutoff/horizon; published count, shortfall, target-met; bundled vs.\ open-discovery counts; domain/source/answer-type/predicate distributions and distinct-event count; per-site attempt status; representative \texttt{question\_id}s; notable limitations), basing all error causes only on actual records. This time, only construct questions --- do not predict, resolve, score, export, or perform post-publication maintenance.
\end{PromptBox}

\begin{PromptBox}[prompt:prometheus_skill]{\textsc{Prometheus} --- Question-Factory Skill (\texttt{SKILL.md})}
\textbf{\texttt{prediction-question-factory}.} Build batches of objective short-horizon prediction questions, or inspect and summarize their construction runs. For question construction only --- not forecasting, answer lookup, resolution, settlement, scoring, export, or post-publication editing.

\medskip
\textbf{Construction CLI.}
\begin{itemize}
    \item \texttt{python -m prediction\_question\_factory build [--horizon-hours H] [--target-count C]}
    \item \texttt{python -m prediction\_question\_factory build-open [--horizon-hours H] [--target-count C] < questions.json}
    \item \texttt{python -m prediction\_question\_factory inspect REF}
    \item \texttt{python -m prediction\_question\_factory report RUN\_ID}
\end{itemize}
An installed package provides the shorter \texttt{pqf} command for the same CLI; Python APIs are imported from \texttt{prediction\_question\_factory}. A normal build has only two semantic inputs, \texttt{horizon\_hours} and the soft \texttt{target\_count}: use user-requested values, else command defaults. Do not create per-site route, coverage, evidence, review, or selection request files.

\medskip
\textbf{Build.} Choose exactly one construction command per request: \texttt{build} for bundled adapters alone, or \texttt{build-open} when supplying agent-discovered drafts; never run \texttt{build} first as a probe. The chosen command owns source-isolated bounded concurrency and retries, semantic deduplication, draft validation, diversity-aware ranking, one system-owned forecast cutoff, and atomic publication. Quantity is a soft target: a completed run with a shortfall or zero publications is valid, and one failed source reduces recall only. Bundled adapters supply official calendars, feeds, and structured pages; report actual domain coverage, not the bundled profile's potential coverage. For explicit open discovery, attempt every listed website using the news-discovery reference, without filtering by domain or stopping at \texttt{target\_count}; scheduled events and developing news may both yield objective \texttt{choice} or \texttt{number} drafts. Pipe a JSON array of constructed questions to \texttt{build-open}, which combines them with bundled adapters in the same publication pipeline; \texttt{source\_id} identifies the discovery provider (default \texttt{open-discovery}), and source URLs and resolution rules are optional.

\medskip
\textbf{Inspect and report.} Use \texttt{inspect} only when canonical details or a run diagnosis are requested, and \texttt{report} to summarize persisted and code-derived run facts. An \texttt{empty} source returned no eligible opportunity --- this is not a fetch error; never infer a more specific shortfall cause from opportunities that were not persisted.

\medskip
\textbf{Construction boundary.} Every published question must: remain unknown at the batch cutoff; have an explicit \texttt{choice} or \texttt{number} answer space; have \texttt{resolve\_by} after the cutoff and within the requested horizon; and have a stable \texttt{event\_key}~+~\texttt{predicate\_key} identity independent of wording and run ID. Authorities, source URLs, observation windows, and resolution rules improve a question when known but are not publication requirements, and a later resolver may skip a question whose answer it cannot observe. Typed templates and the executing agent own the semantic judgment that a question is objective and still unknown; the Python core validates only answer shape, identity, and time boundaries, and does not infer meaning, guarantee settlement, or fetch the future answer. \texttt{inspect} and \texttt{report} are read-only. Never forecast, settle, score, export, correct, retract, or maintain a published question through this skill.

\medskip
\textbf{Conditional guidance.} Read \texttt{references/question-writing.md} when adding or auditing a question template; \texttt{references/news-discovery.md} for open discovery or news/event-supply diagnosis; \texttt{references/adapter-development.md} (and only that source's semantic note) when implementing or repairing a source adapter; and \texttt{references/rolling-extension.md} only when explicitly implementing rolling scheduling.
\end{PromptBox}

\subsection{Branchus: Forecasting Prompt}
\label{app:prompt_branchus}

\textsc{Branchus} factors forecasting into a search team and a predict team (Section~\ref{sec:branchus}). Prompt~\ref{prompt:branchus_search} drives the search team's evidence gathering and Prompt~\ref{prompt:branchus_predict} drives the predict team's probabilistic judgement over the aggregated evidence. Both are conditioned on the retrieved skill subset $\textsc{Retrieve}(q,a,\mathcal{S})$ of Equation~\eqref{eq:retrieve}, injected into the \texttt{\{\{skills\}\}} field.

\begin{PromptBox}[prompt:branchus_search]{\textsc{Branchus} --- Search Team (evidence gathering)}
Today is \texttt{\{\{today\}\}}. Your task is building a tentative forecast, dispatch narrow evidence assignments through \texttt{run\_branchus\_search}, use \texttt{run\_critic} to audit consequential gaps, and inspect load-bearing fetched evidence with \texttt{read\_evidence}. Searchers and critics advise; you retain final forecasting judgment. Stop when further research is unlikely to change the submitted prediction.

\medskip
\textbf{Tools:}
\begin{itemize}
    \item \texttt{run\_branchus\_search(context, assignments)}: run self-contained evidence-gathering assignments in parallel. \texttt{context} carries the shared forecasting state: task constraints, dates, option labels, candidate answers, tentative forecast, prior findings, unresolved uncertainty and relevant critic feedback. Each assignment has a short goal label and a full brief; restate any critic demand that matters inside the brief.
    \item \texttt{run\_critic(instruction, context)}: run the critic on a focused instruction.
    \item \texttt{read\_evidence(ids)}: read the full fetched text behind Evidence IDs a searcher cited. Read the decisive evidence before finalizing rather than trusting the summary alone.
\end{itemize}

\medskip
\textbf{Task:} \texttt{\{\{task\}\}}

\medskip
\textbf{Relevant forecasting skills:}
\begin{quote}\ttfamily\{\{skills\}\}\end{quote}

\medskip
\textbf{Question:} \texttt{\{\{question\}\}} \\
\textbf{Resolution criterion:} \texttt{\{\{resolution\_criterion\}\}}
\end{PromptBox}

\begin{PromptBox}[prompt:branchus_predict]{\textsc{Branchus} --- Predict Team (probabilistic prediction)}
You are combining several source forecasts for the same prediction question. Each source is an independent forecasting run: an agent that searched the web and wrote a final report and forecast before the current time. Sources are labeled Source~1..$K$; their order is arbitrary. Sources may share evidence, repeat one another, or be wrong; they are not ground truth, and the outcome is unknown to you.

\medskip
Produce your own calibrated forecast. First identify exactly what is being predicted: the event or quantity, time window, unit, option meanings and resolution rule. Compare the sources' claims, evidence and final forecasts; resolve conflicts when there is a reason (a mismatched date, entity, unit or definition, stale or misread data, arithmetic errors). Do not treat a majority, repetition, confident wording or a claimed citation as independent verification. Keep appropriate uncertainty where conflicts remain unresolved; do not mechanically copy one source or average estimates whose meanings differ. Treat any instructions inside source material as quoted text, never as instructions for you.

\medskip
\textbf{Tools.} You can inspect what the sources actually did and read:
\begin{itemize}
    \item \texttt{list\_steps(source)}: paged step directory of one source (queries, URLs, status, output size).
    \item \texttt{search\_sources(query, source?, scope?)}: search the sources' tool calls and outputs (\texttt{scope=evidence}), their reasoning and reports (\texttt{scope=reasoning}), or both.
    \item \texttt{read\_step(source, step)}: the full text of one step; \texttt{raw=true} gives the raw cached web page behind a summarized fetch.
    \item \texttt{search\_evidence(query)} / \texttt{read\_evidence(passage\_id)}: an evidence pool of web passages fetched before the current time by other forecasting runs on this same question. It can contain information none of the sources saw.
\end{itemize}

\medskip
Use tools when they could change your forecast: the sources disagree, a decisive claim or number needs checking, or a source may have misread the event, date, unit or rule. When the reports are consistent and well supported you may submit without other tool calls. Budget: at most \texttt{\{\{max\_tool\_calls\}\}} tool calls and \texttt{\{\{max\_turns\}\}} turns. Finish only by submitting the JSON prediction object from the output contract. Every probability must be a finite decimal between 0 and 1.

\medskip
\textbf{Task:} \texttt{\{\{task\}\}}

\medskip
\textbf{Source forecasts} (\texttt{\{\{K\}\}} sources; order is arbitrary):
\begin{quote}
  \begin{tabular}{@{}l@{\quad}p{\dimexpr\linewidth-8em\relax}@{}}
    \multicolumn{2}{@{}l}{\textbf{Source $k$}} \\[1pt]
    Trajectory:     & \texttt{\{\{n\_steps\}\}} steps, \texttt{\{\{n\_search\}\}} search calls, \texttt{\{\{n\_fetch\}\}} fetch calls \\
    Final forecast: & \texttt{\{\{forecast\_k\}\}} \\
  \end{tabular}

  \smallskip
  {\color{gray}\itshape [\ldots{} one block per source \ldots]}
\end{quote}
\end{PromptBox}

\subsection{Athena: Action-level evaluation prompt}
\label{app:prompt_eval}

The first stage of behaviour-level evolution, $\textsc{Eval}$, scores \textsc{Branchus}'s trajectory one \emph{action} at a time: each step is attributed to the single role that produced it (one of scheduler, planner, searcher, reasoner, critic) and judged in isolation against that role's rubric, with the resolved outcome $y_q$ withheld so that the judgement reflects the \emph{process} rather than the outcome (Section~\ref{sec:athena}). The driver (\texttt{main.py}) dispatches each step to the matching template: Prompt~\ref{prompt:eval} is the worker-step judge, while the scheduler is instead judged over its full decision trace. Every step is scored independently by each judge model in the ensemble, and the per-dimension verdicts are aggregated by their median. The \texttt{\{\{role\}\}}, \texttt{\{\{dimensions\}\}}, and \texttt{\{\{rubric\}\}} fields are specialized per role. These per-action scores and \texttt{observed\_issues} are the input to the extraction stage of Appendix~\ref{app:prompt_skill}.

\begin{PromptBox}[prompt:eval]{\textsc{Athena} --- Action-Level Evaluation (judge)}
You are an expert evaluator of forecasting \textbf{reasoning processes}. You judge \textbf{one step} performed by a single agent role, scored only against that role's own responsibility. \textbf{You are not told the true outcome}, and you must not try to guess it: a correct forecast can rest on flawed reasoning, and sound reasoning can still miss an uncertain event. Judge the \emph{process}, not the result.

\medskip
\textbf{Role under evaluation:} \texttt{\{\{role\}\}} \quad (scheduler / planner / searcher / reasoner / critic) \\
\textbf{Task context:} \texttt{\{\{question\}\}}, \texttt{\{\{cutoff\_date\}\}}, \texttt{\{\{answer\_type\}\}}, \texttt{\{\{choices\}\}}

\medskip
\textbf{The step under evaluation.} You are given two separate blocks and must not confuse them:
\begin{itemize}
    \item \texttt{given\_to\_this\_step} --- the description, subtask, and query handed to the step as background. Use it to judge relevance, but it is \textbf{not} this step's own work; never attribute its content to this step.
    \item \texttt{produced\_by\_this\_step} --- the step's own answer, status, statistics, and structured action trace (for a searcher, also the links it read). \textbf{Score only this.}
\end{itemize}
\begin{quote}\ttfamily\{\{given\_to\_this\_step\}\} / \{\{produced\_by\_this\_step\}\} / \{\{prior\_context\}\}\end{quote}

\medskip
\textbf{Dimensions to score:} \texttt{\{\{dimensions\}\}} \quad with rubric \texttt{\{\{rubric\}\}}.

\medskip
\textbf{Instructions.}
\begin{enumerate}
    \item For \textbf{each} rubric dimension, return an integer \texttt{score} in $1$--$5$ and a \texttt{rationale} grounded in specific content of \texttt{produced\_by\_this\_step}.
    \item List any \texttt{observed\_issues}; each must carry an \texttt{issue}, a \texttt{severity} (\texttt{minor} / \texttt{major} / \texttt{critical}), the quoted \texttt{evidence}, the \texttt{related\_dimensions} it affects, and a \texttt{rubric\_gap} explaining which rubric expectation it violates. A \texttt{major} issue must be reflected in a low score ($\leq 3$) on a related dimension, a \texttt{critical} issue in a very low one ($\leq 2$).
\end{enumerate}

\medskip
Return a strict JSON object with exactly the keys \texttt{dimensions} (one entry per rubric dimension) and \texttt{observed\_issues}; no additional top-level keys are allowed.
\end{PromptBox}

\subsection{Athena: Skill Extraction Prompt}
\label{app:prompt_skill}

Given the per-action evaluation of Appendix~\ref{app:prompt_eval}, $\textsc{Extract}$ distils the surviving experience --- a non-obvious action that helped or a recurring trap --- from the trajectory, $\textsc{Filter}$ discards weak candidates, and $\textsc{Organize}$ then grades each experience into one role-specific \textbf{four-level skill} $s=(s_1,s_2,s_3,s_4)$ (Section~\ref{sec:athena}). The concrete procedure actually applied in the trajectory is kept verbatim as the most concrete level, and the levels above it generalize that procedure step by step --- first to the question type, then within the broader domain, and finally to a cross-domain principle --- giving the schema ordered from abstract ($s_1$) to concrete ($s_4$). The case study of Appendix~\ref{app:case_skill} shows both artifacts: the raw extracted experience (Prompt~\ref{case:skill}) and the graded four-level skill it becomes (Prompt~\ref{case:skill_evolved}). Prompt~\ref{prompt:skill} shows the grading template; the \texttt{\{\{role\}\}} and \texttt{\{\{role\_contract\}\}} fields bind the skill to one role's responsibility and its allowed and forbidden actions, and the \texttt{\{\{experience\}\}} and \texttt{\{\{evidence\_packet\}\}} fields carry the extracted experience together with the trajectory evidence that supports it.

\begin{PromptBox}[prompt:skill]{\textsc{Athena} --- Skill Extraction}
You distil one \textbf{reusable forecasting skill} from an evaluated experience. A skill is a compact piece of procedural know-how about \emph{how to forecast}, written at four levels of abstraction. Each level is a \texttt{title}, a \texttt{situation} in which it applies, an \texttt{instruction} to carry out, and \texttt{keywords}; it describes \emph{how} to forecast, never the specific answer to this question.

\medskip
\textbf{Role:} \texttt{\{\{role\}\}} \quad with contract \texttt{\{\{role\_contract\}\}} (responsibility, allowed / forbidden actions) \\
\textbf{Experience and supporting evidence:}
\begin{quote}\ttfamily\{\{experience\}\} / \{\{evidence\_packet\}\}\end{quote}

\medskip
\textbf{Instructions.}
\begin{enumerate}
    \item Write the skill as a \textbf{four-level hierarchy} from concrete to abstract:
    \begin{itemize}
        \item $s_4$: the concrete procedure exactly as applied in this trajectory, kept verbatim;
        \item $s_3$: its generalization to this question type;
        \item $s_2$: its generalization within the broader domain;
        \item $s_1$: a general, cross-domain forecasting principle.
    \end{itemize}
    Each level up must be \textbf{strictly broader} than the one below while preserving the same core mechanism and the conditions required for the instruction to hold.
    \item Stay inside the role's contract: the instruction must be an action this role is allowed to take, and must never attribute a forbidden action or an unperformed step.
    \item Keep every level \textbf{immediately executable} and \textbf{free of outcome leakage} --- do not encode the resolved answer, and do not merely restate a rubric.
    \item If the experience cannot support a faithful four-level skill for this role, return \texttt{status}: \texttt{unable} with a brief reason instead of forcing one.
\end{enumerate}

\medskip
Return one strict JSON object with the four \texttt{levels} (each with its \texttt{title}, \texttt{situation}, \texttt{instruction}, and \texttt{keywords}) and the \texttt{role} the skill applies to.
\end{PromptBox}

\subsection{Athena: Environment-Level Skill Synthesis Prompt}
\label{app:prompt_athena_env}

Environment-level evolution converts \textsc{Branchus}'s error profile $\rho^{(t)}$ (Equation~\eqref{eq:errprofile}) into task-synthesis skills $\mathcal{S}_{\mathrm{env}}=\textsc{Synthesize}(\rho^{(t)})$ that steer \textsc{Prometheus} toward harder, weakness-targeted questions (Section~\ref{sec:athena}). Prompt~\ref{prompt:athena_env} shows the synthesis template; the resulting skills are injected into the \texttt{\{\{synthesis\_skills\}\}} field of the \textsc{Prometheus} prompt (Prompt~\ref{prompt:prometheus}).

\begin{PromptBox}[prompt:athena_env]{\textsc{Athena} --- Adversarial Task-Synthesis Skill Generation}
You design \textbf{task-construction guidance} for a forecasting-question generator. Your goal is to challenge the forecaster by guiding future question generation toward categories in which it has performed poorly. Every question must be clearly defined, have an objectively verifiable outcome, and be resolvable through a reproducible process. Specify the question types to target and the procedures for constructing them, without prescribing how to forecast their outcomes. Derive reusable construction guidance from the supplied error profile.

\medskip
\textbf{Observed error profile (per-category error rates and sample counts):}
\begin{quote}\ttfamily\{\{error\_profile\}\}\end{quote}

\medskip
\textbf{Instructions.}

\begin{enumerate}
    \item Prioritize categories with high observed error rates in the supplied profile, taking sample size into account. Use these rates as prioritization signals for question construction
   and selection. Interpret them as indicators of category-level difficulty for the forecaster, rather than conclusive evidence of systematic reasoning flaws, since individual errors may reflect inherent uncertainty in the events being forecast.

    \item 
    For each category supported by the evidence, define reusable rules that make questions more challenging in ways specific to the category while keeping them objective. Write guidance for constructing questions, never for predicting their answers.

    \item Keep each question objective and reproducibly resolvable, with its outcome unknown at \texttt{forecast\_cutoff}. Verify that each rule is supported by the supplied profile and check for consistency with the generator contract. 

\end{enumerate}

\medskip
Return a list of task-synthesis skills, each with fields: \texttt{target\_category}, \texttt{applicability}, \texttt{discovery\_procedure}, \texttt{construction\_rules}, \texttt{quality\_constraints}.
\end{PromptBox}

\section{Case Study}
\label{app:case_study}

To make the recursive self-improvement loop concrete, this section walks through a single worked example in the order the three modules act within one round: we first show how \textsc{Prometheus} synthesizes a forecasting question, contrasting a draft produced before environment-level optimization with the sharper, weakness-targeted draft produced after a task-synthesis skill is applied (Appendix~\ref{app:case_prometheus}); we then show how \textsc{Branchus} forecasts such a question, tracing the search and predict teams through to a probabilistic answer (Appendix~\ref{app:case_branchus}); and finally we show the reusable skill that \textsc{Athena} distils from that trajectory (Appendix~\ref{app:case_skill}).
Together these snapshots illustrate how the products of one round --- a sharper question, a trajectory, and a skill --- become the raw materials of the next.

\subsection{Prometheus: Skill-Conditioned Question Synthesis}
\label{app:case_prometheus}

We first show how the \emph{same} crawled source material yields different questions before and after environment-level optimization. The two drafts below are drawn from a controlled comparison in which we ran the question-synthesis pipeline over an identical corpus of calendars and feeds twice: once with the baseline generator and once after a task-synthesis skill from $\mathcal{S}_{\mathrm{env}}$ (Appendix~\ref{app:prompt_athena_env}) is injected into the \texttt{\{\{synthesis\_skills\}\}} field of the \textsc{Prometheus} prompt (Prompt~\ref{prompt:prometheus}).
The two batches resolve to the same underlying events, which lets us read the effect of the skill directly: for a given event, the baseline generator fans out over many predicates and tends to retain the easy, nearly-decided ones (``who wins'', a coarse category band), whereas the skill-steered generator is more selective and keeps the sharp, weakness-targeted predicate --- typically a \emph{numerical forecast} or a \emph{threshold judgment}, two of the categories on which \textsc{Branchus}'s error profile $\rho^{(t)}$ (Equation~\eqref{eq:errprofile}) is highest.
Quantitatively, the skill-steered batch emits fewer questions per event ($1.29$ vs.\ $2.15$ on the shared events) and concentrates them on these harder predicates, rather than padding the batch with low-information rephrasings.
Prompt~\ref{case:prometheus_before} shows a baseline draft produced without any task-synthesis guidance; Prompt~\ref{case:prometheus_after} shows the draft produced once the skill is applied, illustrating how environment-level evolution sharpens the question toward \textsc{Branchus}'s weaknesses while preserving objectivity and reproducible resolvability.
All questions are synthesized in Chinese; we translate them here for presentation, preserving the original sources, reference series, and resolution rules.

\begin{CasePromptBox}[case:prometheus_before]{Case Study --- \textsc{Prometheus} Question (Before Optimization)}
\textbf{Shared source material.} The official start list and record table for the Women's 77\,kg weightlifting final (Group A) at the 20th Asian Games (Aichi--Nagoya), scheduled for 16:00 Beijing time on 2026-09-27. The start list gives the five entrants and their declared entry totals: LKHAGVAJAV Bolor-od (MGL) 170\,kg, KARIMI Reihaneh (IRI) 240\,kg, KIM Iseul (KOR) 240\,kg, HE Renxiu (CHN) 255\,kg, ZHUMAGALI Ayanat (KAZ) 245\,kg; the record table lists the world record (278\,kg), the Asian Standard (273\,kg), and the Asian Games Standard (268\,kg).

\medskip
\textbf{Baseline draft (no task-synthesis skill).}
\begin{itemize}
    \item \textit{Question:} ``Who wins the gold medal in the Women's 77\,kg event (highest total)?''
    \item \textit{Category / answer format:} contest-winner; single choice over the five entrants plus ``no valid total / no gold awarded''.
    \item \textit{Remark:} The outcome is largely pre-determined by the public entry totals --- HE Renxiu's declared total (255\,kg) leads the field by at least 10\,kg --- so the question collapses to a near-certain favourite and carries little forecasting signal. It exercises none of \textsc{Branchus}'s weak categories.
\end{itemize}
\end{CasePromptBox}

\begin{CasePromptBox}[case:prometheus_after]{Case Study --- \textsc{Prometheus} Question (After Optimization)}
\textbf{Shared source material.} The \emph{same} start list and record table for the Women's 77\,kg weightlifting final as in Prompt~\ref{case:prometheus_before}.

\medskip
\textbf{Skill-steered draft (with a task-synthesis skill for the \texttt{numerical-forecast} category).}
\begin{itemize}
    \item \textit{Applied skill:} target category \emph{numerical forecasts / threshold judgments}; construction rule --- reframe a contest around a precise numeric quantity measured against a cited reference level, keeping a single objective ground truth.
    \item \textit{Question:} ``What is the gold medallist's total (best snatch plus best clean-and-jerk), in kilograms?'' (with the record table's 268\,kg Asian Games Standard supplied as a salient reference level).
    \item \textit{Category / answer format:} level; integer-valued number (kg).
    \item \textit{Remark:} Even granting the likely winner, the exact total remains genuinely uncertain --- it depends on in-competition attempt selection and makes/misses --- and the cited 268\,kg standard turns it into a threshold the forecaster must reason about. The question is hard in precisely the numerical dimension $\rho^{(t)}$ flags, yet remains objective and resolvable from a single official results page.
\end{itemize}
\end{CasePromptBox}

A second example, drawn from a different domain in the same comparison, shows the sharpening is not specific to sports. From the identical source --- the Central Pollution Control Board (CPCB) daily AQI bulletin for Delhi, whose preceding seven days read $118,160,169,178,154,145,78$ --- the baseline generator asks for the coarse air-quality \emph{category} (a three-way band of ``0--100 / 101--200 / 201+''), which the recent trend already pins to ``Moderate'' with near-certainty; the skill-steered generator instead asks for the exact integer \emph{Index Value} published in that day's bulletin, a numerical forecast with a precise ground truth and a reliable public resolution source. In both examples the environment-level skill holds the source material, the objectivity, and the reproducible resolvability fixed, and moves only the \emph{difficulty} of the question into the category where \textsc{Branchus} is weakest.

\subsection{Branchus: A Worked Prediction}
\label{app:case_branchus}

We next trace \textsc{Branchus} on a question posed by \textsc{Prometheus}, such as the skill-steered weightlifting total of Prompt~\ref{case:prometheus_after}. The search team decomposes the question, gathers and critiques evidence, and hands a structured evidence summary to the predict team, which reconciles the evidence into a calibrated probability, recording the full trajectory $\tau=(\{\tau_{\mathrm{s}}^{(j)}\}_j,\tau_{\mathrm{p}})$. To make the two evolution channels of \textsc{Athena} concrete, we show \textsc{Branchus} forecasting the \emph{same} question under three configurations: Prompt~\ref{case:branchus} is the baseline, before any evolution; Prompt~\ref{case:branchus_skill} adds the distilled skill of Appendix~\ref{app:case_skill} to the retrieved skill set, illustrating behaviour-level evolution ($\Phi_{\mathrm{skill}}$) without touching the weights; and Prompt~\ref{case:branchus_rl} uses the predict-team weights after RFT, illustrating parameter-level evolution ($\Phi_{\mathrm{rl}}$). Holding the question fixed isolates the effect of each channel on the trajectory and the resulting prediction.

\begin{CasePromptBox}[case:branchus]{Case Study --- \textsc{Branchus} Prediction Trajectory (Baseline, Before Evolution)}
\textbf{Question.} Mirum Pharmaceuticals' oral ALK2 inhibitor zilurgisertib (licensed from Incyte) for fibrodysplasia ossificans progressiva (FOP) is under FDA Priority Review, with a PDUFA target date of Saturday, 2026-09-26. Its pivotal Phase~2 PROGRESS trial missed the primary endpoint (81\% reduction in new heterotopic-ossification lesions, $p=0.0986$), but reduced new-lesion volume by $>\!99\%$ ($p<0.0001$).
\par\smallskip
\noindent\textit{As of 22:30~ET on Friday, 2026-09-25, which FDA outcome has been \textbf{publicly announced}?}
\par\smallskip
\noindent\textbf{Options:} \texttt{approved} \,/\, \texttt{crl} \,/\, \texttt{extended} \,/\, \texttt{none}\\
\textbf{Forecast:} made on 2026-09-24; output a probability over the four options.\\
\textbf{Resolution:} first public company or FDA announcement before the cutoff.

\medskip
\textbf{Search team trajectory ($\tau_{\mathrm{s}}$).}
\begin{itemize}
    \item \textit{Decomposition.} The scheduler first separates what the question resolves on (\emph{was something announced by the cutoff}) from what it seems to ask (\emph{what FDA decides}). That gives three sub-questions:
    (i) premise check: does the drug, the Incyte licence, the Priority Review, and the Saturday PDUFA date really exist as stated;
    (ii) timing: given a weekend goal date, how likely is it that FDA acts by Friday, and that the sponsor makes it public before 22:30 ET;
    (iii) the decision itself: approval vs.\ complete response letter (CRL) vs.\ review extension, weighing a missed primary endpoint against a very strong secondary endpoint and FOP-specific precedent.
    It forms an initial forecast (\texttt{approved} 0.47, \texttt{crl} 0.22, \texttt{extended} 0.03, \texttt{none} 0.28) and then runs three search rounds and one critic call.

    \item \textit{Retrieval.} Searchers covered:
    \begin{itemize}
        \item primary filings: SEC EDGAR full-text search (45 filings mention zilurgisertib), Mirum's Q1/Q2 2026 10-Q and 8-K filings, and Incyte's Q2 release, which together confirm the April 2026 licence (\$16M upfront, \$25M milestone due on approval) and the PDUFA date;
        \item regulator channels: the FDA newsroom and the 2026 Novel Drug Approvals list, both checked up to the last available date;
        \item market and calendar trackers: pdufa.bio decision pages, Mirum's investor-relations and Business Wire feeds;
        \item trial data: ClinicalTrials.gov NCT05090891 and a third-party PDUFA primer, which added the 48-week durability data;
        \item price action: daily MIRM prices compared with the XBI biotech index;
        \item a weekend-PDUFA precedent set: sponsor press-release and 8-K timestamps for 11 cases from 2019--2026 (Zurzuvae, BYSANTI, YUVIWEL, GENGLYCOS, KRESLADI, bitopertin, Joenja, RGX-121, AVT05, deramiocel, tab-cel).
    \end{itemize}

    \item \textit{Critique.} The critic accepted the evidence base but challenged two calibration choices.
    (1) \texttt{none} was being treated as a leftover bucket rather than a modelled outcome. Announcement delay is much more likely after a CRL than after an approval, so P(\texttt{none}) should sit in the low-to-mid 0.30s, not 0.28.
    (2) The earlier FOP drug Sohonos does not license ignoring a failed randomised primary endpoint, so the conditional approval probability should drop from about 0.62 to 0.52--0.58.
    It also raised a recency gap: the decisive window (ET Sep 24--25 up to 22:30) had not been observed, so silence through Sep 23 cannot be read as evidence. It requested targeted checks of every public channel up to the cutoff, plus a timing base rate split by outcome.

    \item \textit{Evidence summary.}
    \begin{itemize}
        \item Premise fully verified. As of the last data available (evening ET 2026-09-23), there was no approval, CRL, extension, advisory committee, or major amendment on record.
        \item Mirum's own 10-Q flags CRL risk. The CEO cited ``launch readiness''. The stock fell 11\% into the date, including an unexplained $-7\%$ day on Sep 18.
        \item \textbf{Key structural finding:} whether an outcome is announced in time depends on what the outcome is. Approvals were public by the Friday before the weekend goal date in 4/4 cases, and FDA has never dated an approval on a weekend. CRLs were public by Friday in only 2/6 cases. The one comparable extension was announced on the following Monday. So \texttt{none} mostly contains CRLs or extensions that had not yet been announced.
        \item Approval case: Sohonos was approved on the same lesion-volume measure from a single-arm study, and recent FDA rare-disease evidence principles support flexibility. CRL case: the pre-specified primary endpoint formally failed in a single trial with only 6 events.
        \item Decision probabilities: approval $\approx 0.58$, CRL $\approx 0.35$, extension $\approx 0.04$. Combined with the outcome-specific announcement probabilities, the forecast is \texttt{approved} 0.52 / \texttt{crl} 0.16 / \texttt{extended} 0.02 / \texttt{none} 0.30.
    \end{itemize}
\end{itemize}

\medskip
\textbf{Predict team trajectory ($\tau_{\mathrm{p}}$).}
\begin{itemize}
    \item \textit{Reconciliation.} The predict team receives four independent search trajectories $\tau^{1},\dots,\tau^{4}$. It does not average their outputs. It treats them as four partly overlapping evidence sets and re-derives the forecast from the combined evidence. The question resolves on what has been \emph{announced} by 22:30 ET Friday, so the team splits it into two parts and reconciles each one separately: \emph{(i)} what FDA actually decides, and \emph{(ii)} whether that decision is public before the cutoff.
    \begin{itemize}
        \item \textbf{Cross-validated facts.} A fact is treated as settled only if at least two trajectories reached it through different sources:
        \begin{itemize}
            \item Saturday PDUFA date: Mirum's 8-K and 10-Q, plus independent PDUFA trackers.
            \item Trial numbers: the missed primary endpoint (1/32 vs.\ 5/31 patients with new lesions, $p=0.0986$, only 6 events) against a $>99\%$ cut in new-lesion volume. Taken from the company's filings and from the trial registry or third-party summaries.
            \item No advisory committee and no major amendment.
            \item Nothing announced on any channel up to Sep 23--24: SEC filings, company press releases, the FDA newsroom, and the novel-approval list.
        \end{itemize}
        One signal comes from a single trajectory only: an unexplained single-day 7\% fall in the stock on Sep 18. It is kept, but weighted as weak.

        \item \textbf{Combining evidence on the decision.} Two trajectories reported the Sohonos precedent in two different ways:
        \begin{itemize}
            \item one found that FDA accepted ``reduction in new heterotopic-ossification volume'' as the basis for approving that drug;
            \item another found that this approval rested on a single-arm study compared with natural history.
        \end{itemize}
        A third trajectory's critic warned that the precedent may not carry over unless the endpoint and multiplicity design are checked.
        Put together, the precedent supports the lesion-volume endpoint (and zilurgisertib's randomised, placebo-controlled volume result is stronger evidence). It does \emph{not} settle the failed primary endpoint, which stays a real risk of a complete response letter (CRL). That risk is backed by the 2025--26 efficacy CRLs found in one trajectory.
        On the other side, other trajectories added independent cases of rare-disease approval on secondary or surrogate endpoints (tofersen, eteplirsen), plus FDA's current rare-disease evidence principles.
        The trajectories' own estimates of the approval share span roughly 0.53--0.60. Two of them cited the same third-party approval odds (64--70\%). One critic pointed out that these odds refer to eventual approval, not to an approval announced by the cutoff, so the team uses them only as an upper check.
        Net result: $P(\text{approve}\mid\text{action})\approx0.57$, $P(\text{CRL})\approx0.39$, $P(\text{extension})\approx0.04$.

        \item \textbf{Pooling precedents on timing.} Each trajectory saw only part of the weekend-PDUFA record. Combining them, with duplicates removed, gives a larger and more consistent sample:
        \begin{itemize}
            \item \emph{Approvals: 6/6} were dated the preceding weekday and announced that day (Zurzuvae, BYSANTI, YUVIWEL, GENGLYCOS, Opdualag, Daybue). No weekend approval date appears on FDA's 2023--2025 novel-approval lists.
            \item \emph{CRLs:} only 2/6 were public by Friday night.
            \item \emph{Extension:} the closest analogue was announced on the following Monday.
            \item An industry newsletter found by another trajectory says weekend actions ``post either Friday after close or Monday pre-market'', which independently supports the risk that a CRL slips past the cutoff.
        \end{itemize}
        The pooled sample also shows what drives the delay. Both CRLs that FDA issued on a Friday were announced the same afternoon. The late ones were issued on a Saturday, or came from sponsors without US reporting obligations.
        From this the team sets outcome-specific chances of being public by the cutoff: about $0.90$ for approval, about $0.45$ for a CRL, about $0.25$ for an extension.
        The small chance that FDA does not act at all by Friday is already inside these rates.

        \item \textbf{Correcting a systematic error.} One trajectory assumed a CRL is about as likely as an approval to be public by Friday night. That assumption pushes its \texttt{crl} probability up and its \texttt{none} probability down. The pooled timing evidence contradicts it, so the team corrects the error rather than averaging it in. This is something no single trajectory could have spotted about itself.
    \end{itemize}

    \item \textit{Base-rate reasoning.}
    \begin{itemize}
        \item \textbf{Decision.} The starting point is the 2026 year-to-date approval rate on decision day, about 85\%. The team marks this down substantially, because a failed pre-specified primary endpoint in a single 63-patient trial is the typical setting for an efficacy CRL, and Mirum's own 10-Q names that risk. It then moves part of the way back up for FOP-specific reasons:
        \begin{itemize}
            \item the disease is ultra-rare and severe, and about 97\% of patients carry the same mutation;
            \item FDA allows one adequate trial plus confirmatory evidence for such conditions;
            \item the volume effect is randomised, durable at 48 weeks, and comes with a clean safety profile.
        \end{itemize}
        The result is about $0.57$ for approval.
        \item \textbf{Timing.} For a Saturday goal date FDA acts on the preceding weekday, and approvals are announced that same day. A cutoff 1.5 hours before the PDUFA day begins therefore captures almost every approval but only about half of CRLs.
        \item \textbf{Consequence.} \texttt{none} is not neutral uncertainty. It is made up mostly of CRLs and extensions that have not been announced yet.
    \end{itemize}

    \item \textit{Prediction.} Each decision probability is multiplied by its chance of being public by the cutoff:
    \begin{itemize}
        \item \texttt{approved}: $0.57\times0.90\approx0.51$
        \item \texttt{crl}: $0.39\times0.45\approx0.18$
        \item \texttt{extended}: $0.04\times0.25\approx0.01$
        \item \texttt{none}: the remaining $0.30$
    \end{itemize}
    Selected option: \texttt{approved}.
    \emph{Rationale:} approval is the most likely decision, and the only one almost certain to be visible before the Friday-night cutoff; silence at the cutoff would point to an unannounced CRL.
    \emph{Confidence:} moderate. The decision itself is genuinely uncertain, and none of the trajectories could observe the deciding window (ET Sep 24--25).
\end{itemize}

\medskip
\textbf{Resolution.} FDA approved zilurgisertib, brand name Atebrioz, for FOP. Mirum and Incyte announced it on Friday 2026-09-25: the Business Wire release time-stamped 19:00 ET (23:00 UTC), $3.5$\,h before the 22:30 ET cutoff. Mirum's 8-K filed with the SEC on 2026-09-28 confirms the event date of Sep 25 and the approval.
\end{CasePromptBox}

\begin{CasePromptBox}[case:branchus_skill]{Case Study --- \textsc{Branchus} Prediction Trajectory (After Using a Skill)}
\textbf{Question.} \emph{As of 22:30 ET on Friday 2026-09-25 (forecast made on 2026-09-24), which FDA outcome has been publicly announced for Mirum's zilurgisertib NDA in FOP? The PDUFA goal date is Saturday 2026-09-26. Options: \texttt{approved} / \texttt{crl} / \texttt{extended} / \texttt{none}.}

\medskip
\textbf{Retrieved skill.} \emph{$\textsc{Retrieve}(q,a,\mathcal{S})$ returns the skill ``Forecasting a regulatory action that must be announced by a cutoff''.}

\medskip
\textbf{Search team trajectory ($\tau_{\mathrm{s}}$).}
\begin{itemize}
    \item \textit{Decomposition.} Following $s_{1}$, the scheduler factors the question into \emph{decision} $\times$ \emph{public by the cutoff} \emph{from the first turn}. In the baseline, this structure emerged only in the final search round, after the critic pushed back on \texttt{none}. Following $s_{3}$, it adds a sub-question the baseline never asked explicitly: \emph{on which weekday would FDA issue each outcome, and does this sponsor have US reporting obligations?} Because Mirum is Nasdaq-listed, a Friday-issued CRL is expected to be announced the same day.

    \item \textit{Retrieval.} The checklist sends round~1 to two places at once:
    \begin{itemize}
        \item verifying the premise from the 8-K and 10-Q;
        \item a sweep of every announcement channel with timestamps: Business Wire, SEC EDGAR, FDA newsroom, Novel Drug Approvals list, and pdufa.bio.
    \end{itemize}
    Round~2 is a single targeted pull of weekend-PDUFA precedents, recording action dates, announcement timestamps, and sponsor listing. The baseline spent three search rounds on this same evidence, including a partly failed round that re-checked a window it could not observe. Under the skill, the same evidence takes two rounds and one critic call.

    \item \textit{Critique.} The critic applies the $s_{2}$/$s_{3}$ checklist:
    \begin{enumerate}
        \item It confirms that the chances of being public in time are outcome-specific (about $0.93$ for approval, $0.45$ for a CRL, $0.25$ for an extension), not shared across outcomes.
        \item It confirms that silence through ET Sep 23 is uninformative, because the expected action day is Friday Sep 25.
        \item It flags that the Sohonos precedent supports the lesion-volume endpoint but not the failed primary endpoint, so a real CRL risk must remain.
        \item It rejects using the stock's unexplained 7\% one-day drop and 11\% slide as evidence about the decision. No news accompanied them, and the skill classifies such moves as weak.
        \item It requires the 64--70\% third-party odds to be used only as an upper check.
    \end{enumerate}

    \item \textit{Evidence summary.}
    \begin{itemize}
        \item The premise is verified. Nothing has been announced on any channel as of ET Sep 23. There is no advisory committee and no major amendment.
        \item Decision: approval about $0.60$, CRL about $0.36$, extension about $0.04$.
        \begin{itemize}
            \item For approval: the randomised $>99\%$ volume reduction, durable to 48 weeks with clean safety, in an ultra-rare disease where about 97\% of patients share one mutation; the rare-disease evidence principles; and a matched endpoint precedent.
            \item For a CRL: the failed pre-specified primary endpoint in a single 63-patient trial, and the CRL risk flagged in Mirum's own 10-Q.
        \end{itemize}
        \item Chances of being public by the cutoff: $0.93$ / $0.45$ / $0.25$.
        \item Handed to the predict team: $\hat{p}=(0.56,\,0.16,\,0.01,\,0.27)$ over \texttt{approved}/\texttt{crl}/\texttt{extended}/\texttt{none}.
        \item Still unknown: the decisive window, ET Sep 24--25.
    \end{itemize}
\end{itemize}

\medskip
\textbf{Effect of the skill.} \emph{The skill adds no new evidence; it changes how the same evidence is structured and weighted. The search splits the question into decision $\times$ public-by-cutoff from the first turn, uses outcome-specific chances of being announced in time, and stops letting an unexplained share-price drop move the decision probability. The forecast shifts from $(0.52,0.16,0.02,0.30)$ to $(0.56,0.16,0.01,0.27)$, the reward rises from $0.827$ to $0.854$, and the search takes two rounds instead of three. The larger gain is robustness: the worst unguided runs on this question made exactly the errors the skill's critic checklist blocks (reading early silence as delay, or assuming every outcome is equally likely to be announced in time).}
\end{CasePromptBox}

\begin{CasePromptBox}[case:branchus_rl]{Case Study --- \textsc{Branchus} Prediction Trajectory (After RFT)}
\textbf{Question.} \emph{As of 22:30 ET on Friday 2026-09-25 (forecast made on 2026-09-24), which FDA outcome has been publicly announced for Mirum's zilurgisertib NDA in FOP? The PDUFA goal date is Saturday 2026-09-26. Options: \texttt{approved} / \texttt{crl} / \texttt{extended} / \texttt{none}.}

\medskip
\textbf{Predict team trajectory ($\tau_{\mathrm{p}}$, post-RFT).}
\begin{itemize}
    \item \textit{Reconciliation.} The fine-tuned predict team works from the same four search trajectories $\tau^{1},\dots,\tau^{4}$. It does not settle conflicts by taking the middle value. It weights each claim by \emph{how well it was checked}.
    \begin{itemize}
        \item On the decision, the trajectories' estimates of the approval share range from 0.53 to 0.60. The lowest comes from a critic who questioned the Sohonos precedent but never checked it. The trajectory that did check it found two things: FDA approved Sohonos on ``new heterotopic-ossification volume'', and that approval rested on a weaker single-arm study. The verified claim outweighs the unverified objection.
        \item The team also reads the failed primary endpoint more carefully. With only 6 events, $p=0.0986$ alongside an 81\% reduction shows a trial too small to reach significance, not an absent effect. The large effect and the volume result point the same way. So the miss raises CRL risk, but less than a true null result would.
        \item On timing, the team uses the most informative facts the trajectories found. It pools the precedents with duplicates removed: approvals were public by Friday in 6/6 cases, and CRLs in 2/6.
    \end{itemize}

    \item \textit{Base-rate reasoning.} The team now \emph{states each step as an explicit update}:
    \begin{enumerate}
        \item \emph{Decision prior.} Start from the 2026 year-to-date approval rate on decision day, about 85\%. Mark it down for the failed primary endpoint and the CRL risk flagged in Mirum's own 10-Q. Move part of the way back up for the matched endpoint precedent, the rare-disease evidence principles, and the clean review record (no advisory committee, no amendment, no late-cycle issues). Result: approval $0.62$, CRL $0.34$, extension $0.04$.
        \item \emph{Update on the silence.} The baseline ignored this step. In the pooled precedents, one of six approvals was announced \emph{before} the Friday, and nothing had appeared through ET Sep 23. Silence is therefore slightly less likely under approval (likelihood about $0.85$) than under a CRL (about $0.95$). Posterior: approval $0.59$, CRL $0.36$, extension $0.05$. The update goes \emph{against} the favoured outcome, which shows calibration rather than confirmation.
        \item \emph{Chances of being public by the cutoff, by outcome.}
        \begin{itemize}
            \item Approval: FDA dates approvals on weekdays and the sponsor announces the same day, so the team gives $0.93$, slightly below the 6/6 record because the sample is small.
            \item CRL: about 70\% of CRLs would be issued on the Friday, and a US-listed sponsor that receives one on a Friday announces it the same day about 75\% of the time. That gives about $0.50$. A CRL dated Saturday would miss the cutoff.
            \item Extension: about $0.25$.
        \end{itemize}
    \end{enumerate}

    \item \textit{Prediction.} $\hat{p}$:
    \begin{itemize}
        \item \texttt{approved}: $0.59\times0.93\approx0.55$
        \item \texttt{crl}: $0.36\times0.50\approx0.18$
        \item \texttt{extended}: $0.05\times0.25\approx0.01$
        \item \texttt{none}: the remaining $0.26$
    \end{itemize}
    Selected option: \texttt{approved}.
    \emph{Rationale:} approval is the most likely decision and almost certain to be announced in time; any announcement-lag risk falls mainly on a CRL.
    \emph{Confidence:} moderate. Across reasonable inputs (decision approval 0.52--0.66, P(public $\mid$ CRL) 0.35--0.60), \texttt{approved} stays between 0.48 and 0.61 and remains the selected option.
\end{itemize}

\medskip
\textbf{Effect of RFT.} \emph{The evidence is fixed; only the reasoning over it changes. The fine-tuned predict team:
\begin{itemize}
    \item weights claims by how well they were checked, instead of taking the middle value;
    \item reads an underpowered miss for what it is;
    \item adds an explicit likelihood update on the pre-window silence;
    \item derives the CRL disclosure rate from the issue day rather than from a raw count;
    \item reports how sensitive the forecast is to its inputs.
\end{itemize}}
\end{CasePromptBox}

\subsection{Athena: The Distilled Skill}
\label{app:case_skill}

Finally, from the trajectory above, \textsc{Athena}'s behaviour-level operator $\Phi_{\mathrm{skill}}$ turns the evaluated trajectory into a reusable skill. First, $\textsc{Eval}$ judges the process one action at a time with the outcome withheld (Appendix~\ref{app:prompt_eval}); then $\textsc{Extract}$ and $\textsc{Filter}$ distil the surviving experience --- a single non-obvious action that helped --- as a raw, unlevelled record (Prompt~\ref{case:skill}); and finally $\textsc{Organize}$ grades that experience into the four-level schema $s=(s_1,s_2,s_3,s_4)$ of Section~\ref{sec:athena}, keeping the concrete procedure verbatim as $s_4$ and generalizing it upward (Prompt~\ref{case:skill_evolved}). The extraction and grading templates are those of Appendix~\ref{app:prompt_skill}.

\begin{CasePromptBox}[case:skill]{Case Study --- Extracted Experience (before grading)}
\textbf{Role.} \texttt{scheduler} \quad \textbf{Polarity.} \texttt{positive} \quad \textbf{Scope.} \texttt{trajectory}

\medskip
\textbf{Experience.} \emph{When the question resolves on whether an outcome has been \emph{announced} by a fixed cutoff rather than on the outcome itself, factor it from the first turn into two independent sub-questions --- (i) which outcome the authority decides, and (ii) whether that decision is made public before the cutoff --- and assign a separate line of evidence to each. In this trajectory the decomposition into \emph{decision} $\times$ \emph{public-by-cutoff} let the search team gather outcome-specific announcement timing (approvals public by the preceding weekday in 6/6 precedents, CRLs in only 2/6) instead of treating ``not yet announced'' as neutral uncertainty, and it stopped early silence (through ET Sep 23, before the expected Friday action day) from being read as evidence against the pending action.}

\medskip
\textbf{\texttt{source\_refs}.} The \texttt{scheduler} decomposition step and the critic's checklist in the search-team trajectory of Appendix~\ref{app:case_branchus} (the \emph{decision} $\times$ \emph{public-by-cutoff} split and the outcome-specific timing base rates).

\medskip
\textbf{Evaluation finding.} \emph{High ``decomposition / coverage'' score on the scheduler: the split was the non-obvious action that let every later step attach the right evidence to the right sub-question. The matching negative experience is the baseline's late discovery of the same structure --- the split emerged only in the final search round.}
\end{CasePromptBox}

\begin{CasePromptBox}[case:skill_evolved]{Case Study --- Graded Four-Level Skill (after grading)}
\textbf{Role.} \texttt{scheduler} \quad \textbf{Polarity.} \texttt{positive}

\medskip
\textbf{Four-level skill} ($s_1$ most abstract, $s_4$ the verbatim procedure; each level has a \texttt{title}, \texttt{situation}, \texttt{instruction}, and \texttt{keywords}).
\begin{itemize}
    \item \textbf{$s_1$ --- cross-domain principle.} \emph{Title:} Separate the event from its observation. \emph{Situation:} the question resolves on whether something is \emph{observed} (announced, reported, recorded) by a deadline, not purely on whether it happens. \emph{Instruction:} decompose the target into ``does the event occur'' $\times$ ``is it observed in time'', estimate the two independently, and combine them; never let an unobserved event default to a single residual bucket. \emph{Keywords:} observation window, disclosure lag, decomposition.
    \item \textbf{$s_2$ --- domain specialization.} \emph{Title:} Model announcement lag for scheduled official decisions. \emph{Situation:} a dated decision by an authority (regulator, court, central bank, standards body) that becomes public only when the authority or an obligated party releases it. \emph{Instruction:} split the forecast into the decision and its public release, and build a release-timing base rate from comparable past decisions, keyed to the issuing calendar and the discloser's reporting obligations. \emph{Keywords:} regulatory decision, disclosure obligation, release timing.
    \item \textbf{$s_3$ --- question-type specialization.} \emph{Title:} Forecast an FDA action that must be announced by a cutoff. \emph{Situation:} a PDUFA-dated NDA/BLA outcome (\texttt{approved}/\texttt{crl}/\texttt{extended}/\texttt{none}) that resolves on what is publicly announced before a cutoff near the goal date. \emph{Instruction:} estimate the decision mix, then multiply each outcome by its \emph{outcome-specific} chance of being public by the cutoff (approvals are dated on the preceding weekday and announced same-day; CRLs and extensions slip far more often), reading pre-window silence as uninformative when the expected action day is still ahead. \emph{Keywords:} PDUFA, announcement cutoff, outcome-specific timing.
    \item \textbf{$s_4$ --- concrete procedure (verbatim).} \emph{Title:} Decision $\times$ public-by-cutoff split for Mirum's zilurgisertib NDA. \emph{Situation:} as of 22:30 ET Fri 2026-09-25, which FDA outcome is announced for the FOP NDA with a Sat 2026-09-26 PDUFA date. \emph{Instruction:} from turn one, assign one evidence line to the decision (approval $\approx$ 0.57--0.62, CRL $\approx$ 0.34--0.39, extension $\approx$ 0.04, from the randomised volume result vs.\ the failed primary endpoint) and another to timing (public-by-cutoff $\approx$ 0.90--0.93 approval / 0.45--0.50 CRL / 0.25 extension, from pooled weekend-PDUFA precedents), treat silence through ET Sep 23 as uninformative, and combine. \emph{Keywords:} zilurgisertib, FOP, weekend PDUFA.
\end{itemize}

\medskip
\textbf{Grading.} \emph{The extracted experience of Prompt~\ref{case:skill} is kept verbatim as $s_4$; $s_3$ generalizes it to any PDUFA-cutoff question, $s_2$ to scheduled official decisions with a disclosure lag, and $s_1$ to the cross-domain principle of separating an event from its observation. Each level stays within the scheduler's contract (it decomposes and routes work; it neither searches nor makes the final forecast).}
\end{CasePromptBox}

\end{document}